\documentclass{article}

\PassOptionsToPackage{numbers}{natbib}

    \usepackage[preprint]{neurips_2020}

\usepackage[utf8]{inputenc} 
\usepackage[T1]{fontenc}    
\usepackage{hyperref}       
\usepackage{url}            
\usepackage{booktabs}       
\usepackage{amsfonts}       
\usepackage{nicefrac}       
\usepackage{natbib}
\usepackage{microtype}      
\usepackage{graphicx, amsmath, subcaption,amssymb, comment, bm, amsthm, bbm, color, soul}
\usepackage{sidecap, floatrow}
\usepackage{multirow}
\usepackage{algorithm}
\usepackage{algorithmic}

\usepackage[font=small,labelfont=bf,tableposition=top]{caption}

\DeclareCaptionLabelFormat{andtable}{#1~#2  \&  \tablename~\thetable}

\newcommand{\tuple}[1]{\langle #1 \rangle}
\DeclareMathOperator*{\argmax}{arg\,max}

\title{Too many cooks: Bayesian inference for coordinating multi-agent collaboration}

\author{
    Rose E. Wang$^*$\\
    MIT\\
    \texttt{rewang@mit.edu}\\
    \And
    Sarah A. Wu\thanks{indicates equal contribution}\\
    MIT\\
    \texttt{sarahawu@mit.edu}\\
    \And
    James A. Evans \\
    UChicago\\
    \texttt{jevans@uchicago.edu}\\
    \And
    Joshua B. Tenenbaum\\
    MIT\\
    \texttt{jbt@mit.edu}\\
    \And 
    David C. Parkes\\
    Harvard\\
    \texttt{parkes@eecs.harvard.edu}\\
    \And 
    Max Kleiman-Weiner\\
    Harvard, MIT \& Diffeo\\
   \texttt{maxkw@mit.edu}
}

\begin{document}

\maketitle

\begin{abstract}
  Collaboration requires agents to coordinate their behavior on the fly, sometimes cooperating to solve a single task together and other times dividing it up into sub-tasks to work on in parallel. Underlying the human ability to collaborate is theory-of-mind, the ability to infer the hidden mental states that drive others to act. Here, we develop Bayesian Delegation, a decentralized multi-agent learning mechanism with these abilities. Bayesian Delegation enables agents to rapidly infer the hidden intentions of others by inverse planning. We test Bayesian Delegation in a suite of multi-agent Markov decision processes inspired by cooking problems. On these tasks, agents with Bayesian Delegation coordinate both their high-level plans (e.g. what sub-task they should work on) and their low-level actions (e.g. avoiding getting in each other's way). In a self-play evaluation, Bayesian Delegation outperforms alternative algorithms. Bayesian Delegation is also a capable ad-hoc collaborator and successfully coordinates with other agent types even in the absence of prior experience. Finally, in a behavioral experiment, we show that Bayesian Delegation makes inferences similar to human observers about the intent of others. Together, these results demonstrate the power of Bayesian Delegation for decentralized multi-agent collaboration. 

\end{abstract}

\section{Introduction}
\label{introduction}
Working together enables a group of agents to achieve together what no individual could achieve on their own \cite{tomasello2014natural,henrich2015secret}. However, collaboration is challenging as it requires agents to coordinate their behaviors. In the absence of prior experience, social roles, and norms, we still find ways to negotiate our joint behavior in any given moment to work together with efficiency \cite{tomasello2005understanding,misyak2014unwritten}. Whether we are writing a scientific manuscript with collaborators or preparing a meal with friends, core questions we ask ourselves are: how can I help out the group? What should I work on next, and with whom should I do it with? Figuring out how to flexibly coordinate a collaborative endeavor is a fundamental challenge for any agent in a multi-agent world.

Central to this challenge is that agents' reasoning about what they should do in a multi-agent context depends on the future actions and intentions of others. When agents, like people, make independent decisions, these intentions are unobserved. Actions can reveal information about intentions, but predicting them is difficult because of uncertainty and ambiguity -- multiple intentions can produce the same action. In humans, the ability to understand intentions from actions is called theory-of-mind (ToM). Humans rely on this ability to cooperate in coordinated ways, even in novel situations \cite{tomasello2005understanding, shum2019theory}. We aim to build agents with theory-of-mind and use these abilities for coordinating collaboration. 

In this work, we study these abilities in the context of multiple agents cooking a meal together, inspired by the video game \textit{Overcooked} \cite{overcooked2016}. These problems have hierarchically organized sub-tasks and share many features with other object-oriented tasks such as construction and assembly. 
%
These sub-tasks allow us to study agents that are challenged to coordinate in three distinct ways:  (A) Divide and conquer: agents should work in parallel when sub-tasks can be efficiently carried out individually, (B) Cooperation: agents should work together on the same sub-task when most efficient or necessary, (C) Spatio-temporal movement: agents should avoid getting in each other's way at any time. 

To illustrate, imagine the process required to make a simple salad: first chopping both tomato and lettuce and then assembling them together on a plate. Two people might collaborate by first dividing the sub-tasks up: one person chops the tomato and the other chops the lettuce. This doubles the efficiency of the pair by completing sub-tasks in parallel (challenge A). On the other hand, some sub-tasks may require multiple to work together. If only one person can use the knife and only the other can reach the tomatoes, then they must cooperate to chop the tomato (challenge B). In all cases, agents must coordinate their low-level actions in space and time to avoid interfering with others and be mutually responsive (challenge C).

\begin{figure}[bt]
    \begin{subfigure}[b]{.24\linewidth}
      \centering
      \includegraphics[width=25mm]{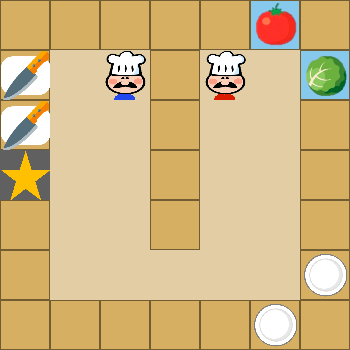}
      \caption{\textit{Partial-Divider}.}
      \label{fig:env_partial}
    \end{subfigure}
    \begin{subfigure}[b]{.34\linewidth}
    \scriptsize{
      \begin{center}
        \includegraphics[width=30mm]{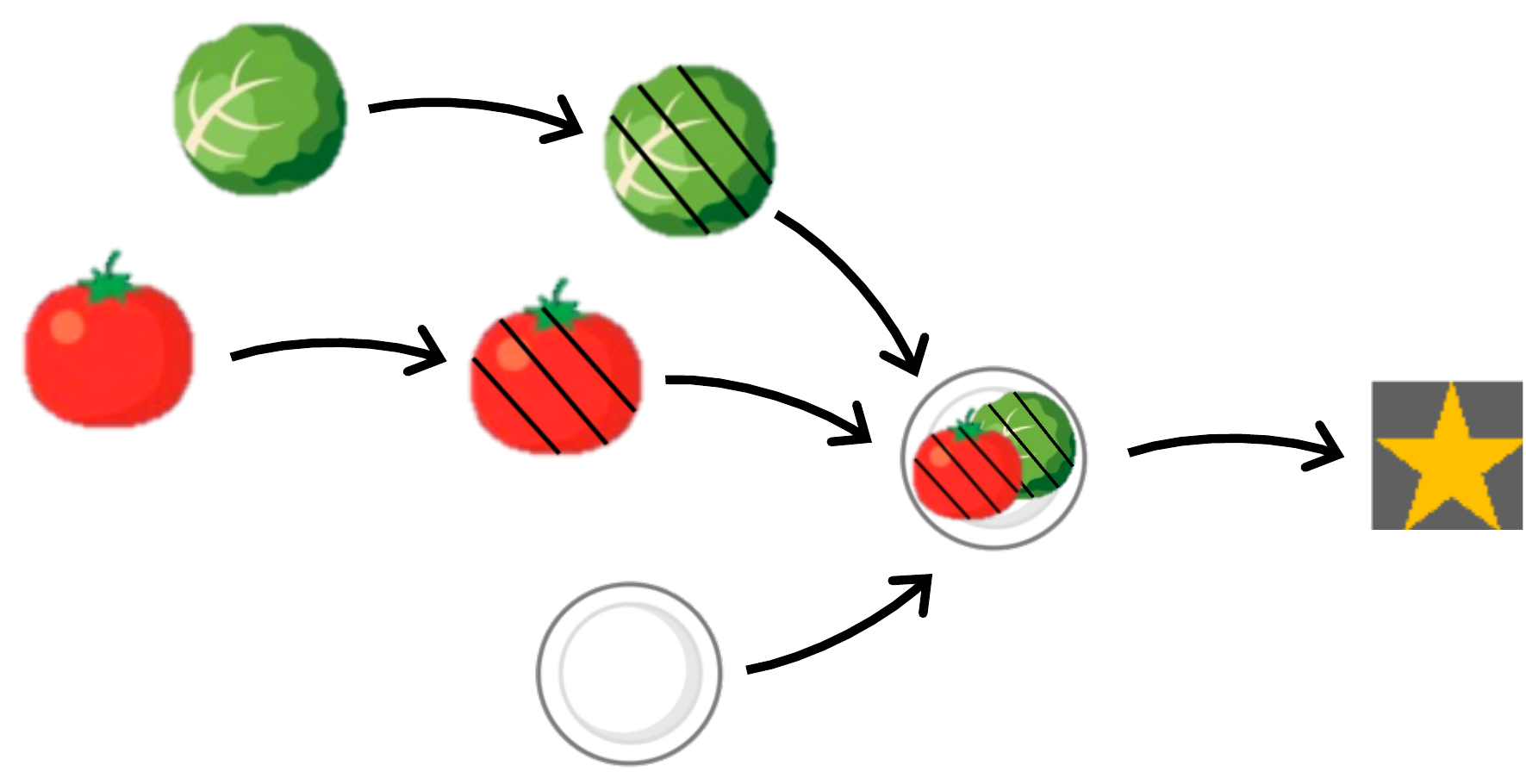} \\
        \vspace{1mm}
        \textbf{Goal} \\
        \texttt{Delivery[Plate[Tomato.chopped, Lettuce.chopped]]}
      \end{center} }
      \caption{\textit{Salad}.}
        \label{fig:env_salad}
  \end{subfigure}
  \begin{subfigure}[b]{.4\linewidth}
      \scriptsize{
      \texttt{Merge(Tomato.unchopped, Knife) \\
        Merge(Lettuce.unchopped, Knife) \\
        Merge(Tomato.chopped, Lettuce.chopped) \\
        Merge([Tomato.chopped Lettuce.chopped], Plate[]) \\
        Merge(Plate[Tomato.chopped, Lettuce.chopped], Delivery)
      }}
      \caption{Example sub-task order for \textit{Salad}.} \label{fig:env_plan}
  \end{subfigure}
  
    \caption{The Overcooked environment. (a) The \textit{Partial-Divider} kitchen offers many counters for objects, but forces agents to move through a narrow bottleneck. (b) The \textit{Salad} recipe in which two chopped foods must be combined on a single plate and delivered, and (c) one of the many possible orderings for completing this task. All sub-tasks are expressed in the \texttt{Merge} operator. Different recipes are possible in each kitchen, allowing for variation in high-level goals while keeping the low-level navigation challenges fixed. 
    \label{fig:env}}
\end{figure}

\paragraph{Related Work}
Our work builds on a long history of using cooking tasks for evaluating multi-agent coordination across hierarchies of sub-tasks \cite{grosz1996collaborative,cohen1991teamwork,tambe1997towards}. Most recently, environments inspired by \textit{Overcooked} have been used in deep reinforcement learning studies where agents are trained using self-play and human data \cite{song2019diversity,carroll2019on}. In contrast, our approach is based on techniques that dynamically learn while interacting rather than requiring large amounts of pre-training experience for a specific environment, team configuration, and sub-task structure. 
Instead our work shares goals with the ad-hoc coordination literature, where agents must adapt on the fly to variations in task, environment, or team \cite{chalkiadakis2003coordination, stone2010ad,barrett2011empirical}. 
However, prior work is often limited to action coordination (e.g,. chasing or hiding) rather than coordinating actions across and within sub-tasks. Our approach to this problem takes inspiration from the cognitive science of how people coordinate their cooperation in the absence of communication \cite{kleiman2016coordinate}. Specifically, we build on recent algorithmic progress in  Bayesian theory-of-mind \cite{ramirez2011goal,nakahashi2016modeling,baker2017rational, shum2019theory} and learning statistical models of others \cite{barrett2012learning,melo2016ad}, and extend these works to decentralized multi-agent contexts.

Our strategy for multi-agent hierarchical planning builds on previous work linking high-level coordination (sub-tasks) to low-level navigation (actions) \cite{amato2019modeling}. In contrast to models which have explicit communication mechanism or centralized controllers \cite{mcintire2016iterated,brunet2008consensus}, our approach is fully decentralized and our agents are never trained together. Prior work has also investigated ways in which multi-agent teams can mesh inconsistent plans (e.g. two agents doing the same sub-task by themselves) into consistent plans (e.g. the agents perform different sub-tasks in parallel) \cite{cox2004efficient, cox2005efficient}, but these methods have also been centralized. 
We draw more closely from decentralized multi-agent planning approaches in which agents aggregate the effects of others and best respond \cite{claes2015effective, claes2017decentralised}. These prior works focus on tasks with spatial sub-tasks called \textit{Spatial Task Allocation Problems} (SPATAPs). However, there are no mechanisms for agents to cooperate on the same sub-task as each sub-task is spatially distinct.

\paragraph{Contributions}
We develop \textit{Bayesian Delegation}, a new algorithm for decentralized multi-agent coordination that rises to the challenges described above. Bayesian Delegation leverages Bayesian inference with inverse planning to rapidly infer the sub-tasks others are working on. Our probabilistic approach allows agents to predict the intentions of other agents under uncertainty and ambiguity. These inferences allow agents to efficiently delegate their own efforts to the most high-value collaborative tasks for collective success. 
We quantitatively measure the performance of Bayesian Delegation in a suite of novel multi-agent environments. First, Bayesian Delegation outperforms existing approaches, completing all environments in less time than alternative approaches and maintaining performance even when scaled up to larger teams. 
Second, we show  Bayesian Delegation is an ad-hoc collaborator. It performs better than baselines when paired with alternative agents. 
Finally, in a behavioral experiment, human participants observed others interact and made inferences about the sub-tasks they were working on. Bayesian Delegation aligned with many of the fine-grained variations in human judgments. Although the model was never trained on human data or other agents' behavior, it was the best ad-hoc collaborator and predictor of human inferences. 


\section{Multi-Agent MDPs with Sub-Tasks \label{section:formalism}} 

A multi-agent Markov decision process (MMDP) with sub-tasks is described as a tuple $\tuple{n, \mathcal{S},\mathcal{A}_{1\ldots n}, T, R, \allowbreak \gamma, \mathcal{T}}$ where $n$ is the number of agents, $s \in \mathcal{S}$ are object-oriented states specified by the locations, status and type of each object and agent in the environment \cite{boutilier1996planning, diuk2008object}. $\mathcal{A}_{1 \ldots n}$ is the joint action space with $a_i \in \mathcal{A}_i$ being the set of actions available to agent $i$; each agent chooses its own actions independently. $T(s,a_{1 \ldots n},s')$ is the transition function which describes the probability of transitioning from state $s$ to $s'$ after all agents act $a_{1 \ldots n}$. $R(s,a_{1 \ldots n})$ is the reward function shared by all agents and $\gamma$ is the discount factor. Each agent aims to find a policy $\pi_i(s)$ that maximizes expected discounted reward.  The environment state is fully observable to all agents, but agents do not observe the policies $\pi_{-i}(s)$ ($-i$ refers to all other agents except $i$) or any other internal representations of others agents. 

Unlike traditional MMDPs, the environments we study have a partially ordered set of sub-tasks $\mathcal{T} = \{ \mathcal{T}_0 \ldots \mathcal{T}_{|\mathcal{T}|}\}$. Each sub-task $\mathcal{T}_i$ has preconditions that specify when a sub-task can be started, and postconditions that specify when it is completed. They provide structure when $R$ is very sparse. These sub-tasks are also the target of high-level coordination between agents. In this work, all sub-tasks can be expressed as \texttt{Merge(X,Y)}, that is, to bring \texttt{X} and \texttt{Y} into the same location. Critically, unlike in SPATAPs, this location is not fixed or predetermined if both \texttt{X} and \texttt{Y} are movable. In the cooking environments we study here, the partial order of sub-tasks refers to a ``recipe''. Figure~\ref{fig:env} shows an example of sub-task partial orders for a recipe. 

The partial order of sub-tasks ($\mathcal{T}$) introduces two coordination challenges. First, \texttt{Merge} does not specify how to implement that sub-task in terms of efficient actions nor which agent(s) should work on it. Second, because the ordering of sub-tasks is partial, the sub-tasks can be accomplished in many different orders. For instance, in the \textit{Salad} recipe (Figure~\ref{fig:env_salad}), once the tomato and lettuce are chopped, they can: (a) first combine the lettuce and tomato and then plate, (b) the lettuce can be plated first and then add the tomato, or (c) the tomato can be plated first and then add the lettuce. These distinct orderings make coordination more challenging since to successfully coordinate, agents must align their ordering of sub-tasks. 

\subsection{Coordination Test Suite}
We now describe the Overcooked inspired environments we use as a test suite for evaluating multi-agent collaboration. Each environment is a 2D grid-world kitchen. 
Figure~\ref{fig:env_partial} shows an example layout. 
The kitchens are built from counters that contain both movable food and plates and immovable stations (e.g. knife stations). The state is represented as a list of entities and their type, location, and status \cite{diuk2008object}. See Table~\ref{tab:objects} for a description of the different entities, the dynamics of object interactions, and the statuses that are possible. Agents (the chef characters) can move north, south, east, west or stay still. All agents move simultaneously. They cannot move through each other, into the same space, or through counters. If they try to do so, they remain in place instead. Agents pick up objects by moving into them and put down objects by moving into a counter while holding them. Agents chop foods by carrying the food to a knife station. Food can be merged with plates. Agents can only carry one object at a time and cannot directly pass to each other.

The goal in each environment is to cook a recipe in as few time steps as possible. The environment terminates after either the agents bring the finished dish specified by the recipe to the star square or 100 time steps elapse. Figure~\ref{fig:env_all} and Figure~\ref{fig:recipe} describe the full set of kitchen and recipes used in our evaluations.

\section{Bayesian Delegation}
We now introduce \textit{Bayesian Delegation}, a novel algorithm for multi-agent coordination that uses inverse planning to make probabilistic inferences about the sub-tasks other agents are performing. Bayesian Delegation models the latent intentions of others in order to dynamically decide whether to divide-and-conquer sub-tasks or to cooperate on the same sub-task. An action planner finds approximately optimal policies for each sub-task. Note that planning is decentralized at both levels, i.e., agents plan and learn for themselves without access to each other's internal representations. 

Inferring the sub-tasks others are working on enables each agent to select the right sub-task when multiple are possible. Agents maintain and update a belief state over the possible sub-tasks that all agents (including itself) are likely working on based on a history of observations that is commonly observed by all.  
Formally, Bayesian Delegation maintains a probability distribution over task allocations. Let $\mathbf{ta}$ be the set of all possible allocations of agents to sub-tasks where all agents are assigned to a sub-task. For example, if there are two sub-tasks ($[\mathcal{T}_1, \mathcal{T}_2]$) and two agents ($[i, j]$), then $\mathbf{ta} = [(i: \mathcal{T}_1, j : \mathcal{T}_2), (i : \mathcal{T}_2, j : \mathcal{T}_1), (i : \mathcal{T}_1, j : \mathcal{T}_1), (i : \mathcal{T}_2, j : \mathcal{T}_2)]$ where $i: \mathcal{T}_1$ means that agent $i$ is ``delegated'' to sub-task $\mathcal{T}_1$. Thus, $\mathbf{ta}$ includes both the possibility that agents will divide and conquer (work on separate sub-tasks) and cooperate (work on shared sub-tasks). If all agents pick the same $ta \in \mathbf{ta}$, then they will easily coordinate. However, in our environments, agents cannot communicate before or during execution. Instead Bayesian Delegation maintains uncertainty about which $ta$ the group is coordinating on, $P(ta)$. 

At every time step, each agent selects the most likely allocation $ta^*= \argmax_{ta} P(ta | H_{0:T})$, where $P(ta | H_{0:T})$ is the posterior over $ta$ after having observed a history of actions $H_{0:T} = [(s_0,\mathbf{a_0}), \ldots (s_T,\mathbf{a}_T)]$ of $T$ time steps and  $\mathbf{a}_t$ are all agents' actions at time step $t$. The agent then plans the next best action according to $ta^*$ using a model-based reinforcement learning algorithm described below.
This posterior is computed according by Bayes rule: 
\begin{align}
    \label{eq:bayes}
    P (ta | H_{0:T})  \propto P(ta) P(H_{0:T} | ta) =  P(ta) \prod_{t=0}^T P(\mathbf{a_t} | s_t, ta)
\end{align}
where $P(ta)$ is the prior over $ta$ and $P(\mathbf{a_t} | s_t, ta)$ is the likelihood of actions at time step $t$ for all agents. Note that these belief updates do not explicitly consider the private knowledge that each agent has about their own intention at time $T-1$. Instead each agent performs inference based only on the history observed by all, i.e., the information a third-party observer would have access to \cite{nagel1986view}. 
The likelihood of a given $ta$ is the likelihood that each agent $i$ is following their assigned task ($\mathcal{T}_i$) in that $ta$. 
\begin{align}
    \label{eq:softmax}
    P(\mathbf{a_t} | s_t, ta) \propto \prod_{i:\mathcal{T} \in ta} exp(\beta * Q^*_{\mathcal{T}_i}(s,a_i))
\end{align}
where $Q^*_{\mathcal{T}_i}(s,a_i)$, is the expected future reward of $a$ towards the completion of sub-task $\mathcal{T}_i$ for agent $i$.
The soft-max accounts for non-optimal and variable behavior as is typical in Bayesian theory-of-mind \cite{kleiman2016coordinate,baker2017rational,shum2019theory}. $\beta$ controls the degree to which an agent believes others are perfectly optimal. When $\beta \xrightarrow{} 0$, the agent believes others are acting randomly. When $\beta \xrightarrow{} \infty$, the agent believes others are perfectly maximizing. Since the likelihood is computed by planning, this approach to posterior inference is called inverse planning. Note that even though agents see the same history of states and actions, their belief updates will not necessarily be the same because updates come from $Q_{\mathcal{T}_i}$, which is computed independently for each agent and is affected by stochasticity in exploration. 

The prior over $P(ta)$ is computed from the environment state. First, $P(ta)=0$ for all $ta$ that have sub-tasks without satisfied preconditions. 
We set the remaining priors to $P(ta) \propto \sum_{\mathcal{T} \in ta} \frac{1}{V_{\mathcal{T}(s)}}$, where $V_{\mathcal{T}(s)}$ is the estimated value of the current state under sub-task $\mathcal{T}$. 
This gives $ta$ that can be accomplished in less time a higher prior weight. Priors are reinitialized when new sub-tasks have their preconditions satisfied and when others are completed. Figure~\ref{fig:beliefs} shows an example of the dynamics of $P(ta)$ during agent interaction. The figure illustrates how Bayesian delegation enables agents to dynamically align their beliefs about who is doing what (i.e., assign high probability to a single $ta$).

Action planning transforms sub-task allocations into efficient actions and provides the critical likelihood for Bayesian Delegation (see Equation~\ref{eq:bayes}). Action planning takes the $ta$ selected by Bayesian Delegation and outputs the next best action while modeling the movements of other agents. In this work, we use bounded real-time dynamic programming (BRTDP) extended to a multi-agent setting to find approximately optimal Q-values and policies \cite{mcmahan2005bounded}. See Appendix~\ref{appendix:compexperiments} for implementation details.

Agents use $ta^*$ from Bayesian Delegation to address two types of low-level coordination problems: (1) avoiding getting in each others way while working on distinct sub-tasks, and (2) cooperating efficiently when working on a shared sub-task. $ta^*$ contains agent $i$'s best guess about the sub-tasks carried out by others, $\mathcal{T}_{-i}$. In the first case, $\mathcal{T}_{i} \neq \mathcal{T}_{-i}$. Agent $i$ first creates models of the others performing $\mathcal{T}_{-i}$ assuming others agents are stationary ($\pi_{\mathcal{T}_{-i}}^0(s)$, level-0 models).  
These level-0 models are used to reduce the multi-agent transition function to a single agent transition function $T'$ where the transitions of the other agents are assumed to follow the level-0 policies, $T'(s'|s,a_{-i}) = \sum_{a_{i}} T(s'|s,a_{-i},a_i) \prod_{A \in -i}\pi_{\mathcal{T}_{A}}^0(s)$.  
Running BRTDP on this transformed environment finds an approximately optimal level-1 policy $\pi_{\mathcal{T}_i}^1(s)$ for agent $i$ that ``best responds'' to the level-0 models of the other agents. This approach is similar to level-K or cognitive hierarchy \cite{wright2010beyond, kleiman2016coordinate, shum2019theory}.

When $\mathcal{T}_i = \mathcal{T}_{-i}$, agent $i$ attempts to work together on the same sub-task with the other agent(s). The agent simulates a fictitious centralized planner that controls the actions of all agents working together on the same sub-task \cite{kleiman2016coordinate}. This transforms the action space: if both $i$ and $j$ are working on $\mathcal{T}_i$,  then $\mathcal{A}' = a_i \times a_j$. Joint policies $\pi^J_{\mathcal{T}_i}(s)$ can similarly be found by single-agent planners such as BRTDP. Agent $i$ then takes the actions assigned to it under $\pi^J_{\mathcal{T}_i}(s)$. Joint policies enable emergent decentralized cooperative behavior---agents can discover efficient and novel ways of solving sub-tasks as a team such as passing objects across counters. Since each agent is solving for their own $\pi^J_{\mathcal{T}_i}(s)$, these joint policies are not guaranteed to be perfectly coordinated due to stochasticity in the planning process. Note that although we use BRTDP, any other model-based reinforcement learner or planner could also be used.

\section{Results
\label{section:compexperiments}}

\begin{table}[bt]
  \caption{Self-play performance of our model and alternative models with two versus three agents. All metrics are described in the text. See Figure~\ref{fig:perf} for more detailed results on Time Steps and Completion for two agents in self-play, and see Figure~\ref{fig:shuffles} for more detailed results on shuffles. Averages $\pm$ standard error of the mean.
  \label{table:perf}}
  \centering
  \begin{tabular}{llcccc}
    \toprule
                     &  & Time Steps                  & Completion            & Shuffles \\
                     &  & ($\downarrow$ better)       & ($\uparrow$ better)   & ($\downarrow$ better) \\
    \midrule
    \multirow{5}{*}{Two agents} & BD (ours)                   & \textbf{35.29 $\pm$ 1.40}   & \textbf{0.98 $\pm$ 0.06} & \textbf{1.01 $\pm$ 0.05} \\
    & UP                      & 50.42 $\pm$ 2.04            & 0.94 $\pm$ 0.05          & 5.32 $\pm$ 0.03 \\ 
    & FB                      & 37.58 $\pm$ 1.60            & 0.95 $\pm$ 0.04          & 2.64 $\pm$ 0.03 \\
    & D\&C          & 71.57 $\pm$ 2.40            & 0.61 $\pm$ 0.07          & 13.08 $\pm$ 0.05 \\
    & Greedy  & 71.11 $\pm$ 2.41            & 0.57 $\pm$ 0.08          & 17.17 $\pm$ 0.06 \\
    \midrule
    \multirow{4}{*}{Three agents} & BD (ours)    & \textbf{34.52 $\pm$ 1.66}   & \textbf{0.96 $\pm$ 0.08} & 1.64 $\pm$ 0.05 \\
    & UP     & 56.84 $\pm$ 2.12            & 0.91 $\pm$ 0.22          & 5.02 $\pm$ 0.12 \\
    & FB     & 41.34 $\pm$ 2.27            & 0.92 $\pm$ 0.08          & \textbf{1.55 $\pm$ 0.05} \\
    & D\&C      & 67.21 $\pm$ 2.31           & 0.67 $\pm$ 0.15         & 4.94 $\pm$ 0.09 \\
    &  Greedy & 75.87 $\pm$ 2.32             & 0.62 $\pm$ 0.22         & 12.04 $\pm$ 0.13 \\ 
    \bottomrule
  \end{tabular}
\end{table}

\begin{figure}[tb]
  \centering
  \newcommand{\rw}{26mm}
  \newcommand{\lw}{16mm}
  \newcommand{\gw}{20mm}
  \newcommand{\bw}{20mm}
  \newcommand{\graphw}{18mm}
  \newcommand{\verticalspace}{1mm}
  \newcommand{\leftratio}{0.17}
  \newcommand{\rightratio}{0.27}

  \begin{subfigure}[t]{\leftratio\linewidth}
    \centering    
    \phantom{a}
    
    \includegraphics[width=\lw]{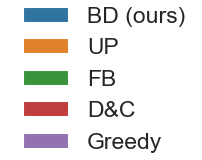}
  \end{subfigure}
  \begin{subfigure}[t]{\rightratio\linewidth}
  \centering
    \scriptsize{\textit{Tomato}}
    
    \includegraphics[width=\rw]{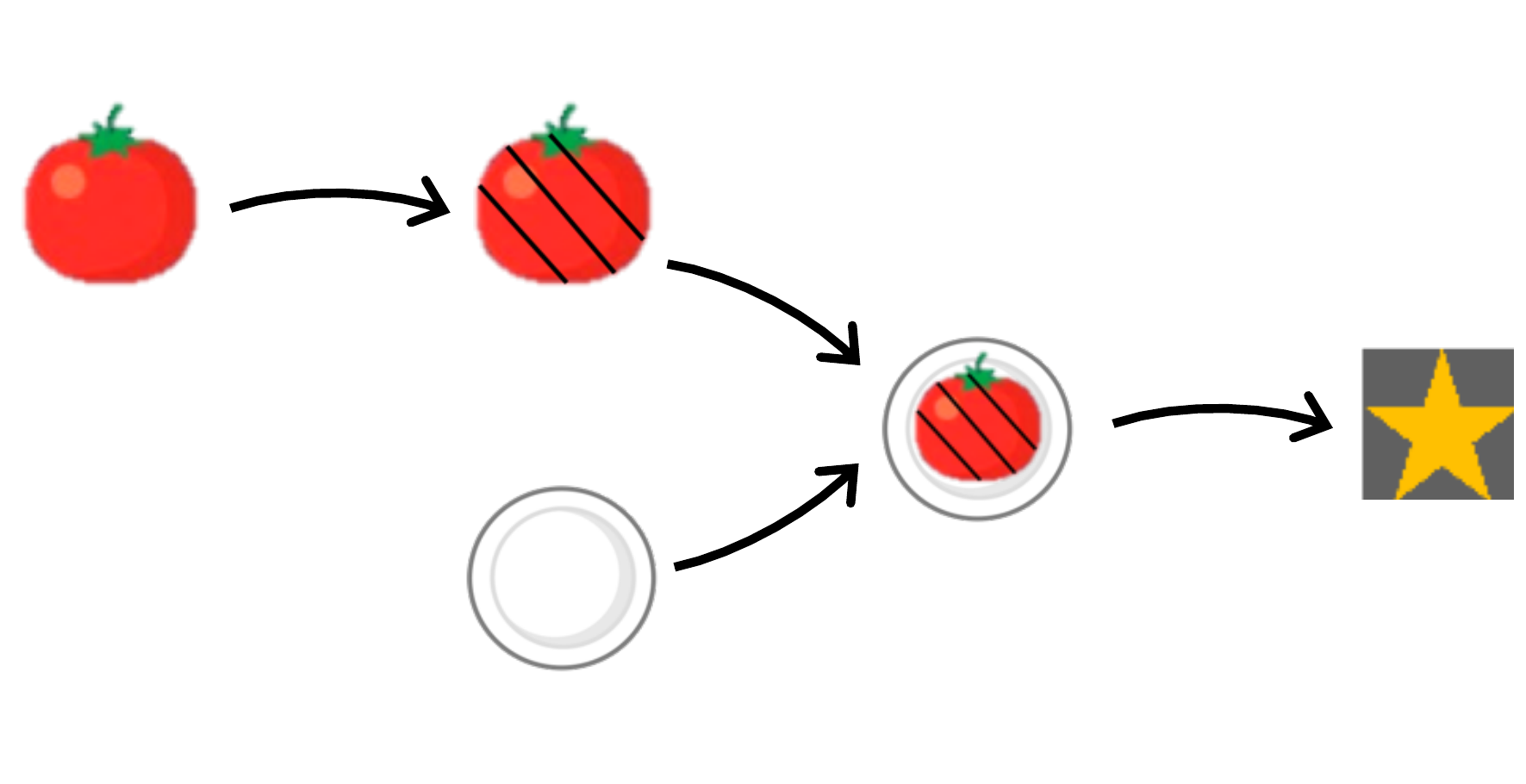}
  \end{subfigure}
  \begin{subfigure}[t]{\rightratio\linewidth}
  \centering
    \scriptsize{\textit{Tomato-Lettuce}}
    
    \includegraphics[width=\rw]{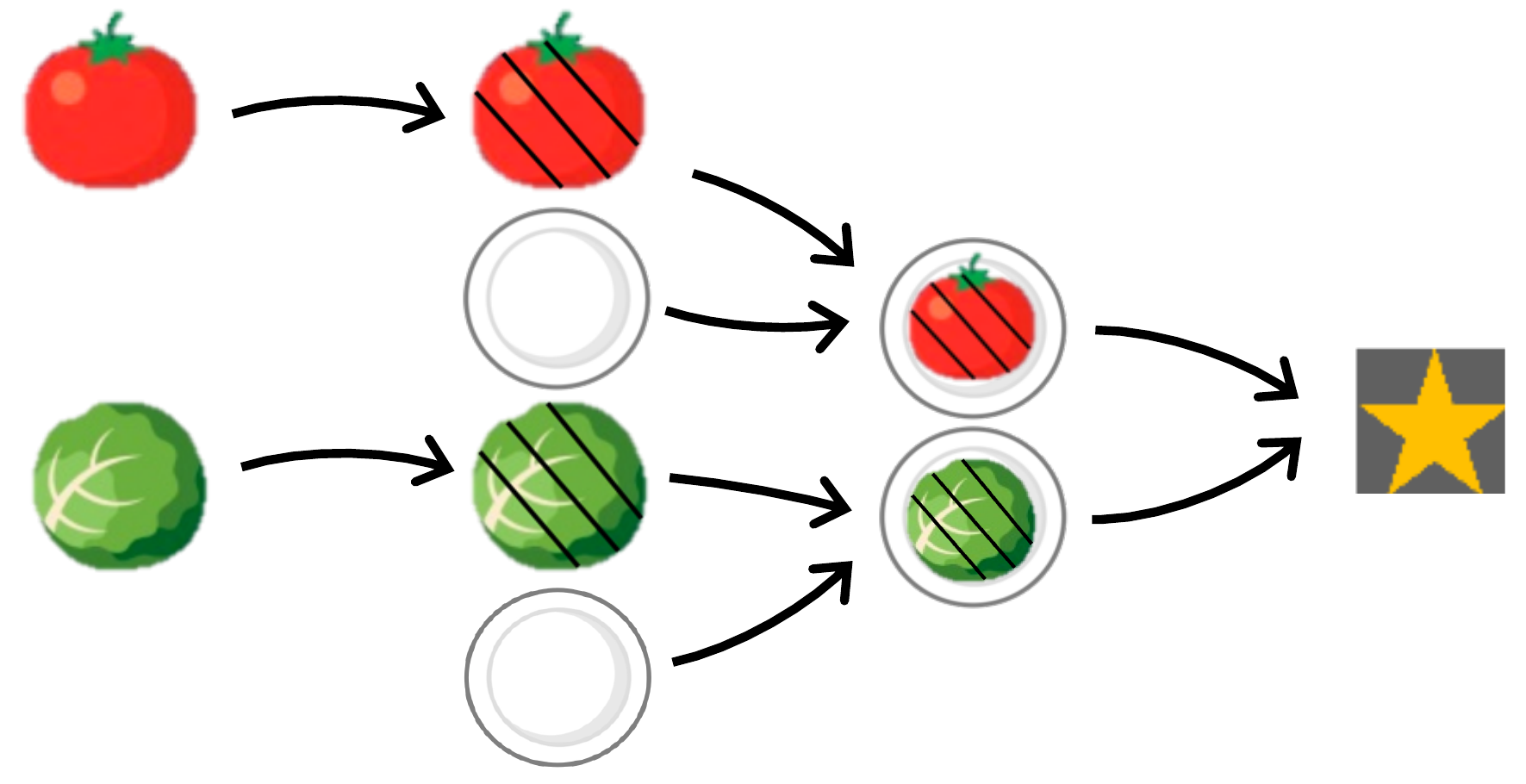}
  \end{subfigure}
  \begin{subfigure}[t]{\rightratio\linewidth}
  \centering
    \scriptsize{\textit{Salad}}
    
    \includegraphics[width=\rw]{images/recipes/Salad}
  \end{subfigure}

  \begin{subfigure}[t]{\leftratio\linewidth}
  \centering
      \rotatebox{90}{\scriptsize{\quad\textit{Open-Divider}}}\includegraphics[width=\gw]{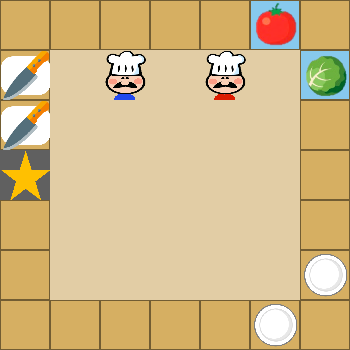}
  \end{subfigure}
  \begin{subfigure}[t]{\rightratio\linewidth}
  \centering
    \includegraphics[width=\graphw]{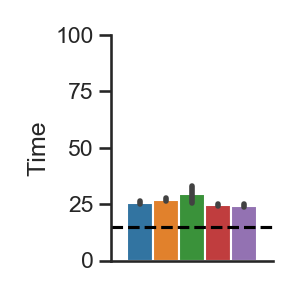}
    \includegraphics[width=\graphw]{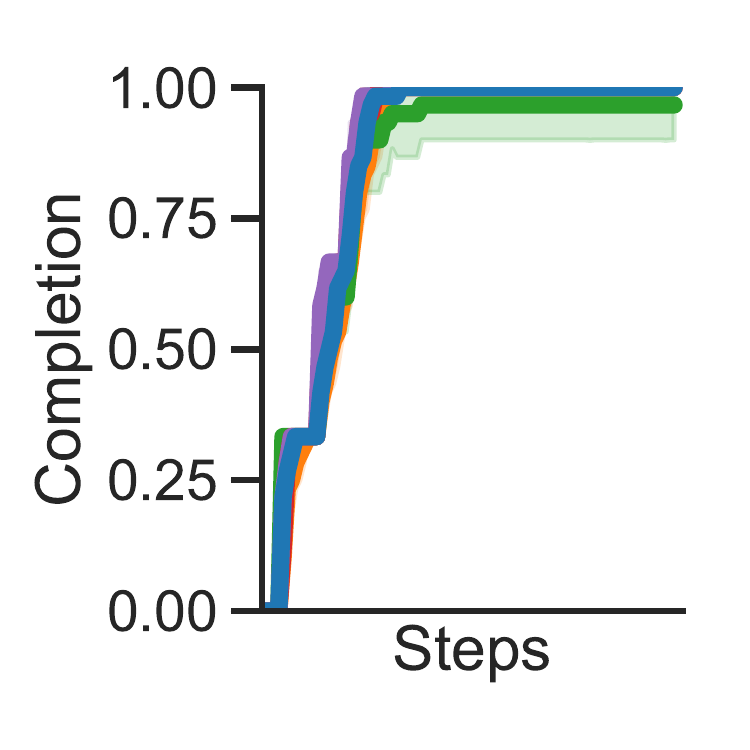}
  \end{subfigure}
  \begin{subfigure}[t]{\rightratio\linewidth}
  \centering
    \includegraphics[width=\graphw]{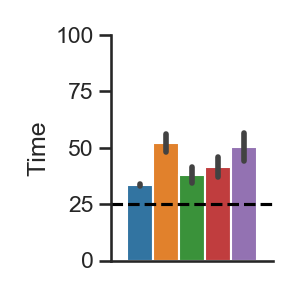}
    \includegraphics[width=\graphw]{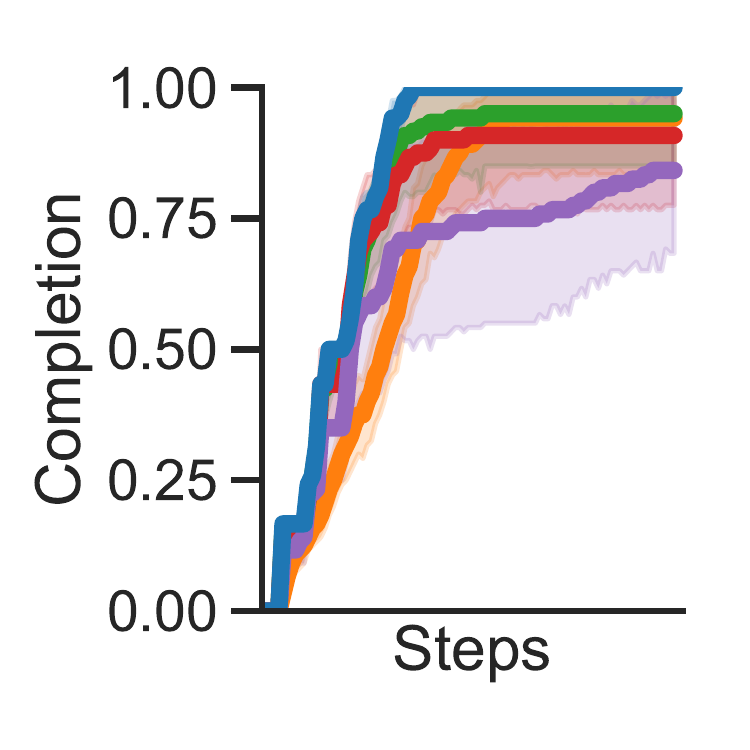}
  \end{subfigure}
  \begin{subfigure}[t]{\rightratio\linewidth}
  \centering
    \includegraphics[width=\graphw]{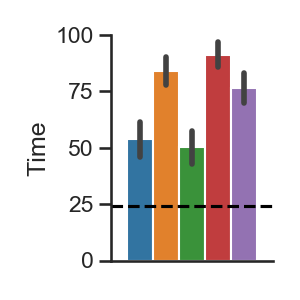}
    \includegraphics[width=\graphw]{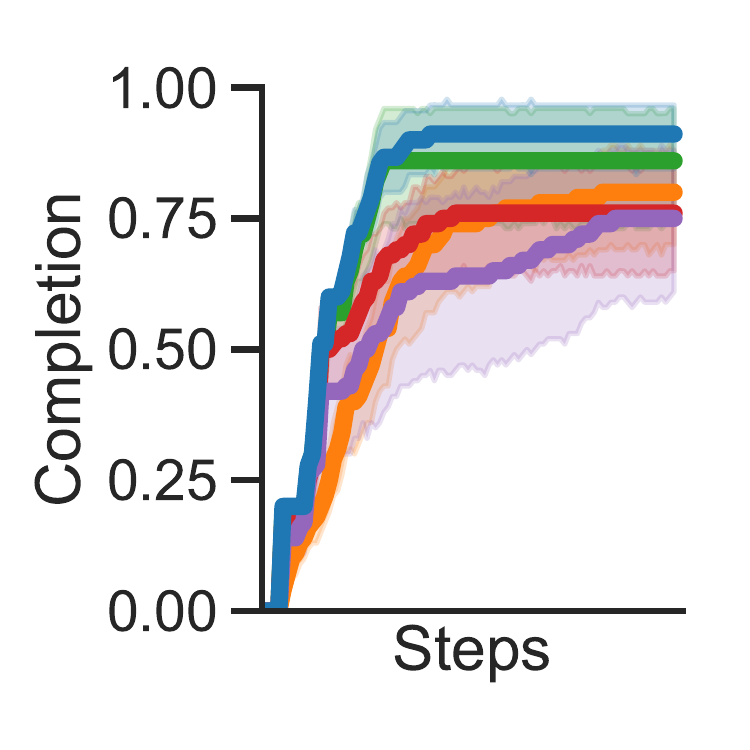}
  \end{subfigure}

  \vspace{\verticalspace}
  \begin{subfigure}[t]{\leftratio\linewidth}
  \centering
    \rotatebox{90}{\scriptsize{\quad\textit{Partial-Divider\phantom{p}}}}\includegraphics[width=\gw]{images/levels/partial.png}
  \end{subfigure}
  \begin{subfigure}[t]{\rightratio\linewidth}
  \centering
    \includegraphics[width=\graphw]{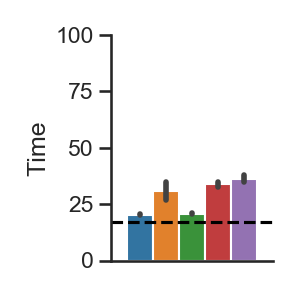}
    \includegraphics[width=\graphw]{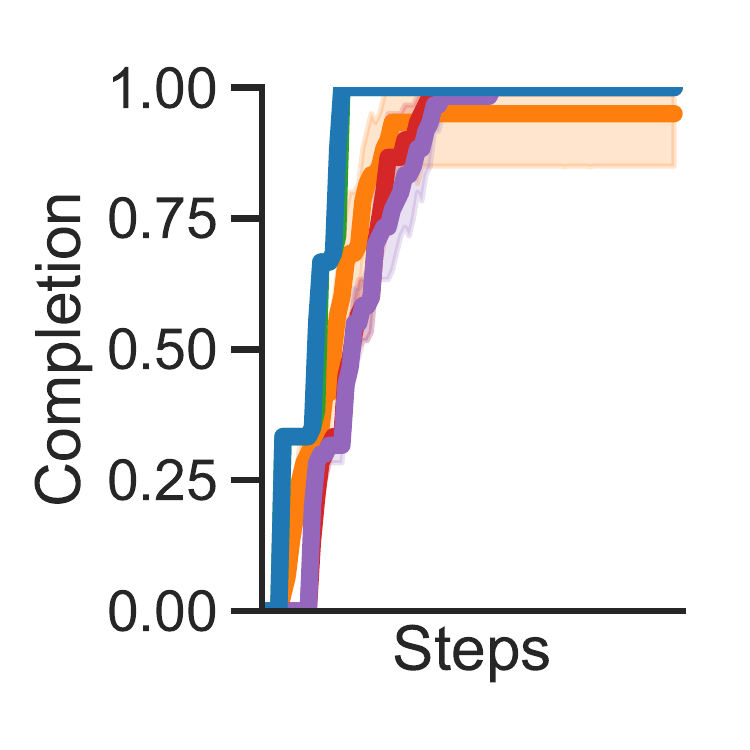}
  \end{subfigure}
  \begin{subfigure}[t]{\rightratio\linewidth}
  \centering
    \includegraphics[width=\graphw]{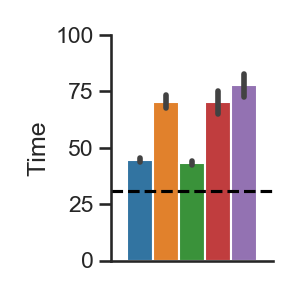}
    \includegraphics[width=\graphw]{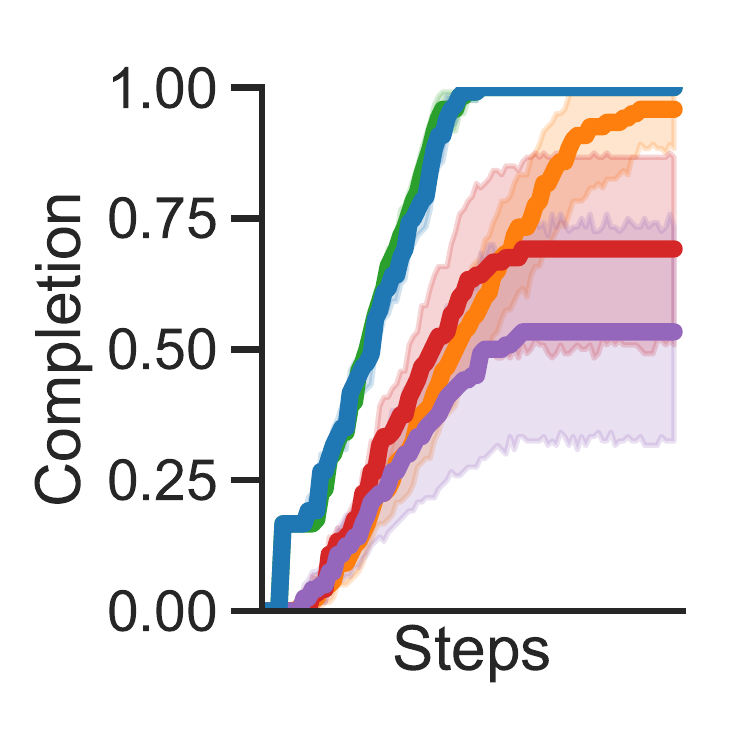}
  \end{subfigure}
  \begin{subfigure}[t]{\rightratio\linewidth}
  \centering
    \includegraphics[width=\graphw]{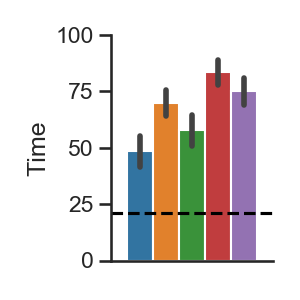}
    \includegraphics[width=\graphw]{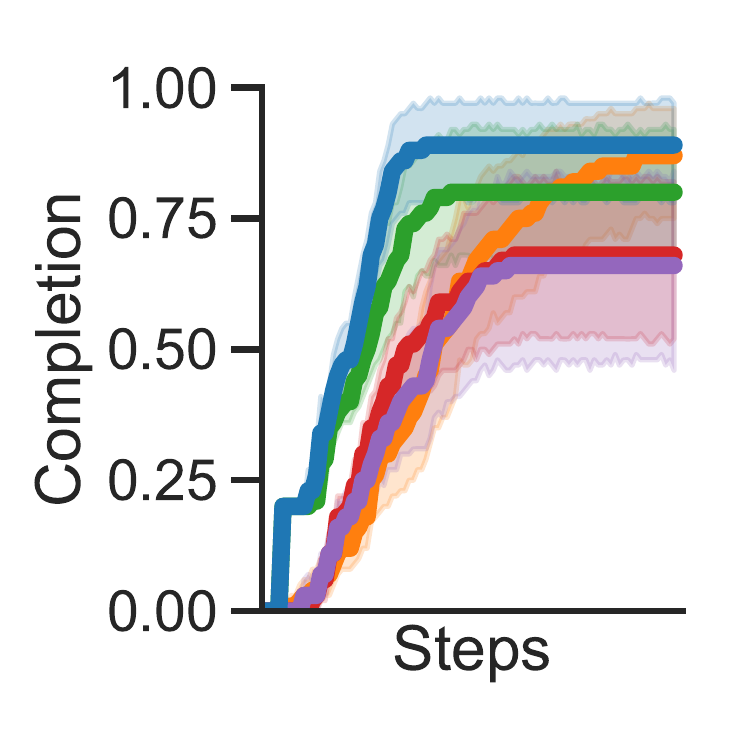}
  \end{subfigure}
  
  \vspace{\verticalspace}
  \begin{subfigure}[t]{\leftratio\linewidth}
  \centering
      \rotatebox{90}{\scriptsize{\quad\textit{Full-Divider\phantom{p}}}}\includegraphics[width=\gw]{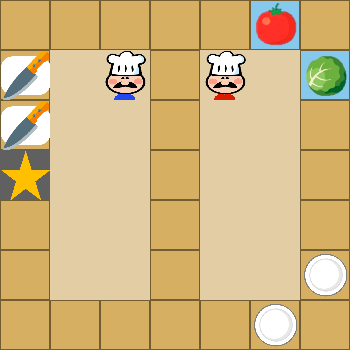}
  \end{subfigure}
  \begin{subfigure}[t]{\rightratio\linewidth}
  \centering
    \includegraphics[width=\graphw]{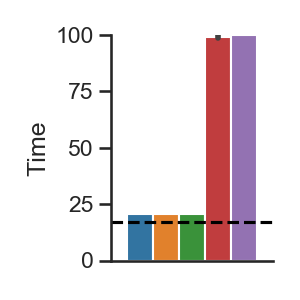}
    \includegraphics[width=\graphw]{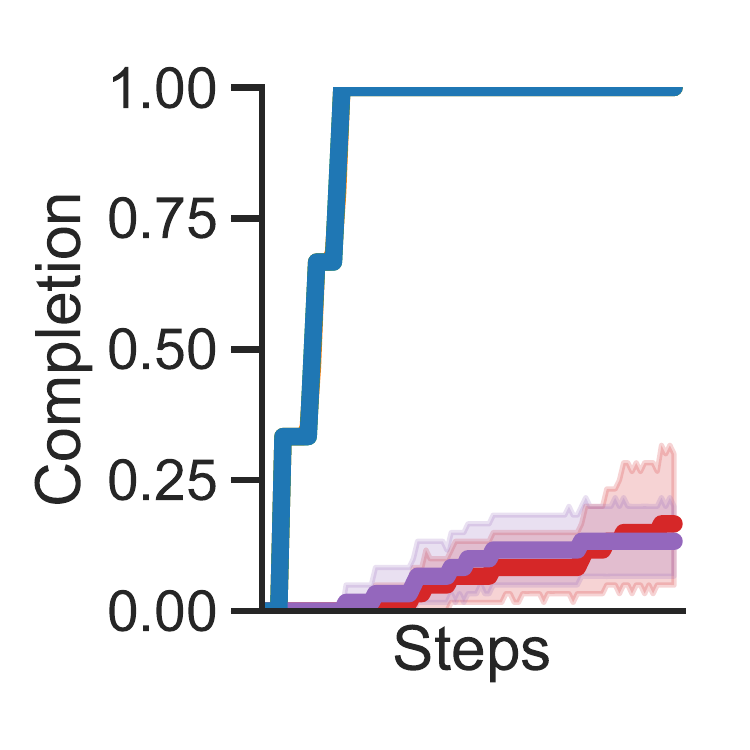}
  \end{subfigure}
  \begin{subfigure}[t]{\rightratio\linewidth}
  \centering
    \includegraphics[width=\graphw]{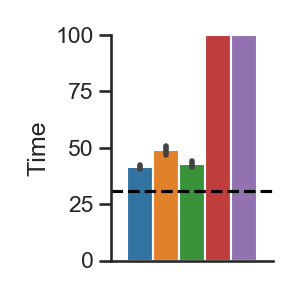}
    \includegraphics[width=\graphw]{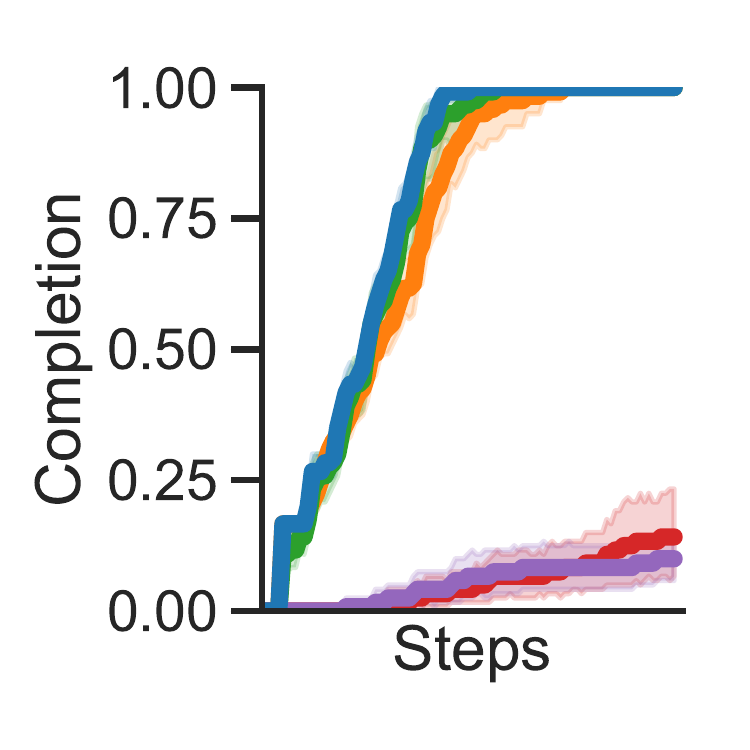}
  \end{subfigure}
  \begin{subfigure}[t]{\rightratio\linewidth}
  \centering
    \includegraphics[width=\graphw]{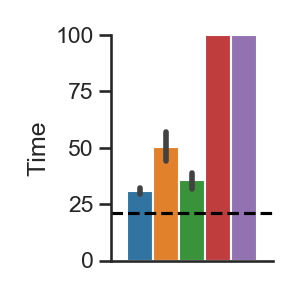}
    \includegraphics[width=\graphw]{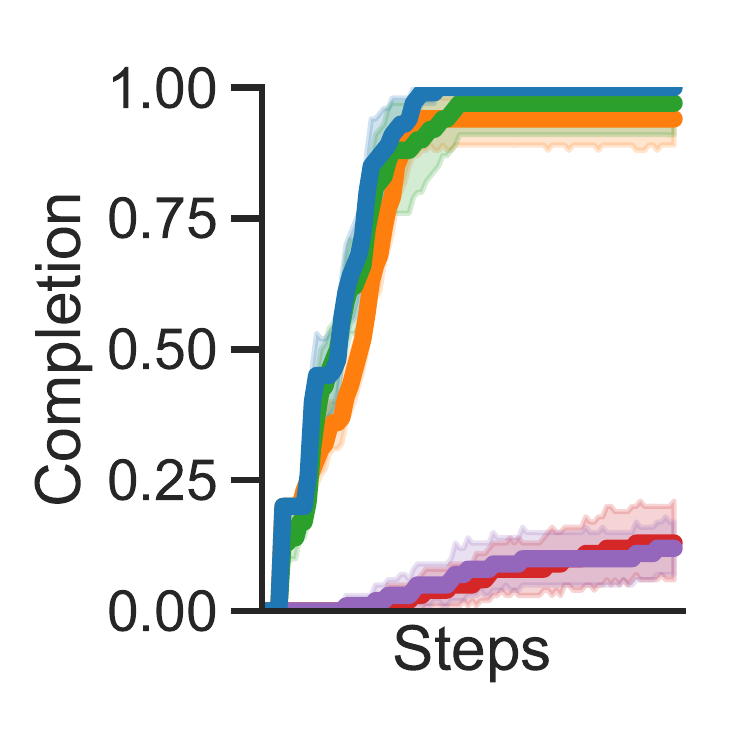}
  \end{subfigure}
\caption{Performance results for each kitchen-recipe composition (lower is better) for two agents in self-play. The row shows the kitchen and the column shows the recipe. Within each composition, the left graph shows the number of time steps needed to complete all sub-tasks. The dashed lines on the left graph represent the optimal performance of a centralized team. The right graph shows the fraction of sub-tasks completed over time.
Bayesian Delegation completes more sub-tasks and does so more quickly compared to baselines.\vspace{-2em}
\label{fig:perf} }
\end{figure}

We evaluate the performance of Bayesian Delegation across three different experimental paradigms. First, we test the performance of each agent type when all agents are the same type with both two and three agents (self-play). Second, we test the performance of each agent type when paired with an agent of a different type (ad-hoc). Finally, we test each model's ability to predict human inferences of sub-task allocation after observing the behavior of other agents (human inferences).  

We compare the performance of Bayesian Delegation (BD) to four alternative baseline agents: Uniform Priors (UP), which starts with uniform probability mass over all valid $ta$ and updates through inverse planning; Fixed Beliefs (FB), which does not update $P(ta)$ in response to the behavior of others; Divide and Conquer (D\&C) \cite{ephrati1994divide}, which sets $P(ta)=0$ if that $ta$ assigns two agents to the same sub-task (this is conceptually similar to Empathy by Fixed Weight Discounting \cite{claes2015effective} because agents cannot share sub-tasks and D\&C discounts sub-tasks most likely to be attended to by other agents based on $P(ta|H)$); Greedy, which selects the sub-task it can complete most quickly without considering the sub-tasks other agents are working on. See Appendix~\ref{appendix:compexperiments} for more details.
All agents take advantage of the sub-task structure because end-to-end optimization of the full recipe using techniques such as DQN \cite{mnih2013playing} and Q-learning \cite{watkins1992q} never succeeded under our computational budget.

To highlight the differences between our model and the alternatives, let us consider an example with two possible sub-tasks ($[\mathcal{T}_1, \mathcal{T}_2]$) and two agents ($[i, j]$). The prior for Bayesian Delegation puts positive probability mass on $\mathbf{ta} = [(i: \mathcal{T}_1, j : \mathcal{T}_2), (i : \mathcal{T}_2, j : \mathcal{T}_1), (i : \mathcal{T}_1, j : \mathcal{T}_1), (i : \mathcal{T}_2, j : \mathcal{T}_2)]$ where $i: \mathcal{T}_1$ means that agent $i$ is assigned to sub-task $\mathcal{T}_1$. The UP agent proposes the same $\mathbf{ta}$, but places uniform probability across all elements, i.e., $P(ta) = \frac{1}{4}$ for all $ta \in \mathbf{ta}$.  FB would  propose the same $\mathbf{ta}$ with the same priors as Bayesian Delegation, but would never update its beliefs. The D\&C agent does not allow for joint sub-tasks, so it would reduce to $\mathbf{ta} = [(i: \mathcal{T}_1, j : \mathcal{T}_2), (i : \mathcal{T}_2, j : \mathcal{T}_1)]$. Lastly, Greedy makes no inferences so each agent $i$ would propose $\mathbf{ta} = [(i : \mathcal{T}_1), (i : \mathcal{T}_2)]$. Note that $j$ does not appear.

In the first two computational experiments, we analyze the results in terms of three key metrics. The two pivotal metrics are the number of time steps to complete the full recipe and the total fraction of sub-tasks completed. We also analyze average number of shuffles, a measure of uncoordinated behavior. A \textit{shuffle} is any action that negates the previous action, such as moving left and then right, or picking an object up and then putting it back down (see Figure~\ref{fig:shuffles-example} for an example). All experiments show the average performance over 20 random seeds. Agents are evaluated in 9 task-environment combinations (3 recipes $\times$ 3 kitchens). See Appendix~\ref{a:repo} for videos of agent behavior. 

\paragraph{Self-play} 
Table~\ref{table:perf} quantifies the performance of all agents aggregated across the 9 environments. Bayesian Delegation outperforms all baselines and completes recipes with less time step and fewer shuffles. The performance gap was even larger with three agents. Most other agents performed worse with three agents than they did with two, while the performance of Bayesian Delegation did not suffer. Figure~\ref{fig:perf} breaks down performance by kitchen and recipe. All five types of agents are comparable when given the recipe \textit{Tomato} in \textit{Open-Divider}, but when faced with more complex situations, Bayesian Delegation outperform the others. For example, without the ability to represent shared sub-tasks, D\&C and Greedy fail in \textit{Full-Divider} because they cannot explicitly coordinate on the same sub-task to pass objects across the counters. Baseline agents were also less capable of low-level coordination resulting in more inefficient shuffles (Figure~\ref{fig:shuffles}). A breakdown of three agent performance is shown in Figure~\ref{fig:perf3}.

Learning about other agents is especially important for more complicated recipes that can be completed in different orders. In particular, FB and Greedy, which do not learn, have trouble with the \textit{Salad} recipe on \textit{Full Divider}. There are two challenges in this composition. One is that the \textit{Salad} recipe can be completed in three different orders: once the tomato and lettuce are chopped, they can be (a) first combined together and then plated, (b) the lettuce can be plated first and then the tomato added or (c) the tomato can be plated first and then the lettuce added. The second challenge is that neither agent can perform all the sub-tasks by themselves, thus they must converge to the same order. Unless the agents that do not learn coordinate by luck, they have no way of recovering. Figure~\ref{fig:permutations} shows the diversity of orderings used across different runs of Bayesian Delegation. Another failure mode for agents lacking learning is that FB and Greedy frequently get stuck in cycles in which both agents are holding objects that must be merged (e.g., a plate and lettuce). They fail to coordinate their actions such that one puts their object down in order for the other to pick it up and merge. Bayesian Delegation can break these symmetries by yielding to others so long as they make net progress towards the completion of one of the sub-tasks. For these reasons, only Bayesian Delegation performs on par (if not more efficiently) with three agents than with two agents. As additional agents join the team, aligning plans becomes even more important in order for agents to avoid performing conflicting or redundant sub-tasks. 

\begin{figure}[bt]
    \centering
    \newcommand{\gw}{55mm}
  \begin{subfigure}[c]{0.34\linewidth}
  \centering
      \includegraphics[width=\gw]{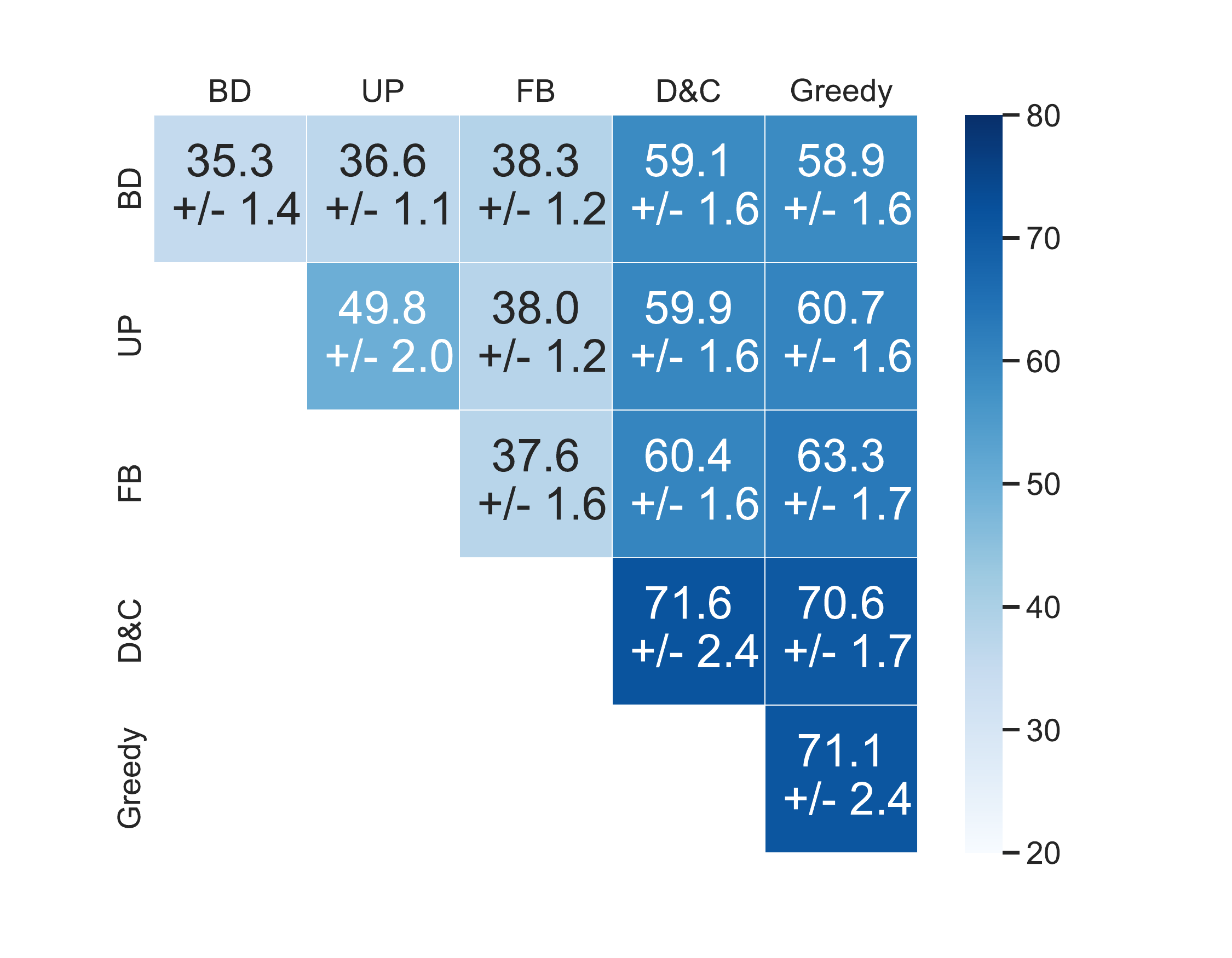}
      \label{fig:adhoc_aggregate}
  \end{subfigure}
  \begin{subfigure}[c]{0.64\linewidth}
  \centering
    \begin{tabular}[c]{lccc}
    \toprule
       & Time Steps  & Completion   &  Shuffles \\
       & ($\downarrow$ better) & ($\uparrow$ better) & ($\downarrow$ better) \\    \midrule
      BD (ours) & \textbf{48.25 $\pm$ 0.75} & \textbf{0.90 $\pm$ 0.01} & \textbf{3.96 $\pm$ 0.33}\\
      UP & 48.84 $\pm$ 0.77 & 0.89 $\pm$ 0.01  & 4.17 $\pm$ 0.34\\
      FB & 50.00 $\pm$ 0.78 & 0.87 $\pm$ 0.01 & 5.11 $\pm$ 0.42\\
      D\&C & 62.49 $\pm$ 0.83  & 0.77 $\pm$ 0.01 & 6.84 $\pm$  0.43 \\
      Greedy & 63.40 $\pm$ 0.84 & 0.76 $\pm$ 0.01 & 6.61 $\pm$ 0.41\\  \bottomrule
    \end{tabular}
    \label{fig:adhoc_table}
  \end{subfigure}
  \vspace{-1em}
    \caption{Ad-hoc performance of different agent pairs in time steps (the lower and lighter, the better).
    (Left) Rows and columns correspond to different agents. Each cell is the average performance of one the row agent playing with the column agent.  
    (Right) Mean performance ($\pm$ SE) of agents when paired with the others.     \label{fig:adhoc}
    }
\end{figure}

\paragraph{Ad-hoc} 
Next, we evaluated the ad-hoc performance of the agents. We show that Bayesian Delegation is a successful ad-hoc collaborator. Each agent was paired with the other agent types. None of the agents had any prior experience with the other agents. Figure~\ref{fig:adhoc} shows the performance of each agent when matched with each other and in aggregate across all recipe-kitchen combinations. Bayesian Delegation performed well even when matched with baselines. When paired with UP, D\&C, and Greedy, the dyad performed better than when UP, D\&C, and Greedy were each paired with their own type. Because Bayesian Delegation can learn in-the-moment, it can overcome some of the ways that these agents get stuck. UP  performs better when paired with Bayesian Delegation or FB compared to self-play, suggesting that as long as one of the agents is initialized with smart priors, it may be enough to compensate for the other's uninformed priors. D\&C and Greedy perform better when paired with Bayesian Delegation, FB, or UP. Crucially, these three agents all represent cooperative plans where both agents cooperate on the same sub-task.  Figure~\ref{fig:adhoc_breakdown} breaks down the ad-hoc performance of each agent pairing by recipe and kitchen. 

\paragraph{Human inferences}
Finally, we evaluated the performance of Bayesian Delegation in predicting human inferences in these same settings. We designed a novel behavioral experiment where human participants first observed a few time steps of two agents behaving, and then made inferences about the sub-tasks each agent was carrying out. We measured the correlation between their judgements and the model predictions.

Participants ($N=45$) were recruited on Amazon Mechanical Turk and shown six scenes of two agents working together in the same scenarios as in the previous experiment (see for the Figure~\ref{fig:traj} for the six scenarios) and asked to make inferences about the agent's intentions as the interaction unfolded. Participants made judgments on a continuous slider [0, 1] with endpoints labeled ``not likely at all'' to ``certainly'' (Figure~\ref{fig:human_expt_example}). Beliefs were normalized into a probability distribution for each subject and then averaged across all subjects. These scenes include a variety of coordinated plans such as instances of clear task allocation (e.g., Figure~\ref{fig:traj1}) and of ambiguous plans where the agent intentions become more clear over time as the interaction continues (e.g., Figure~\ref{fig:traj4}).   These averages were compared to the beliefs formed by our model ($P(ta|H)$) after observing the same trajectory $H$. Each participant made 51 distinct judgments.

\begin{figure}[bt]
    \newcommand{\gw}{28mm}
    \newcommand{\plotr}{0.19}
    \begin{subfigure}[b]{\plotr\columnwidth}
      \centering
        \includegraphics[width=\gw]{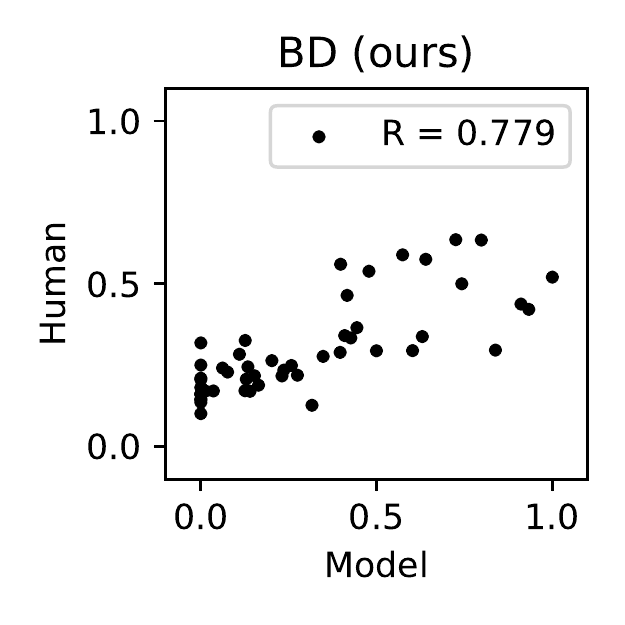}
      \label{fig:correlation_full}
    \end{subfigure}
    \begin{subfigure}[b]{\plotr\columnwidth}
      \centering
        \includegraphics[width=\gw]{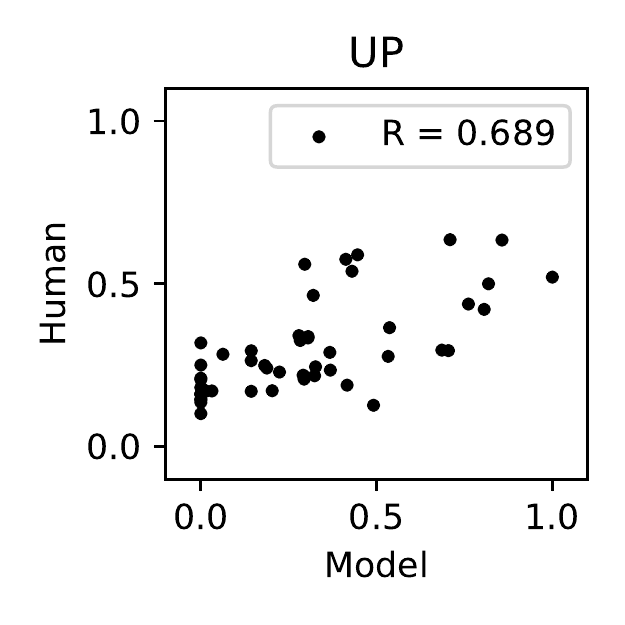}
      \label{fig:correlation_uniform}
    \end{subfigure}
    \begin{subfigure}[b]{\plotr\columnwidth}
      \centering
        \includegraphics[width=\gw]{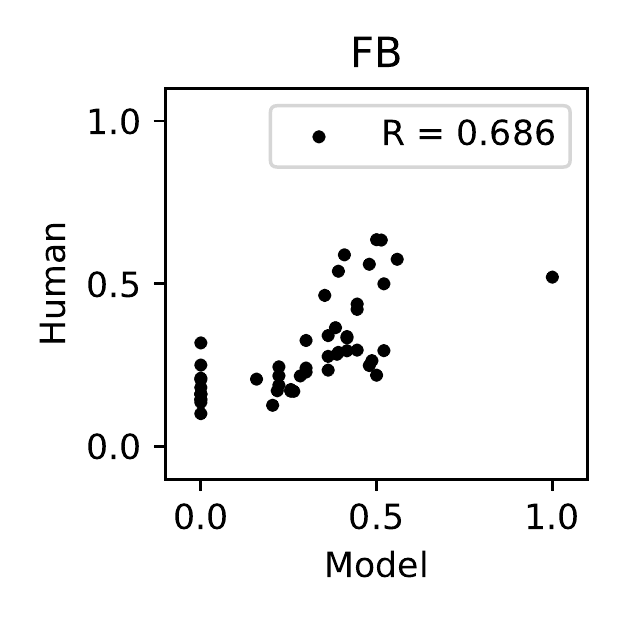}
      \label{fig:correlation_no-bayes}
    \end{subfigure}
    \begin{subfigure}[b]{\plotr\columnwidth}
      \centering
        \includegraphics[width=\gw]{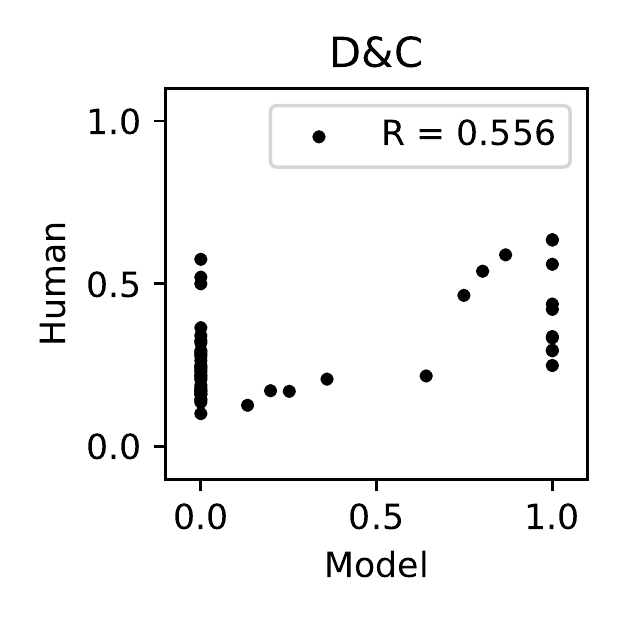}
      \label{fig:correlation_no-jp}
    \end{subfigure}
    \begin{subfigure}[b]{\plotr\columnwidth}
      \centering
        \includegraphics[width=\gw]{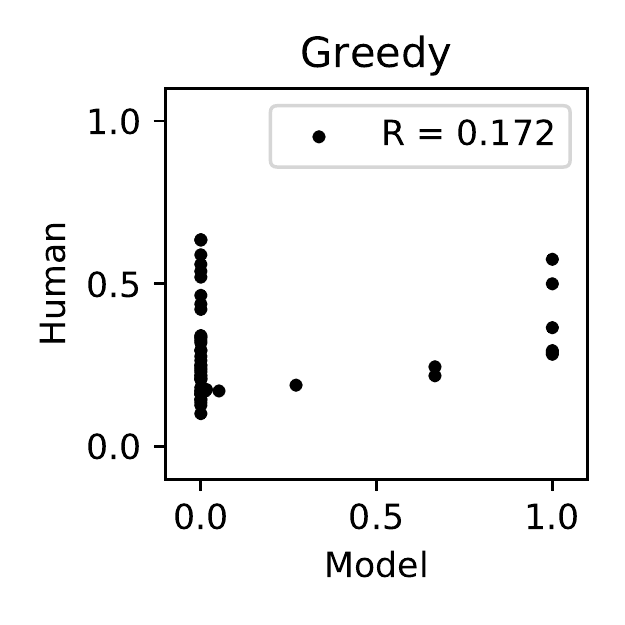}
      \label{fig:correlation_no-jp-no-bayes}
    \end{subfigure}
    \vspace{-2em}
    \caption{Correlation between model and human inferences. Our model (BD, far left) is the most aligned with human judgements compared to the baselines. \vspace{-2em}
    \label{fig:human_correlation}}
\end{figure}

Figure~\ref{fig:traj} shows the inferences made by humans and our model for each scenario at each time step. Figure~\ref{fig:human_correlation} quantifies the overall correspondence of each model with human judgements, and shows that all four baseline models are less aligned with human judgements than Bayesian Delegation.

In particular, priors may help capture people's initial beliefs about high-level plans among agents in a scene, and belief updates may be important for tracking how people's predictions evolve over the course of an interaction. For instance, updates are especially important in Figure~\ref{fig:traj}(d) where the observed actions are initially ambiguous. When the left agent moves to put the tomato on the counter; this may be interpreted as performing either \texttt{Merge(Tomato.chopped, Plate[])} or \texttt{Merge(Lettuce.unchopped, Knife)}, but after only a few time steps, the former emerges as more probable.
Without any training on human Bayesian Delegation captures some of the fine-grained structure of human sub-task inferences. These results suggest that Bayesian Delegation may help us build better models of human theory-of-mind towards the ultimate goal of building machines that can cooperate with people. 

\section{Discussion}
We developed Bayesian Delegation, a new algorithm inspired by and consistent with human theory-of-mind.
Bayesian Delegation enables efficient ad-hoc coordination by rapidly inferring the sub-tasks of others. Agents dynamically align their beliefs about who is doing what and determine when they should help another agent on the same sub-task and when they should work divide and conquer for increased efficiency. It also enables them to complete sub-tasks that neither agent could achieve on its own. Our agents reflect many natural aspects of human theory-of-mind and cooperation \cite{tomasello2014natural}. Indeed, like people, Bayesian Delegation makes predictions about sub-task allocations from only sparse data, and does so in ways consistent with human judgments. 

While Bayesian Delegation reflects progress towards human-like coordination, there are still limitations which we hope to address in future work. One challenge is that when agents jointly plan for a single sub-task, they currently have no way of knowing when they have completed their individual ``part'' of the joint effort. Consider a case where one agent needs to pass both lettuce and tomato across the divider for the other to chop it. After dropping off the lettuce, the first agent should reason that it has fulfilled its role in that joint plan and can move on to next task, i.e., that the rest of the sub-task depends only on the actions of the other agent. 
If agents were able to recognize when their own specific roles in sub-tasks are finished they could look ahead to future sub-tasks that will need to be done even before their preconditions are satisfied.
At some point, as one scales up the number of agents, there can be ``too many cooks'' in the kitchen! Other, less flexible mechanism and representations  will likely play a crucial role in coordinating the behavior of larger groups of agents such as hierarchies, norms, and conventions \cite{young1993evolution, bicchieri2006grammar, lewis1969convention, lerer2019learning}. These representations are essential for building agents that can form longer term collaborations which persist beyond a single short interaction and are capable of partnering with human teams and with each other.

\newpage
\section*{Broader Impact}
We foresee several benefits of our work. Mechanisms that can align an AI agent's beliefs with those of other agents without communication could be valuable for producing assistive technologies that help people in their daily lives. These future technologies could range from virtual assistants on a computer to household robots for the elderly. In many cases, a common communication channel may not be feasible. It is possible that Bayesian Delegation could help align the goals and intentions of agents to those of people and prevent AI systems from acting in ways inimical to humans. Finally, our model makes agent's decision-making process interpretable: to understand why agents take certain actions, we can simply review the beliefs of those agents leading up to that action. This interpretability is not present in many end-to-end trained deep reinforcement learning policies. 

At the highest level, we also think it is worth considering \textit{what} general sub-tasks agents should infer, who gets to decide those sub-tasks, and how to represent sub-tasks that are aligned with human values i.e., $V_{\mathcal{T}}$ where $\mathcal{T}$ is a specific ``human value'' \cite{russell2019human}. Finally, another risk in our work that is left for future efforts is not having communication to resolve conflicts. The ability to communicate meaning and intention could be helpful in resolving such uncertainties, if agents had a common communication channel. The ability to make intentions clear is extremely important if systems like are deployed in sensitive applications like such as surveillance, law enforcement or military applications. 

\begin{ack}
We thank Alex Peysakhovich, Barbara Grosz, DJ Strouse, Leslie Kaelbling, Micah Carroll, and Natasha Jaques for insightful ideas and comments. This work was funded by the Harvard Data Science Initiate, Harvard CRCS, Templeton World Charity Foundation, The Future of Life Institute, DARPA Ground Truth, and The Center for Brains, Minds and Machines (NSF STC award CCF-1231216). 
\end{ack}

\bibliographystyle{ACM-Reference-Format}  
\bibliography{paper.bib}


\begin{thebibliography}{00}


\ifx \showCODEN    \undefined \def \showCODEN     #1{\unskip}     \fi
\ifx \showDOI      \undefined \def \showDOI       #1{#1}\fi
\ifx \showISBNx    \undefined \def \showISBNx     #1{\unskip}     \fi
\ifx \showISBNxiii \undefined \def \showISBNxiii  #1{\unskip}     \fi
\ifx \showISSN     \undefined \def \showISSN      #1{\unskip}     \fi
\ifx \showLCCN     \undefined \def \showLCCN      #1{\unskip}     \fi
\ifx \shownote     \undefined \def \shownote      #1{#1}          \fi
\ifx \showarticletitle \undefined \def \showarticletitle #1{#1}   \fi
\ifx \showURL      \undefined \def \showURL       {\relax}        \fi
\providecommand\bibfield[2]{#2}
\providecommand\bibinfo[2]{#2}
\providecommand\natexlab[1]{#1}
\providecommand\showeprint[2][]{arXiv:#2}

\bibitem[\protect\citeauthoryear{Amato, Konidaris, Kaelbling, and How}{Amato
  et~al\mbox{.}}{2019}]%
        {amato2019modeling}
\bibfield{author}{\bibinfo{person}{Christopher Amato}, \bibinfo{person}{George
  Konidaris}, \bibinfo{person}{Leslie~Pack Kaelbling}, {and}
  \bibinfo{person}{Jonathan~P. How}.} \bibinfo{year}{2019}\natexlab{}.
\newblock \showarticletitle{Modeling and Planning with Macro-Actions in
  Decentralized POMDPs}.
\newblock \bibinfo{journal}{{\em Journal of Artificial Intelligence
  Research\/}}  \bibinfo{volume}{64} (\bibinfo{year}{2019}),
  \bibinfo{pages}{817--859}.
\newblock


\bibitem[\protect\citeauthoryear{Baker, Jara-Ettinger, Saxe, and
  Tenenbaum}{Baker et~al\mbox{.}}{2017}]%
        {baker2017rational}
\bibfield{author}{\bibinfo{person}{Chris~L Baker}, \bibinfo{person}{Julian
  Jara-Ettinger}, \bibinfo{person}{Rebecca Saxe}, {and}
  \bibinfo{person}{Joshua~B Tenenbaum}.} \bibinfo{year}{2017}\natexlab{}.
\newblock \showarticletitle{Rational quantitative attribution of beliefs,
  desires and percepts in human mentalizing}.
\newblock \bibinfo{journal}{{\em Nature Human Behaviour\/}}
  \bibinfo{volume}{1} (\bibinfo{year}{2017}), \bibinfo{pages}{0064}.
\newblock


\bibitem[\protect\citeauthoryear{Barrett, Stone, and Kraus}{Barrett
  et~al\mbox{.}}{2011}]%
        {barrett2011empirical}
\bibfield{author}{\bibinfo{person}{Samuel Barrett}, \bibinfo{person}{Peter
  Stone}, {and} \bibinfo{person}{Sarit Kraus}.}
  \bibinfo{year}{2011}\natexlab{}.
\newblock \showarticletitle{Empirical evaluation of ad hoc teamwork in the
  pursuit domain}. In \bibinfo{booktitle}{{\em The 10th International
  Conference on Autonomous Agents and Multiagent Systems-Volume 2}}.
  International Foundation for Autonomous Agents and Multiagent Systems,
  \bibinfo{pages}{567--574}.
\newblock


\bibitem[\protect\citeauthoryear{Barrett, Stone, Kraus, and Rosenfeld}{Barrett
  et~al\mbox{.}}{2012}]%
        {barrett2012learning}
\bibfield{author}{\bibinfo{person}{Samuel Barrett}, \bibinfo{person}{Peter
  Stone}, \bibinfo{person}{Sarit Kraus}, {and} \bibinfo{person}{Avi
  Rosenfeld}.} \bibinfo{year}{2012}\natexlab{}.
\newblock \showarticletitle{Learning teammate models for ad hoc teamwork}. In
  \bibinfo{booktitle}{{\em AAMAS Adaptive Learning Agents (ALA) Workshop}}.
  \bibinfo{pages}{57--63}.
\newblock


\bibitem[\protect\citeauthoryear{Bicchieri}{Bicchieri}{2006}]%
        {bicchieri2006grammar}
\bibfield{author}{\bibinfo{person}{Cristina Bicchieri}.}
  \bibinfo{year}{2006}\natexlab{}.
\newblock \bibinfo{booktitle}{{\em The grammar of society: The nature and
  dynamics of social norms}}.
\newblock \bibinfo{publisher}{Cambridge University Press}.
\newblock


\bibitem[\protect\citeauthoryear{Boutilier}{Boutilier}{1996}]%
        {boutilier1996planning}
\bibfield{author}{\bibinfo{person}{Craig Boutilier}.}
  \bibinfo{year}{1996}\natexlab{}.
\newblock \showarticletitle{Planning, learning and coordination in multiagent
  decision processes}. In \bibinfo{booktitle}{{\em Proceedings of the 6th
  conference on Theoretical aspects of rationality and knowledge}}. Morgan
  Kaufmann Publishers Inc., \bibinfo{pages}{195--210}.
\newblock


\bibitem[\protect\citeauthoryear{Brunet, Choi, and How}{Brunet
  et~al\mbox{.}}{2008}]%
        {brunet2008consensus}
\bibfield{author}{\bibinfo{person}{Luc Brunet}, \bibinfo{person}{Han-Lim Choi},
  {and} \bibinfo{person}{Jonathan How}.} \bibinfo{year}{2008}\natexlab{}.
\newblock \showarticletitle{Consensus-based auction approaches for
  decentralized task assignment}. In \bibinfo{booktitle}{{\em AIAA guidance,
  navigation and control conference and exhibit}}. \bibinfo{pages}{6839}.
\newblock


\bibitem[\protect\citeauthoryear{Carroll, Shah, Ho, Griffiths, Seshia, Abbeel,
  and Dragan}{Carroll et~al\mbox{.}}{2019}]%
        {carroll2019on}
\bibfield{author}{\bibinfo{person}{Micah Carroll}, \bibinfo{person}{Rohin
  Shah}, \bibinfo{person}{Mark Ho}, \bibinfo{person}{Thomas Griffiths},
  \bibinfo{person}{Sanjit Seshia}, \bibinfo{person}{Pieter Abbeel}, {and}
  \bibinfo{person}{Anca Dragan}.} \bibinfo{year}{2019}\natexlab{}.
\newblock \showarticletitle{On the Utility of Learning about Humans for
  Human-AI Coordination}. In \bibinfo{booktitle}{{\em Advances in Neural
  Information Processing Systems}}.
\newblock


\bibitem[\protect\citeauthoryear{Chalkiadakis and Boutilier}{Chalkiadakis and
  Boutilier}{2003}]%
        {chalkiadakis2003coordination}
\bibfield{author}{\bibinfo{person}{Georgios Chalkiadakis} {and}
  \bibinfo{person}{Craig Boutilier}.} \bibinfo{year}{2003}\natexlab{}.
\newblock \showarticletitle{Coordination in multiagent reinforcement learning:
  A bayesian approach}. In \bibinfo{booktitle}{{\em Proceedings of the second
  international joint conference on Autonomous agents and multiagent systems}}.
  \bibinfo{pages}{709--716}.
\newblock


\bibitem[\protect\citeauthoryear{Claes, Oliehoek, Baier, and Tuyls}{Claes
  et~al\mbox{.}}{2017}]%
        {claes2017decentralised}
\bibfield{author}{\bibinfo{person}{Daniel Claes}, \bibinfo{person}{Frans
  Oliehoek}, \bibinfo{person}{Hendrik Baier}, {and} \bibinfo{person}{Karl
  Tuyls}.} \bibinfo{year}{2017}\natexlab{}.
\newblock \showarticletitle{Decentralised online planning for multi-robot
  warehouse commissioning}. In \bibinfo{booktitle}{{\em Proceedings of the 16th
  Conference on Autonomous Agents and MultiAgent Systems}}. International
  Foundation for Autonomous Agents and Multiagent Systems,
  \bibinfo{pages}{492--500}.
\newblock


\bibitem[\protect\citeauthoryear{Claes, Robbel, Oliehoek, Tuyls, Hennes, and
  Van~der Hoek}{Claes et~al\mbox{.}}{2015}]%
        {claes2015effective}
\bibfield{author}{\bibinfo{person}{Daniel Claes}, \bibinfo{person}{Philipp
  Robbel}, \bibinfo{person}{Frans~A Oliehoek}, \bibinfo{person}{Karl Tuyls},
  \bibinfo{person}{Daniel Hennes}, {and} \bibinfo{person}{Wiebe Van~der Hoek}.}
  \bibinfo{year}{2015}\natexlab{}.
\newblock \showarticletitle{Effective approximations for multi-robot
  coordination in spatially distributed tasks}. In \bibinfo{booktitle}{{\em
  Proceedings of the 2015 International Conference on Autonomous Agents and
  Multiagent Systems}}. International Foundation for Autonomous Agents and
  Multiagent Systems, \bibinfo{pages}{881--890}.
\newblock


\bibitem[\protect\citeauthoryear{Cohen and Levesque}{Cohen and
  Levesque}{1991}]%
        {cohen1991teamwork}
\bibfield{author}{\bibinfo{person}{Philip~R Cohen} {and}
  \bibinfo{person}{Hector~J Levesque}.} \bibinfo{year}{1991}\natexlab{}.
\newblock \showarticletitle{Teamwork}.
\newblock \bibinfo{journal}{{\em No{\^u}s\/}} \bibinfo{volume}{25},
  \bibinfo{number}{4} (\bibinfo{year}{1991}), \bibinfo{pages}{487--512}.
\newblock


\bibitem[\protect\citeauthoryear{Cox and Durfee}{Cox and Durfee}{2004}]%
        {cox2004efficient}
\bibfield{author}{\bibinfo{person}{Jeffrey~S Cox} {and}
  \bibinfo{person}{Edmund~H Durfee}.} \bibinfo{year}{2004}\natexlab{}.
\newblock \showarticletitle{Efficient mechanisms for multiagent plan merging}.
  In \bibinfo{booktitle}{{\em Proceedings of the Third International Joint
  Conference on Autonomous Agents and Multiagent Systems, 2004. AAMAS 2004.}}
  IEEE, \bibinfo{pages}{1342--1343}.
\newblock


\bibitem[\protect\citeauthoryear{Cox and Durfee}{Cox and Durfee}{2005}]%
        {cox2005efficient}
\bibfield{author}{\bibinfo{person}{Jeffrey~S Cox} {and}
  \bibinfo{person}{Edmund~H Durfee}.} \bibinfo{year}{2005}\natexlab{}.
\newblock \showarticletitle{An efficient algorithm for multiagent plan
  coordination}. In \bibinfo{booktitle}{{\em Proceedings of the fourth
  international joint conference on Autonomous agents and multiagent systems}}.
  \bibinfo{pages}{828--835}.
\newblock


\bibitem[\protect\citeauthoryear{Diuk, Cohen, and Littman}{Diuk
  et~al\mbox{.}}{2008}]%
        {diuk2008object}
\bibfield{author}{\bibinfo{person}{Carlos Diuk}, \bibinfo{person}{Andre Cohen},
  {and} \bibinfo{person}{Michael~L Littman}.} \bibinfo{year}{2008}\natexlab{}.
\newblock \showarticletitle{An object-oriented representation for efficient
  reinforcement learning}. In \bibinfo{booktitle}{{\em Proceedings of the 25th
  international conference on Machine learning}}. ACM,
  \bibinfo{pages}{240--247}.
\newblock


\bibitem[\protect\citeauthoryear{Ephrati and Rosenschein}{Ephrati and
  Rosenschein}{1994}]%
        {ephrati1994divide}
\bibfield{author}{\bibinfo{person}{Eithan Ephrati} {and}
  \bibinfo{person}{Jeffrey~S Rosenschein}.} \bibinfo{year}{1994}\natexlab{}.
\newblock \showarticletitle{Divide and conquer in multi-agent planning}. In
  \bibinfo{booktitle}{{\em AAAI}}, Vol.~\bibinfo{volume}{1}.
  \bibinfo{pages}{80}.
\newblock


\bibitem[\protect\citeauthoryear{Fikes and Nilsson}{Fikes and Nilsson}{1971}]%
        {strips}
\bibfield{author}{\bibinfo{person}{Richard~E. Fikes} {and}
  \bibinfo{person}{Nils~J. Nilsson}.} \bibinfo{year}{1971}\natexlab{}.
\newblock \showarticletitle{Strips: A new approach to the application of
  theorem proving to problem solving}.
\newblock \bibinfo{journal}{{\em Artificial Intelligence\/}}
  \bibinfo{volume}{2}, \bibinfo{number}{3} (\bibinfo{year}{1971}),
  \bibinfo{pages}{189 -- 208}.
\newblock
\showISSN{0004-3702}
\showDOI{%
\url{https://doi.org/10.1016/0004-3702(71)90010-5}}


\bibitem[\protect\citeauthoryear{{Ghost Town Games}}{{Ghost Town
  Games}}{2016}]%
        {overcooked2016}
\bibfield{author}{\bibinfo{person}{{Ghost Town Games}}.}
  \bibinfo{year}{2016}\natexlab{}.
\newblock \showarticletitle{Overcooked}.
\newblock  (\bibinfo{year}{2016}).
\newblock


\bibitem[\protect\citeauthoryear{Grosz and Kraus}{Grosz and Kraus}{1996}]%
        {grosz1996collaborative}
\bibfield{author}{\bibinfo{person}{Barbara~J Grosz} {and}
  \bibinfo{person}{Sarit Kraus}.} \bibinfo{year}{1996}\natexlab{}.
\newblock \showarticletitle{Collaborative plans for complex group action}.
\newblock \bibinfo{journal}{{\em Artificial Intelligence\/}}
  \bibinfo{volume}{86}, \bibinfo{number}{2} (\bibinfo{year}{1996}),
  \bibinfo{pages}{269--357}.
\newblock


\bibitem[\protect\citeauthoryear{Henrich}{Henrich}{2015}]%
        {henrich2015secret}
\bibfield{author}{\bibinfo{person}{Joseph Henrich}.}
  \bibinfo{year}{2015}\natexlab{}.
\newblock \bibinfo{booktitle}{{\em The secret of our success: how culture is
  driving human evolution, domesticating our species, and making us smarter}}.
\newblock \bibinfo{publisher}{Princeton University Press}.
\newblock


\bibitem[\protect\citeauthoryear{Kleiman-Weiner, Ho, Austerweil, Littman, and
  Tenenbaum}{Kleiman-Weiner et~al\mbox{.}}{2016}]%
        {kleiman2016coordinate}
\bibfield{author}{\bibinfo{person}{Max Kleiman-Weiner}, \bibinfo{person}{Mark~K
  Ho}, \bibinfo{person}{Joseph~L Austerweil}, \bibinfo{person}{Michael~L
  Littman}, {and} \bibinfo{person}{Joshua~B Tenenbaum}.}
  \bibinfo{year}{2016}\natexlab{}.
\newblock \showarticletitle{Coordinate to cooperate or compete: abstract goals
  and joint intentions in social interaction}. In \bibinfo{booktitle}{{\em
  Proceedings of the 38th Annual Conference of the Cognitive Science Society}}.
\newblock


\bibitem[\protect\citeauthoryear{Lerer and Peysakhovich}{Lerer and
  Peysakhovich}{2019}]%
        {lerer2019learning}
\bibfield{author}{\bibinfo{person}{Adam Lerer} {and} \bibinfo{person}{Alexander
  Peysakhovich}.} \bibinfo{year}{2019}\natexlab{}.
\newblock \showarticletitle{Learning Existing Social Conventions via
  Observationally Augmented Self-Play}. In \bibinfo{booktitle}{{\em Proceedings
  of the 2019 AAAI/ACM Conference on AI, Ethics, and Society}}. ACM,
  \bibinfo{pages}{107--114}.
\newblock


\bibitem[\protect\citeauthoryear{Lewis}{Lewis}{1969}]%
        {lewis1969convention}
\bibfield{author}{\bibinfo{person}{David Lewis}.}
  \bibinfo{year}{1969}\natexlab{}.
\newblock \bibinfo{booktitle}{{\em Convention: A philosophical study}}.
\newblock \bibinfo{publisher}{John Wiley \& Sons}.
\newblock


\bibitem[\protect\citeauthoryear{McIntire, Nunes, and Gini}{McIntire
  et~al\mbox{.}}{2016}]%
        {mcintire2016iterated}
\bibfield{author}{\bibinfo{person}{Mitchell McIntire}, \bibinfo{person}{Ernesto
  Nunes}, {and} \bibinfo{person}{Maria Gini}.} \bibinfo{year}{2016}\natexlab{}.
\newblock \showarticletitle{Iterated multi-robot auctions for
  precedence-constrained task scheduling}. In \bibinfo{booktitle}{{\em
  Proceedings of the 2016 International Conference on Autonomous Agents \&
  Multiagent Systems}}. \bibinfo{pages}{1078--1086}.
\newblock


\bibitem[\protect\citeauthoryear{McMahan, Likhachev, and Gordon}{McMahan
  et~al\mbox{.}}{2005}]%
        {mcmahan2005bounded}
\bibfield{author}{\bibinfo{person}{H~Brendan McMahan}, \bibinfo{person}{Maxim
  Likhachev}, {and} \bibinfo{person}{Geoffrey~J Gordon}.}
  \bibinfo{year}{2005}\natexlab{}.
\newblock \showarticletitle{Bounded real-time dynamic programming: RTDP with
  monotone upper bounds and performance guarantees}. In
  \bibinfo{booktitle}{{\em Proceedings of the 22nd international conference on
  Machine learning}}. ACM, \bibinfo{pages}{569--576}.
\newblock


\bibitem[\protect\citeauthoryear{Melo and Sardinha}{Melo and Sardinha}{2016}]%
        {melo2016ad}
\bibfield{author}{\bibinfo{person}{Francisco~S Melo} {and}
  \bibinfo{person}{Alberto Sardinha}.} \bibinfo{year}{2016}\natexlab{}.
\newblock \showarticletitle{Ad hoc teamwork by learning teammates’ task}.
\newblock \bibinfo{journal}{{\em Autonomous Agents and Multi-Agent Systems\/}}
  \bibinfo{volume}{30}, \bibinfo{number}{2} (\bibinfo{year}{2016}),
  \bibinfo{pages}{175--219}.
\newblock


\bibitem[\protect\citeauthoryear{Misyak, Melkonyan, Zeitoun, and Chater}{Misyak
  et~al\mbox{.}}{2014}]%
        {misyak2014unwritten}
\bibfield{author}{\bibinfo{person}{Jennifer~B Misyak}, \bibinfo{person}{Tigran
  Melkonyan}, \bibinfo{person}{Hossam Zeitoun}, {and} \bibinfo{person}{Nick
  Chater}.} \bibinfo{year}{2014}\natexlab{}.
\newblock \showarticletitle{Unwritten rules: virtual bargaining underpins
  social interaction, culture, and society}.
\newblock \bibinfo{journal}{{\em Trends in cognitive sciences\/}}
  (\bibinfo{year}{2014}).
\newblock


\bibitem[\protect\citeauthoryear{Mnih, Kavukcuoglu, Silver, Graves, Antonoglou,
  Wierstra, and Riedmiller}{Mnih et~al\mbox{.}}{2013}]%
        {mnih2013playing}
\bibfield{author}{\bibinfo{person}{Volodymyr Mnih}, \bibinfo{person}{Koray
  Kavukcuoglu}, \bibinfo{person}{David Silver}, \bibinfo{person}{Alex Graves},
  \bibinfo{person}{Ioannis Antonoglou}, \bibinfo{person}{Daan Wierstra}, {and}
  \bibinfo{person}{Martin Riedmiller}.} \bibinfo{year}{2013}\natexlab{}.
\newblock \showarticletitle{Playing atari with deep reinforcement learning}.
\newblock \bibinfo{journal}{{\em arXiv preprint arXiv:1312.5602\/}}
  (\bibinfo{year}{2013}).
\newblock


\bibitem[\protect\citeauthoryear{Nagel}{Nagel}{1986}]%
        {nagel1986view}
\bibfield{author}{\bibinfo{person}{Thomas Nagel}.}
  \bibinfo{year}{1986}\natexlab{}.
\newblock \bibinfo{booktitle}{{\em The view from nowhere}}.
\newblock \bibinfo{publisher}{Oxford University Press}.
\newblock


\bibitem[\protect\citeauthoryear{Nakahashi, Baker, and Tenenbaum}{Nakahashi
  et~al\mbox{.}}{2016}]%
        {nakahashi2016modeling}
\bibfield{author}{\bibinfo{person}{Ryo Nakahashi}, \bibinfo{person}{Chris~L
  Baker}, {and} \bibinfo{person}{Joshua~B Tenenbaum}.}
  \bibinfo{year}{2016}\natexlab{}.
\newblock \showarticletitle{Modeling Human Understanding of Complex Intentional
  Action with a Bayesian Nonparametric Subgoal Model.}. In
  \bibinfo{booktitle}{{\em AAAI}}. \bibinfo{pages}{3754--3760}.
\newblock


\bibitem[\protect\citeauthoryear{Ram{\i}rez and Geffner}{Ram{\i}rez and
  Geffner}{2011}]%
        {ramirez2011goal}
\bibfield{author}{\bibinfo{person}{Miquel Ram{\i}rez} {and}
  \bibinfo{person}{Hector Geffner}.} \bibinfo{year}{2011}\natexlab{}.
\newblock \showarticletitle{Goal recognition over POMDPs: Inferring the
  intention of a POMDP agent}. In \bibinfo{booktitle}{{\em IJCAI}}. IJCAI/AAAI,
  \bibinfo{pages}{2009--2014}.
\newblock


\bibitem[\protect\citeauthoryear{Russell}{Russell}{2019}]%
        {russell2019human}
\bibfield{author}{\bibinfo{person}{Stuart Russell}.}
  \bibinfo{year}{2019}\natexlab{}.
\newblock \bibinfo{booktitle}{{\em Human compatible: Artificial intelligence
  and the problem of control}}.
\newblock \bibinfo{publisher}{Penguin}.
\newblock


\bibitem[\protect\citeauthoryear{Shum, Kleiman-Weiner, Littman, and
  Tenenbaum}{Shum et~al\mbox{.}}{2019}]%
        {shum2019theory}
\bibfield{author}{\bibinfo{person}{Michael Shum}, \bibinfo{person}{Max
  Kleiman-Weiner}, \bibinfo{person}{Michael~L Littman}, {and}
  \bibinfo{person}{Joshua~B Tenenbaum}.} \bibinfo{year}{2019}\natexlab{}.
\newblock \showarticletitle{Theory of Minds: Understanding Behavior in Groups
  Through Inverse Planning}. In \bibinfo{booktitle}{{\em Proceedings of the
  Thirty-Third AAAI Conference on Artificial Intelligence (AAAI-19)}}.
\newblock


\bibitem[\protect\citeauthoryear{Song, Wang, Lukasiewicz, Xu, and Xu}{Song
  et~al\mbox{.}}{2019}]%
        {song2019diversity}
\bibfield{author}{\bibinfo{person}{Yuhang Song}, \bibinfo{person}{Jianyi Wang},
  \bibinfo{person}{Thomas Lukasiewicz}, \bibinfo{person}{Zhenghua Xu}, {and}
  \bibinfo{person}{Mai Xu}.} \bibinfo{year}{2019}\natexlab{}.
\newblock \showarticletitle{Diversity-driven extensible hierarchical
  reinforcement learning}. In \bibinfo{booktitle}{{\em Proceedings of the AAAI
  Conference on Artificial Intelligence}}, Vol.~\bibinfo{volume}{33}.
  \bibinfo{pages}{4992--4999}.
\newblock


\bibitem[\protect\citeauthoryear{Stone, Kaminka, Kraus, and Rosenschein}{Stone
  et~al\mbox{.}}{2010}]%
        {stone2010ad}
\bibfield{author}{\bibinfo{person}{Peter Stone}, \bibinfo{person}{Gal~A
  Kaminka}, \bibinfo{person}{Sarit Kraus}, {and} \bibinfo{person}{Jeffrey~S
  Rosenschein}.} \bibinfo{year}{2010}\natexlab{}.
\newblock \showarticletitle{Ad hoc autonomous agent teams: Collaboration
  without pre-coordination}. In \bibinfo{booktitle}{{\em Twenty-Fourth AAAI
  Conference on Artificial Intelligence}}.
\newblock


\bibitem[\protect\citeauthoryear{Tambe}{Tambe}{1997}]%
        {tambe1997towards}
\bibfield{author}{\bibinfo{person}{Milind Tambe}.}
  \bibinfo{year}{1997}\natexlab{}.
\newblock \showarticletitle{Towards flexible teamwork}.
\newblock \bibinfo{journal}{{\em Journal of artificial intelligence
  research\/}}  \bibinfo{volume}{7} (\bibinfo{year}{1997}),
  \bibinfo{pages}{83--124}.
\newblock


\bibitem[\protect\citeauthoryear{Tomasello}{Tomasello}{2014}]%
        {tomasello2014natural}
\bibfield{author}{\bibinfo{person}{Michael Tomasello}.}
  \bibinfo{year}{2014}\natexlab{}.
\newblock \bibinfo{booktitle}{{\em A natural history of human thinking}}.
\newblock \bibinfo{publisher}{Harvard University Press}.
\newblock


\bibitem[\protect\citeauthoryear{Tomasello, Carpenter, Call, Behne, and
  Moll}{Tomasello et~al\mbox{.}}{2005}]%
        {tomasello2005understanding}
\bibfield{author}{\bibinfo{person}{Michael Tomasello}, \bibinfo{person}{Malinda
  Carpenter}, \bibinfo{person}{Josep Call}, \bibinfo{person}{Tanya Behne},
  {and} \bibinfo{person}{Henrike Moll}.} \bibinfo{year}{2005}\natexlab{}.
\newblock \showarticletitle{Understanding and sharing intentions: The origins
  of cultural cognition}.
\newblock \bibinfo{journal}{{\em Behavioral and brain sciences\/}}
  \bibinfo{volume}{28}, \bibinfo{number}{05} (\bibinfo{year}{2005}),
  \bibinfo{pages}{675--691}.
\newblock


\bibitem[\protect\citeauthoryear{Watkins and Dayan}{Watkins and Dayan}{1992}]%
        {watkins1992q}
\bibfield{author}{\bibinfo{person}{Christopher~JCH Watkins} {and}
  \bibinfo{person}{Peter Dayan}.} \bibinfo{year}{1992}\natexlab{}.
\newblock \showarticletitle{Q-learning}.
\newblock \bibinfo{journal}{{\em Machine learning\/}} \bibinfo{volume}{8},
  \bibinfo{number}{3-4} (\bibinfo{year}{1992}), \bibinfo{pages}{279--292}.
\newblock


\bibitem[\protect\citeauthoryear{Wright and Leyton-Brown}{Wright and
  Leyton-Brown}{2010}]%
        {wright2010beyond}
\bibfield{author}{\bibinfo{person}{James~R Wright} {and} \bibinfo{person}{Kevin
  Leyton-Brown}.} \bibinfo{year}{2010}\natexlab{}.
\newblock \showarticletitle{Beyond Equilibrium: Predicting Human Behavior in
  Normal-Form Games.}. In \bibinfo{booktitle}{{\em AAAI}}.
\newblock


\bibitem[\protect\citeauthoryear{Young}{Young}{1993}]%
        {young1993evolution}
\bibfield{author}{\bibinfo{person}{H~Peyton Young}.}
  \bibinfo{year}{1993}\natexlab{}.
\newblock \showarticletitle{The evolution of conventions}.
\newblock \bibinfo{journal}{{\em Econometrica: Journal of the Econometric
  Society\/}} (\bibinfo{year}{1993}), \bibinfo{pages}{57--84}.
\newblock


\end{thebibliography}

\newpage 
\appendix

\section{Analysis of Agent Behavior}
\label{sec:agent behavior}

Videos of the agents can be viewed at \href{https://manycooks.github.io/}{\texttt{https://manycooks.github.io/}}. This page summarizes the levels (see Figure~\ref{fig:env_all}) and recipes (see Figure~\ref{fig:recipe}) and includes videos of two and three agents in a variety of environments. It also includes a link to a closed version of the behavioral experiment.

\begin{figure}[ph]
  \newcommand{\gw}{50mm} 
  \newcommand{\nw}{.4}
  \newcommand{\lw}{.08}

  \footnotesize
    \centering
    \begin{subfigure}[b]{\nw\linewidth}
      \centering
      \textbf{Agent 1}
    \end{subfigure}
    \begin{subfigure}[b]{\nw\linewidth}
      \centering
      \textbf{Agent 2}
    \end{subfigure}

    \texttt{Merge(Lettuce.unchopped, Knife)}
    
    \begin{subfigure}[c]{\nw\linewidth}
      \centering
      \includegraphics[width=\gw]{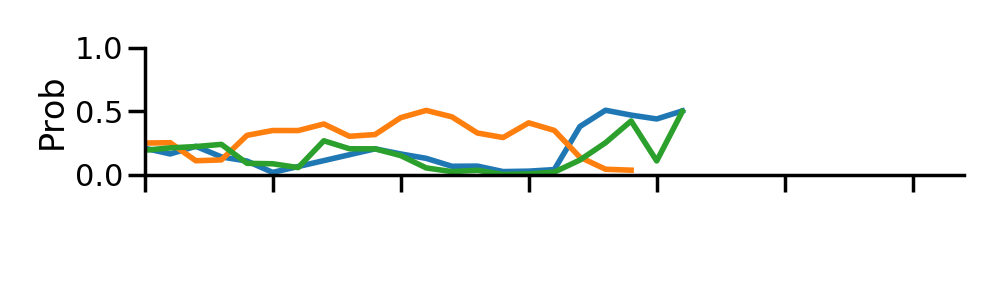}
    \end{subfigure}
    \begin{subfigure}[t]{\lw\linewidth}
        \centering
        \includegraphics[width=12mm]{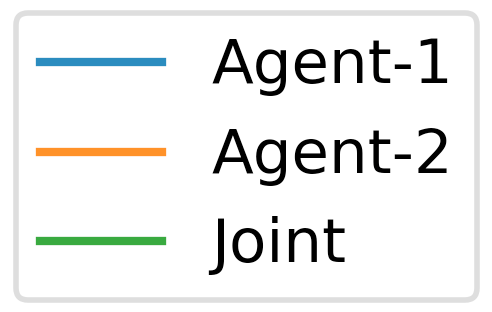}
    \end{subfigure}
    \begin{subfigure}[c]{\nw\linewidth}
      \centering
      \includegraphics[width=\gw]{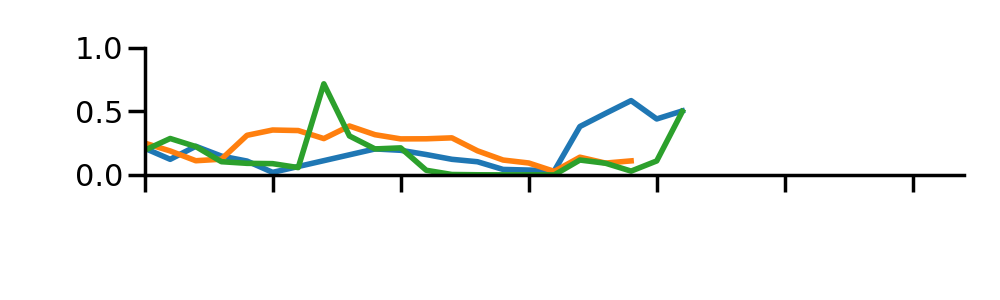}
    \end{subfigure}

    \vspace{-1em}
    \texttt{Merge(Tomato.unchopped, Knife)}  
    
    \begin{subfigure}[t]{\nw\linewidth}
      \centering
      \includegraphics[width=\gw]{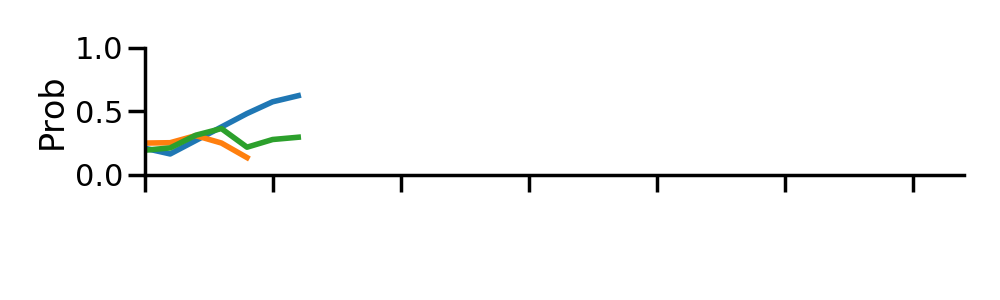}
    \end{subfigure}
    \hspace{\lw\linewidth}%
    \begin{subfigure}[t]{\nw\linewidth}
      \centering
      \includegraphics[width=\gw]{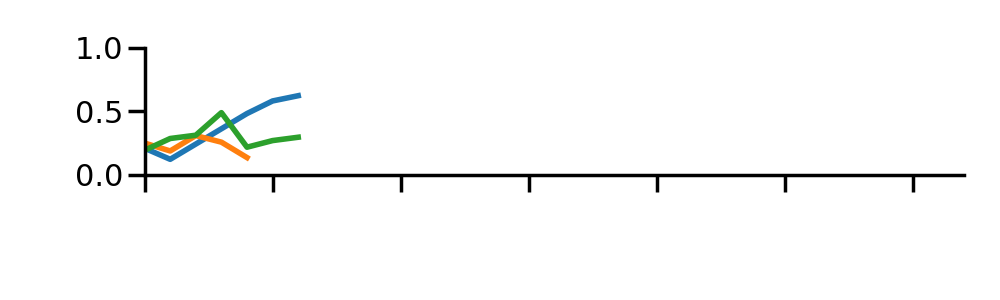}
    \end{subfigure}

    \vspace{-1em}
    \texttt{Merge(Tomato.chopped, Plate[])}  
    
    \begin{subfigure}[t]{\nw\linewidth}
      \centering
      \includegraphics[width=\gw]{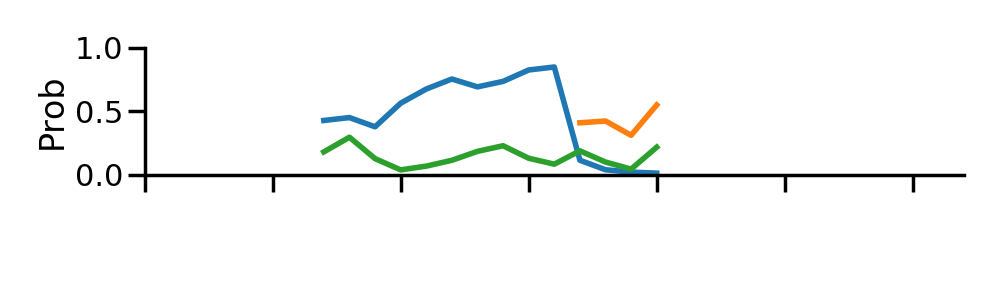}
    \end{subfigure}
    \hspace{\lw\linewidth}%
    \begin{subfigure}[t]{\nw\linewidth}
      \centering
      \includegraphics[width=\gw]{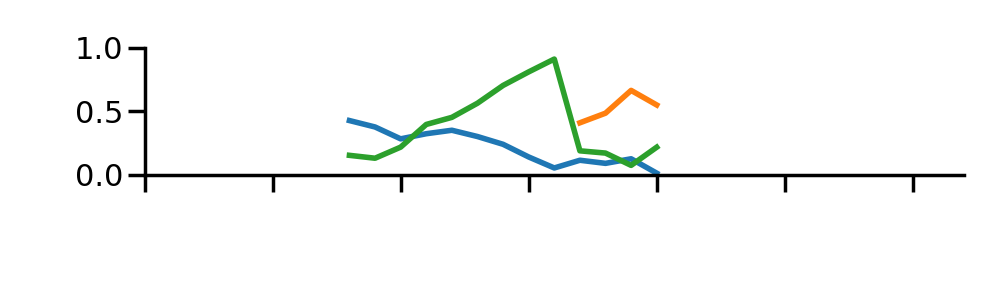}
    \end{subfigure}

    \vspace{-1em}
    \texttt{Merge(Lettuce.chopped, Plate[Tomato])}  
    
    \begin{subfigure}[t]{\nw\linewidth}
      \centering
      \includegraphics[width=\gw]{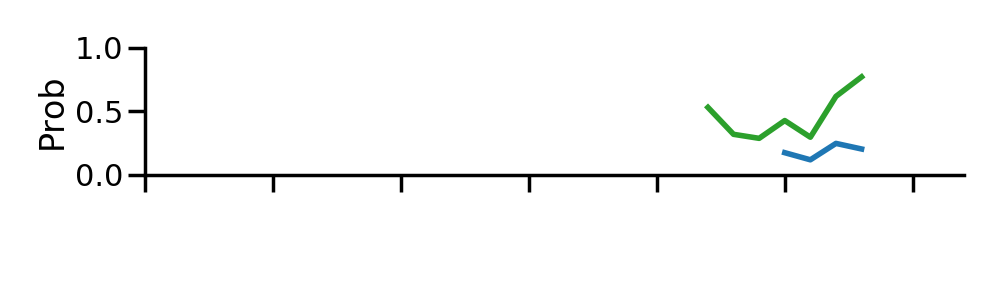}
    \end{subfigure}
    \hspace{\lw\linewidth}%
    \begin{subfigure}[t]{\nw\linewidth}
      \centering
      \includegraphics[width=\gw]{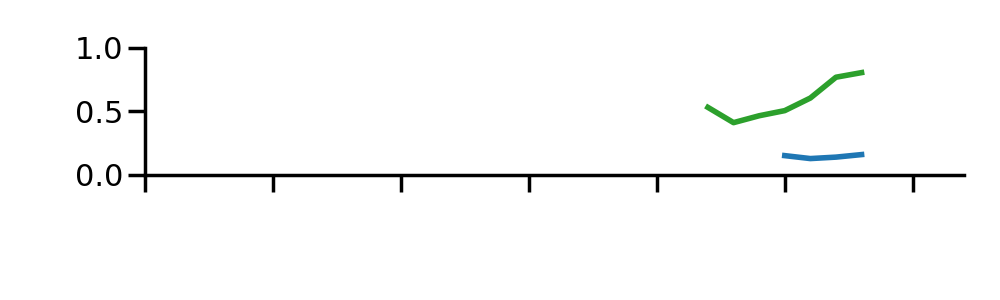}
    \end{subfigure}

    \vspace{-1em}
    \texttt{Merge(Lettuce.chopped, Plate[])}  
    
    \begin{subfigure}[t]{\nw\linewidth}
      \centering
      \includegraphics[width=\gw]{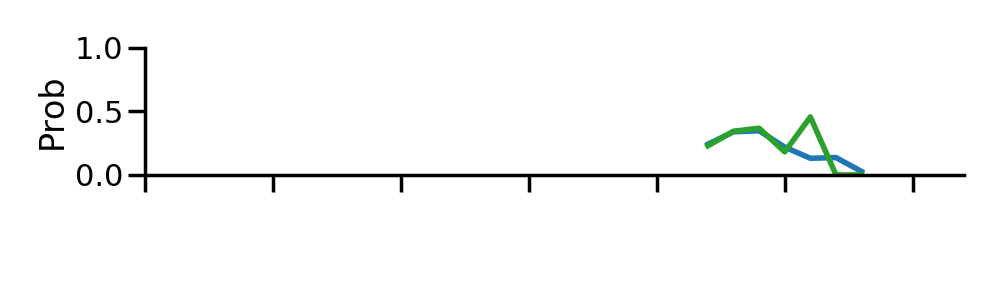}
    \end{subfigure}
    \hspace{\lw\linewidth}%
    \begin{subfigure}[t]{\nw\linewidth}
      \centering
      \includegraphics[width=\gw]{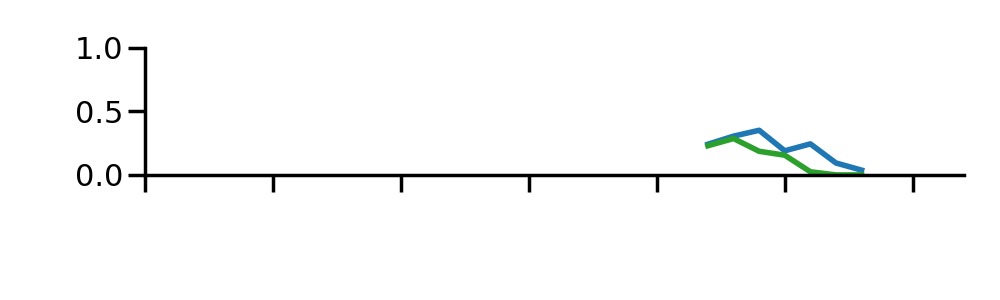}
    \end{subfigure}    

    \vspace{-1em}
    \texttt{Merge(Plate[Tomato.chopped, Lettuce.chopped], Delivery[])}  
    
    \begin{subfigure}[t]{\nw\linewidth}
      \centering
      \includegraphics[width=\gw]{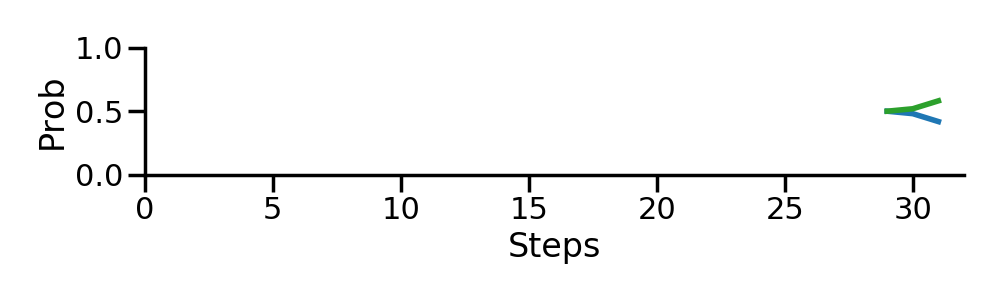}
    \end{subfigure}
    \hspace{\lw\linewidth}%
    \begin{subfigure}[t]{\nw\linewidth}
      \centering
      \includegraphics[width=\gw]{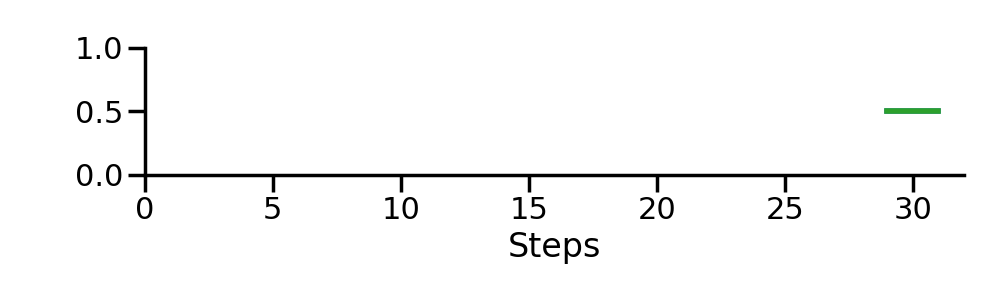} 
    \end{subfigure}      
    
    \caption{Dynamics of the belief state, P($ta$) for each agent during Bayesian delegation with the \textit{Salad} recipe on \textit{Partial-Divider} (Figure~\ref{fig:env}). During the first 7 time steps, only the \texttt{Merge(Lettuce.unchopped, Knife)} and \texttt{Merge(Tomato.unchopped, Knife)} sub-tasks are nonzero because their preconditions are met. These beliefs show alignment across the ordering of sub-tasks as well as within each sub-task. \textit{Salad} can be completed in three different ways (see Figure~\ref{fig:permutations}), yet both agents eventually drop \texttt{Merge(Lettuce.unchopped, Plate[])} in favor of \texttt{Merge(Tomato.unchopped, Plate[])} followed by \texttt{Merge(Lettuce.chopped, Plate[Tomato])}. Agents' beliefs also converge over the course of each specific sub-task. For instance, while both agents are at first uncertain about who should be delegated to \texttt{Merge(Lettuce.unchopped, Knife)}, they eventually align to the same relative ordering. This alignment continues, even though there is never any communication or prior agreement on what sub-task each agent should be doing or when.
    \label{fig:beliefs}
    }
\end{figure}

\begin{figure}[tbh]
  \newcommand{\graphw}{30mm}
  \centering
  \begin{subfigure}[b]{0.25\linewidth}
  \centering
      \includegraphics[width=28mm]{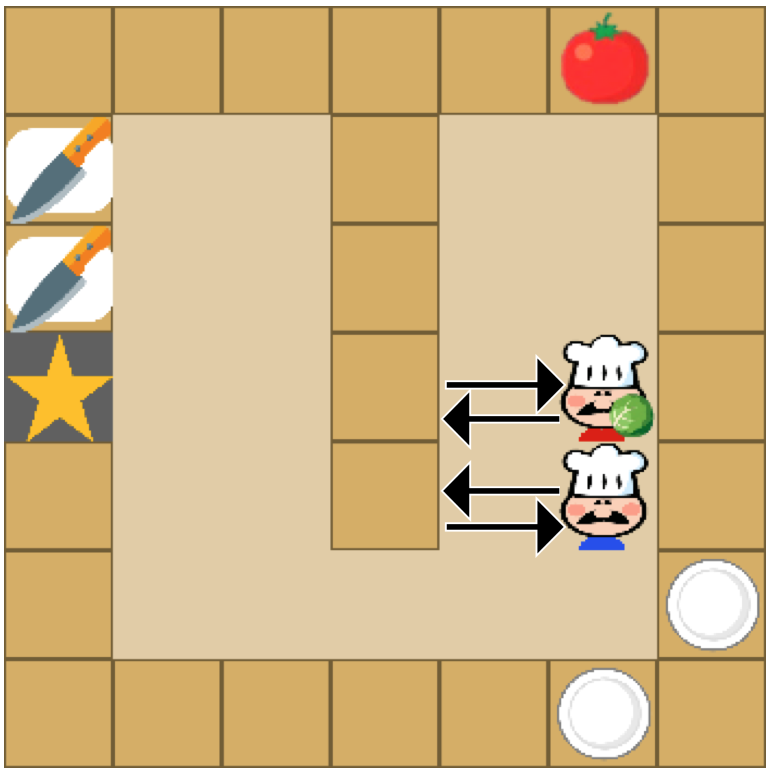}
      \vspace{1mm}
      \caption{Example ``shuffle.''\label{fig:shuffles-example}}
  \end{subfigure}
  \begin{subfigure}[b]{0.15\linewidth}
      \centering
      \includegraphics[width=22mm]{images/legend.png}
      \vspace{8mm}
  \end{subfigure}
  \begin{subfigure}[b]{0.25\linewidth}
    \centering
    \includegraphics[width=\graphw]{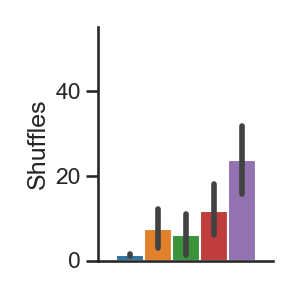}
    \caption{\textit{Open-Divider}.}
  \end{subfigure}
  \begin{subfigure}[b]{0.2\linewidth}
    \centering
    \includegraphics[width=\graphw]{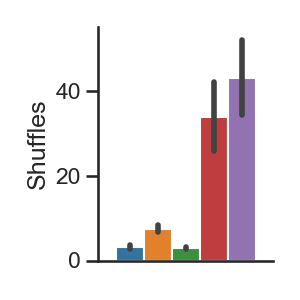}
    \caption{\textit{Partial-Divider}.}
  \end{subfigure}
   
    \caption{Shuffles observed for recipe \textit{Tomato+Lettuce}. (a) Example of a shuffle, where both agents simultaneously move back and forth from left to right, over and over again. This coordination failure prevents them from passing each other. Note that they are not colliding. Average number of shuffles by each agent in the (b) \textit{Open-Divider} and (c) \textit{Partial-Divider} environments. Error bars show the standard error of the mean. Bayesian Delegation and Joint Planning help prevent shuffles, leading to better coordinated behavior. 
    \label{fig:shuffles}}
\end{figure}

\begin{SCfigure}[50][hbt]
  \newcommand{\gw}{50mm}
  \begin{subfigure}[t]{.37\linewidth}
    \includegraphics[width=\gw]{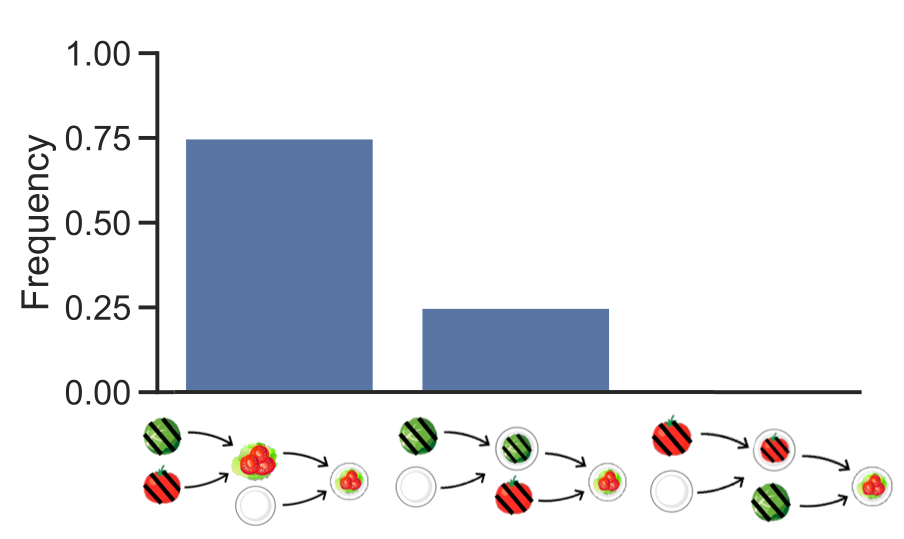}
    \caption{\textit{Open-Divider}.}
  \end{subfigure}
  \begin{subfigure}[t]{.37\linewidth}
    \includegraphics[width=\gw]{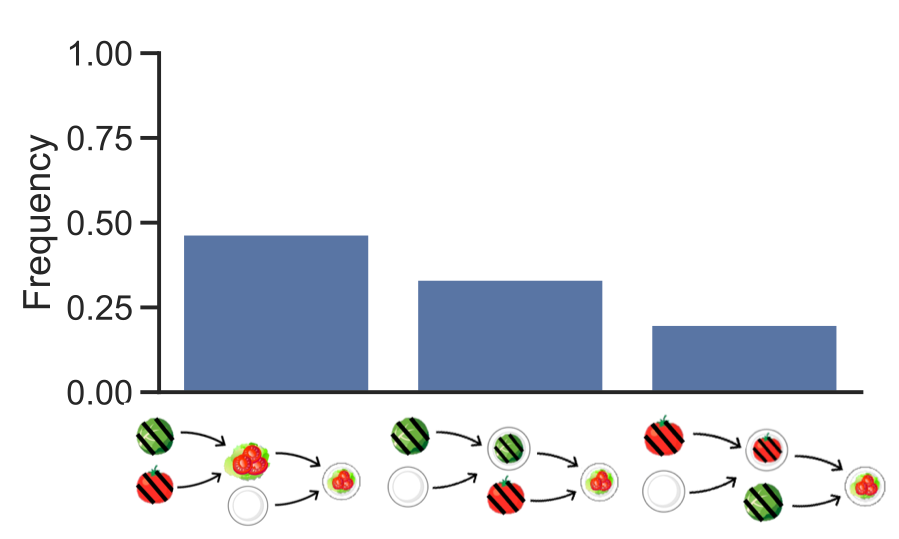}
    \caption{\textit{Partial-Divider}.}
  \end{subfigure}
    \caption{Frequency that three orderings of \textit{Salad} are completed by our model agents, in (a) \textit{Open-Divider} and (b) \textit{Partial-Divider}. \vspace{2em}    \label{fig:permutations}}
\end{SCfigure}

\begin{figure}[tbh]
  \centering
  \newcommand{\rw}{26mm}
  \newcommand{\lw}{16mm}
  \newcommand{\gw}{20mm}
  \newcommand{\bw}{20mm}
  \newcommand{\graphw}{18mm}
  \newcommand{\verticalspace}{1mm}
  \newcommand{\leftratio}{0.17}
  \newcommand{\rightratio}{0.27}

  \begin{subfigure}[t]{\leftratio\linewidth}
  \centering
    \includegraphics[width=\lw]{images/legend.png}
  \end{subfigure}
  \begin{subfigure}[t]{\rightratio\linewidth}
  \centering
    \includegraphics[width=\rw]{images/recipes/SimpleTomato}
  \end{subfigure}
  \begin{subfigure}[t]{\rightratio\linewidth}
  \centering
    \includegraphics[width=\rw]{images/recipes/TL}
  \end{subfigure}
  \begin{subfigure}[t]{\rightratio\linewidth}
  \centering
    \includegraphics[width=\rw]{images/recipes/Salad}
  \end{subfigure}

  \begin{subfigure}[t]{\leftratio\linewidth}
  \centering
      \includegraphics[width=\gw]{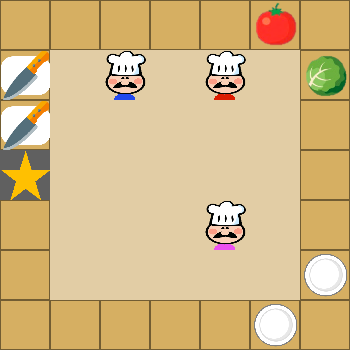}
  \end{subfigure}
  \begin{subfigure}[t]{\rightratio\linewidth}
  \centering
    \includegraphics[width=\graphw]{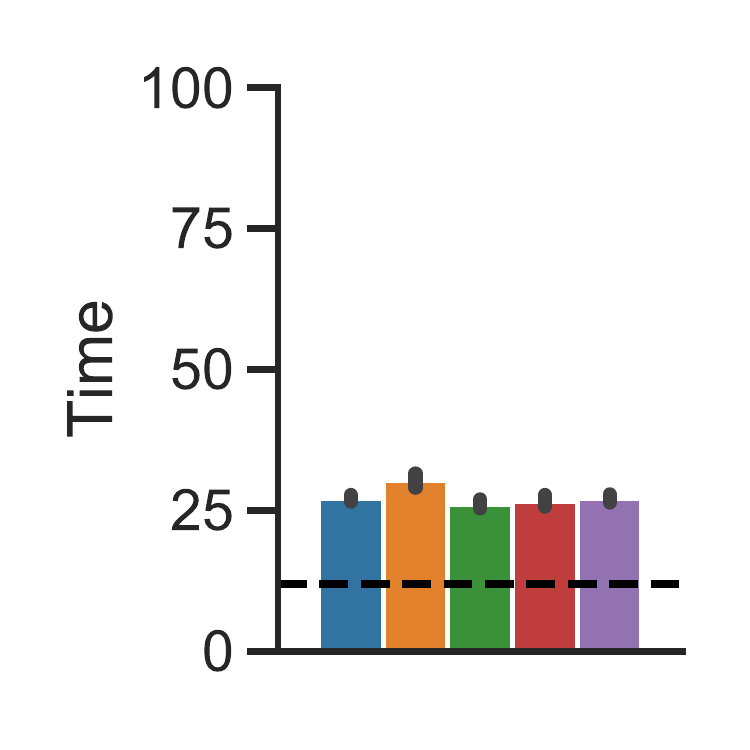}
    \includegraphics[width=\graphw]{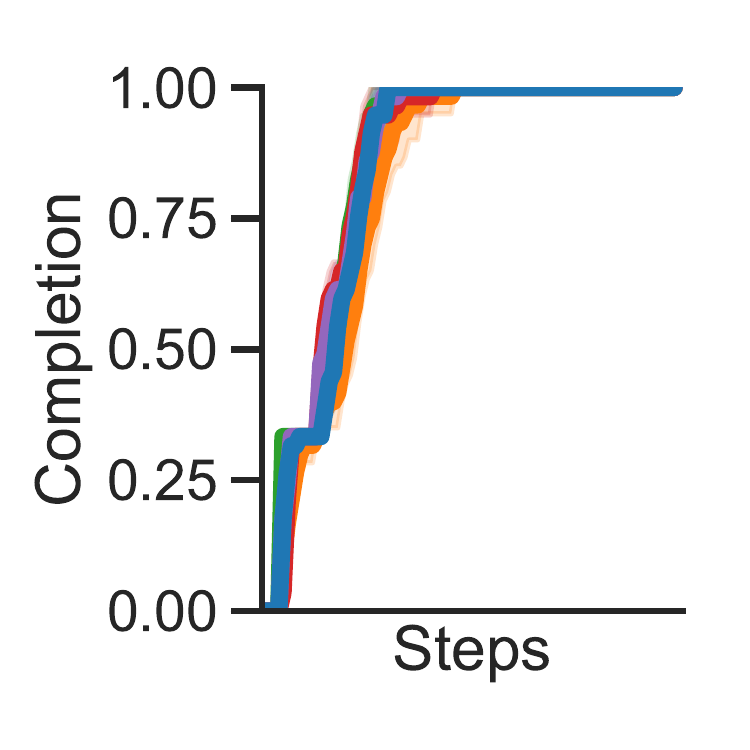}
  \end{subfigure}
  \begin{subfigure}[t]{\rightratio\linewidth}
  \centering
    \includegraphics[width=\graphw]{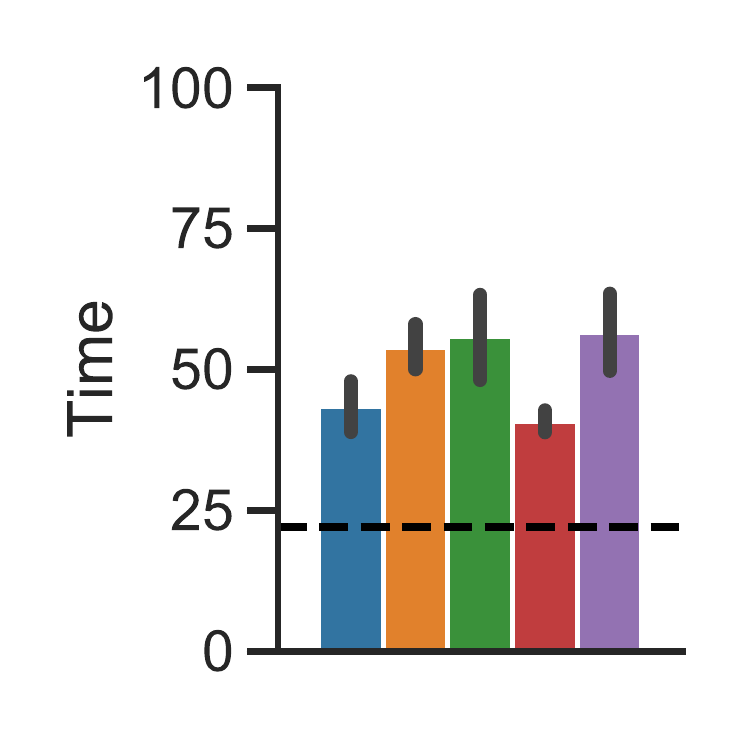}
    \includegraphics[width=\graphw]{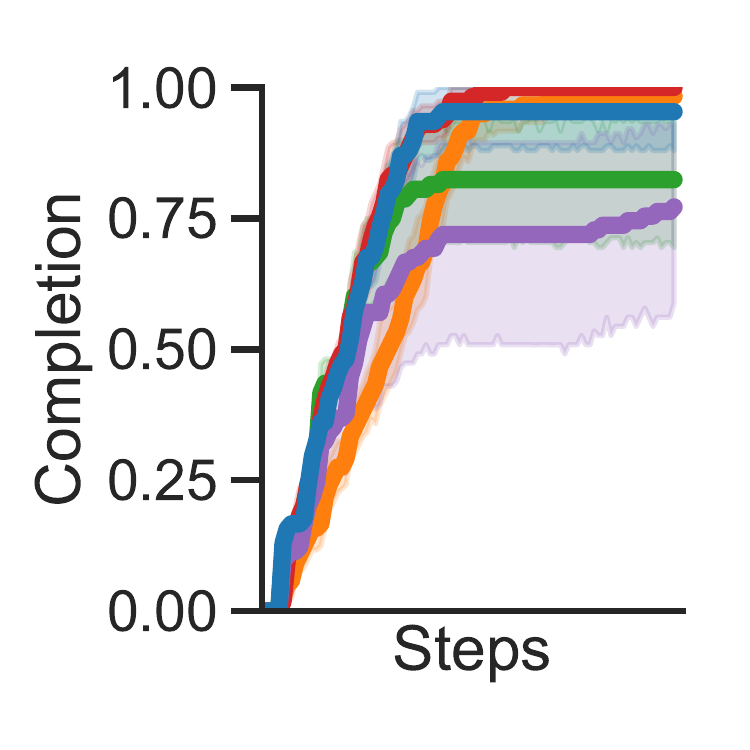}
  \end{subfigure}
  \begin{subfigure}[t]{\rightratio\linewidth}
  \centering
    \includegraphics[width=\graphw]{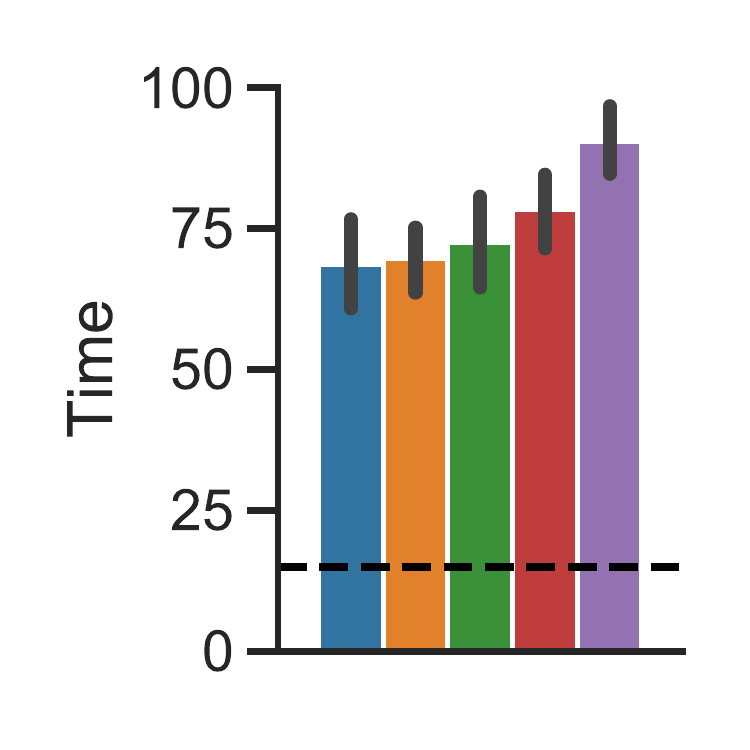}
    \includegraphics[width=\graphw]{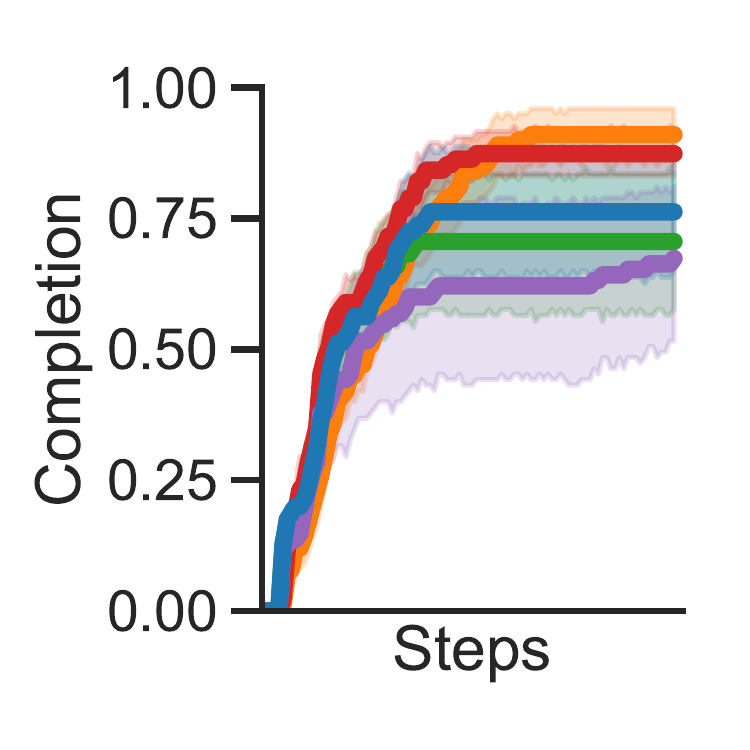}
  \end{subfigure}

  \vspace{\verticalspace}
  \begin{subfigure}[t]{\leftratio\linewidth}
  \centering
    \includegraphics[width=\gw]{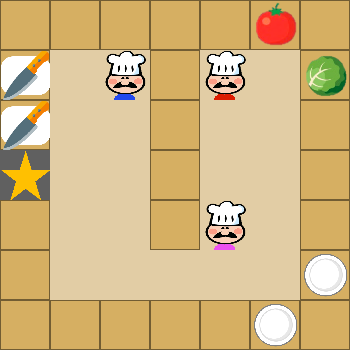}
  \end{subfigure}
  \begin{subfigure}[t]{\rightratio\linewidth}
  \centering
    \includegraphics[width=\graphw]{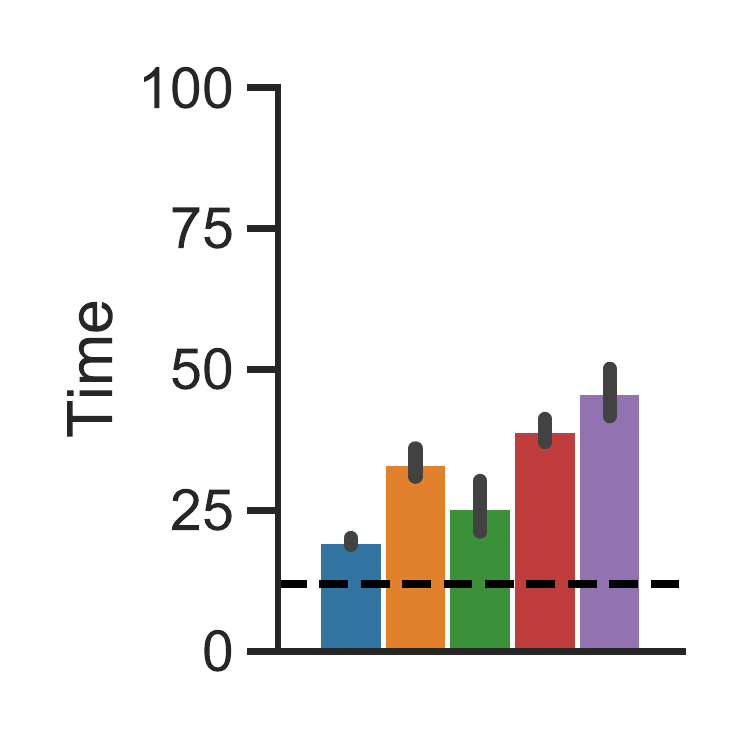}
    \includegraphics[width=\graphw]{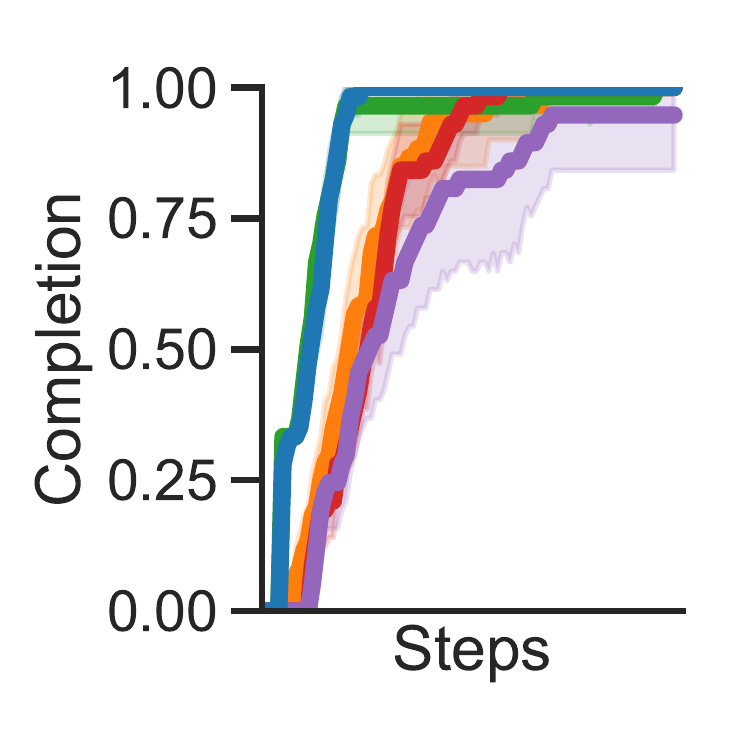}
  \end{subfigure}
  \begin{subfigure}[t]{\rightratio\linewidth}
  \centering
    \includegraphics[width=\graphw]{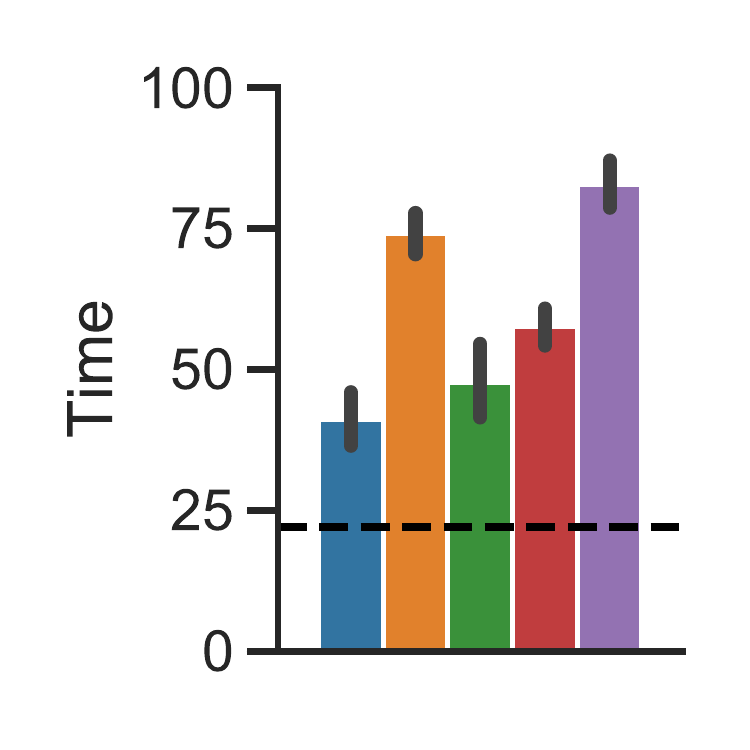}
    \includegraphics[width=\graphw]{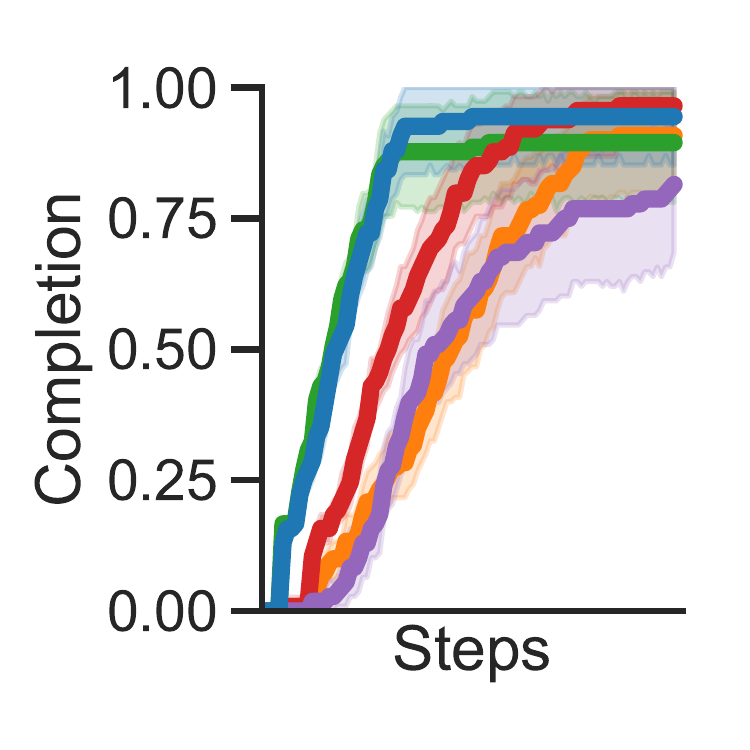}
  \end{subfigure}
  \begin{subfigure}[t]{\rightratio\linewidth}
  \centering
    \includegraphics[width=\graphw]{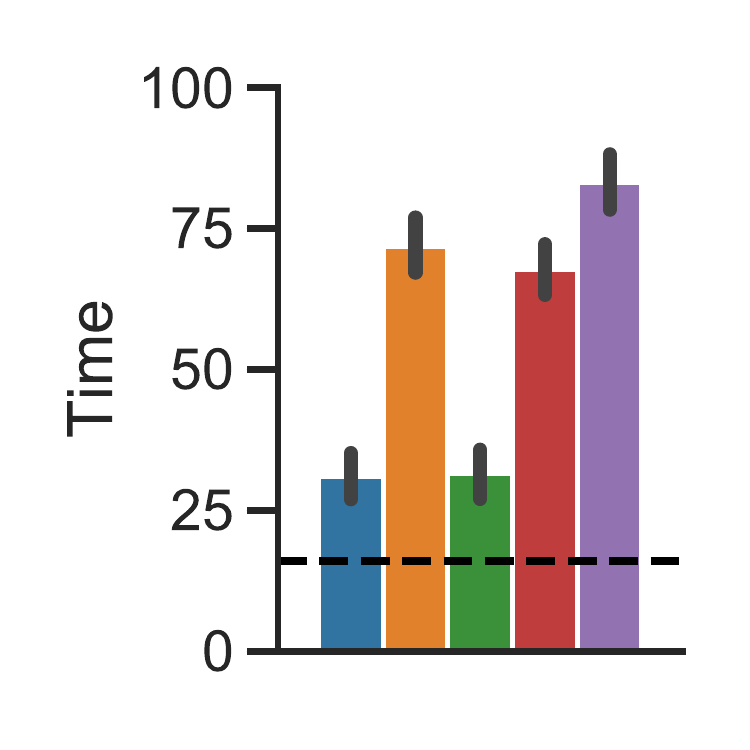}
    \includegraphics[width=\graphw]{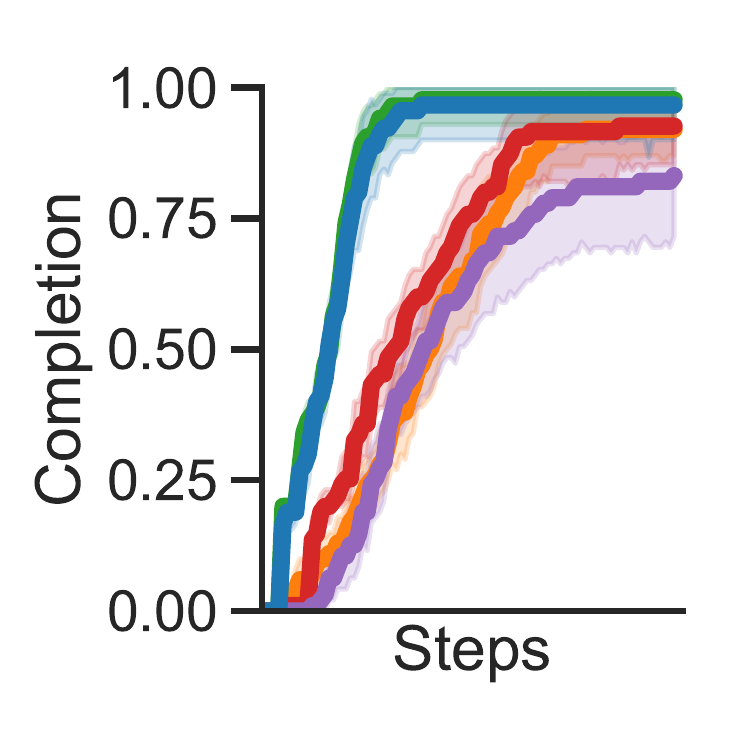}
  \end{subfigure}
  
  \vspace{\verticalspace}
  \begin{subfigure}[t]{\leftratio\linewidth}
  \centering
      \includegraphics[width=\gw]{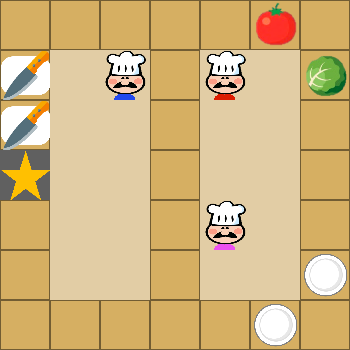}
  \end{subfigure}
  \begin{subfigure}[t]{\rightratio\linewidth}
  \centering
    \includegraphics[width=\graphw]{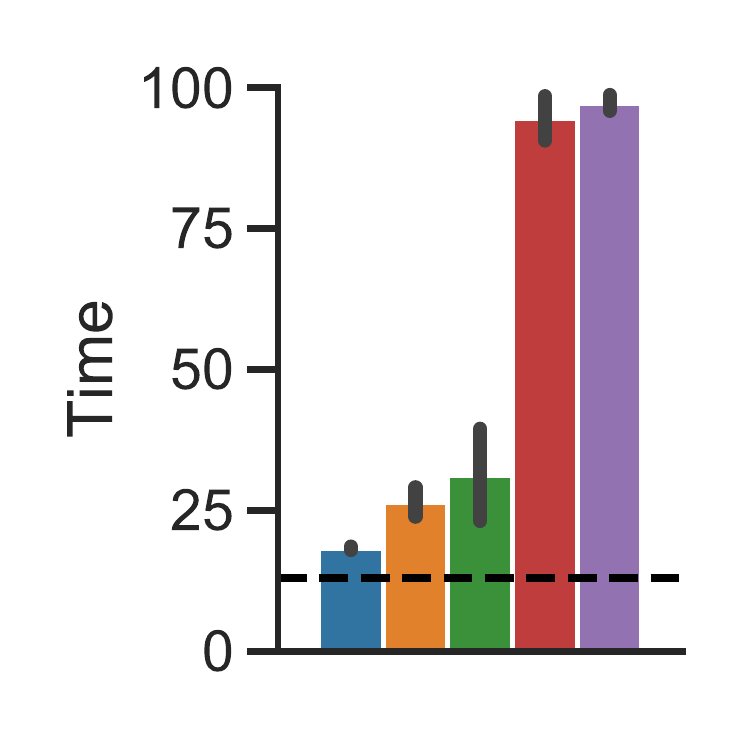}
    \includegraphics[width=\graphw]{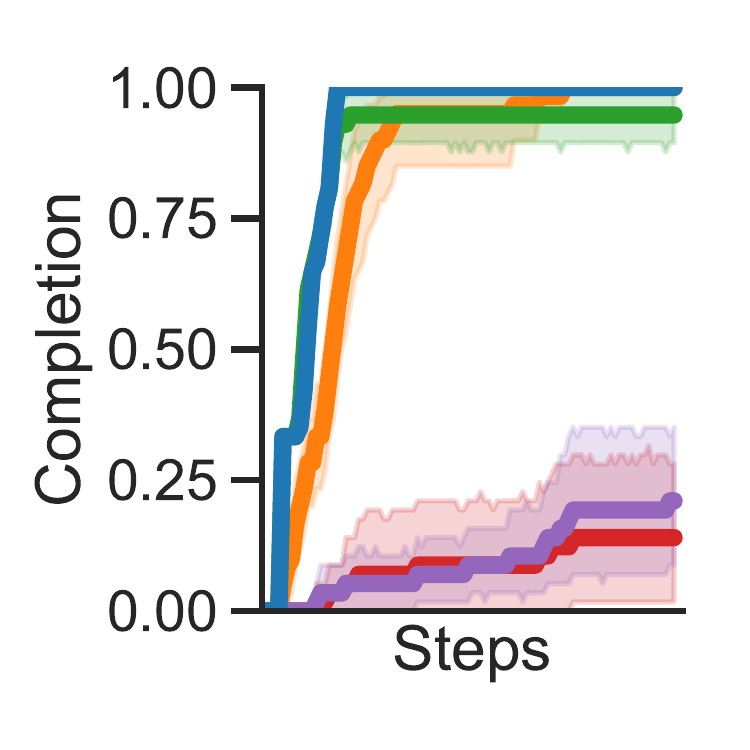}
  \end{subfigure}
  \begin{subfigure}[t]{\rightratio\linewidth}
  \centering
    \includegraphics[width=\graphw]{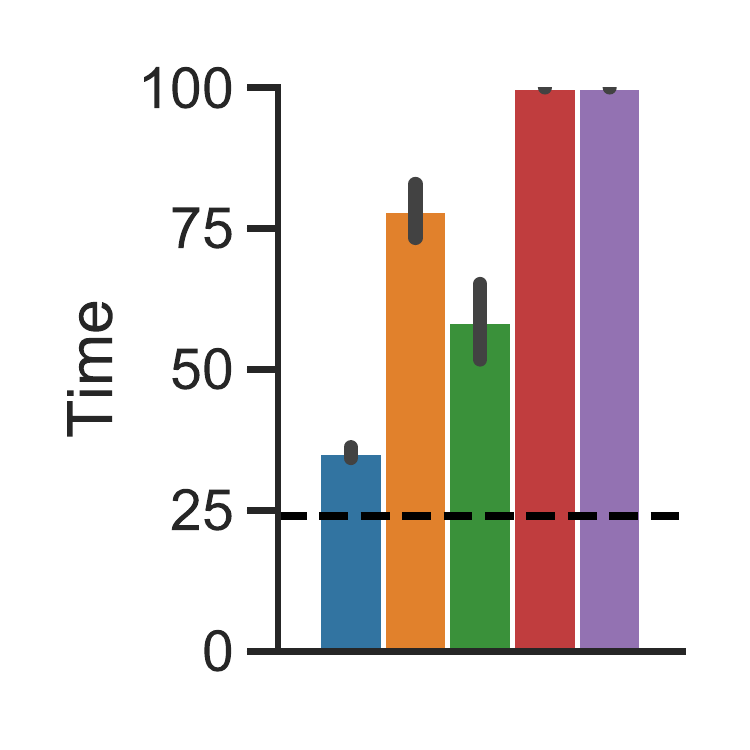}
    \includegraphics[width=\graphw]{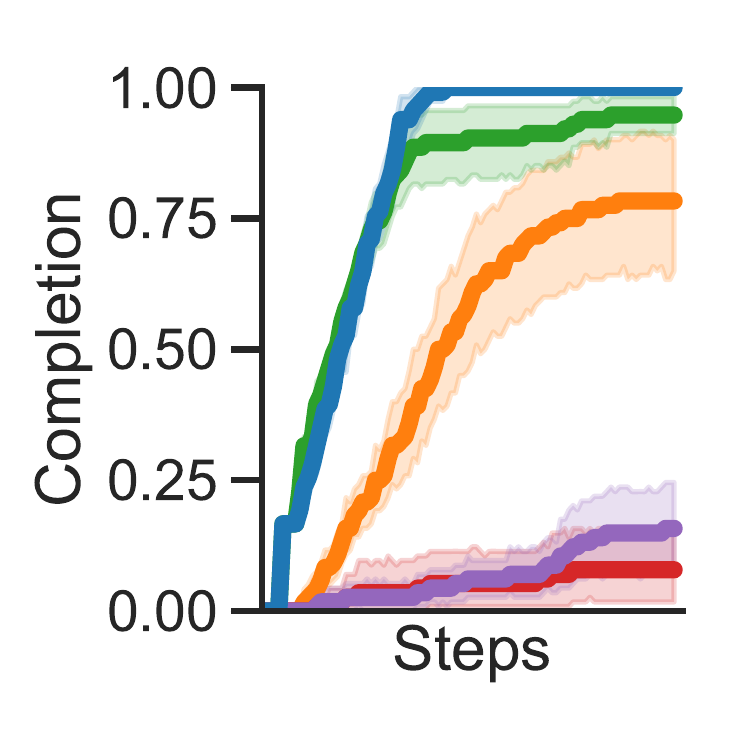}
  \end{subfigure}
  \begin{subfigure}[t]{\rightratio\linewidth}
  \centering
    \includegraphics[width=\graphw]{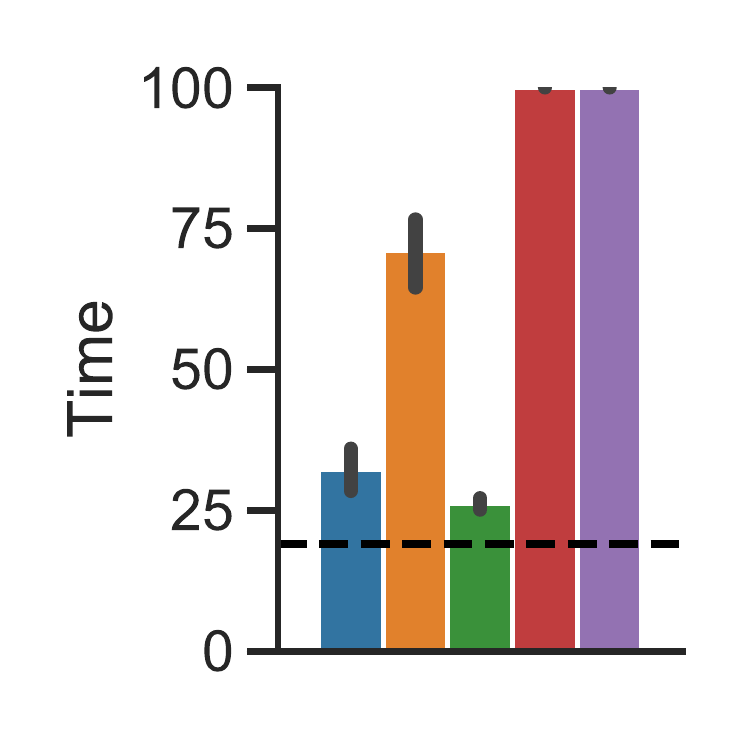}
    \includegraphics[width=\graphw]{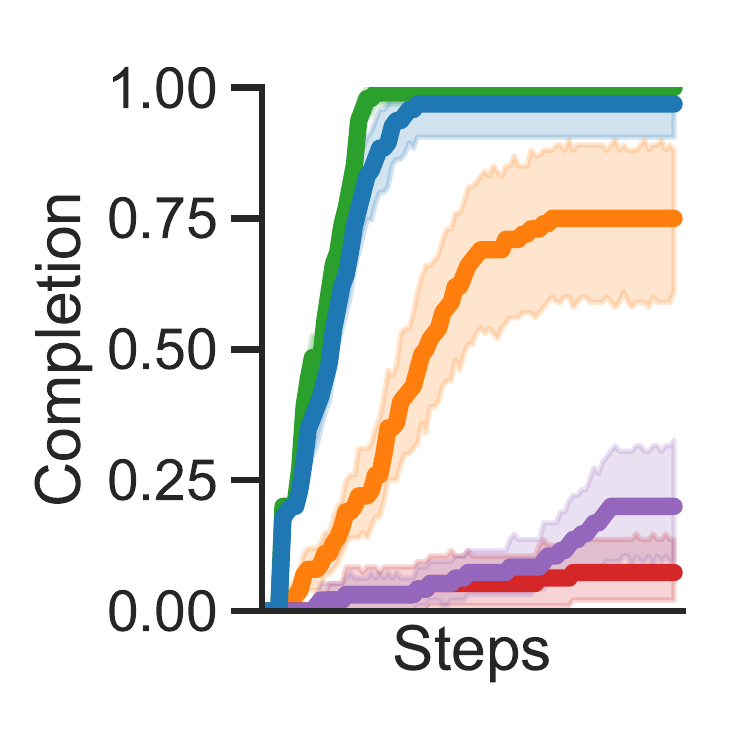}
  \end{subfigure}
\caption{Performance results for each kitchen-recipe composition (lower is better) for \textbf{three} agents. The row shows the kitchen and the column shows the recipe. Within each composition, the left graph shows the number of time steps needed to complete all sub-tasks. The dashed lines on the left graph represent the optimal performance of a centralized team. The right graph shows the fraction of sub-tasks completed over time. The full agent completes more sub-tasks and does so more quickly compared to the alternatives.
\label{fig:perf3} }
\end{figure}

\begin{figure}[tbh]
  \centering
  \newcommand{\rw}{26mm}
  \newcommand{\gw}{20mm}
  \newcommand{\bw}{20mm}
  \newcommand{\graphw}{48mm}
  \newcommand{\verticalspace}{1mm}
  \newcommand{\leftratio}{0.17}
  \newcommand{\rightratio}{0.27}

  \begin{subfigure}[t]{\leftratio\linewidth}
  \centering
    \qquad
  \end{subfigure}
  \begin{subfigure}[t]{\rightratio\linewidth}
  \centering
    \includegraphics[width=\rw]{images/recipes/SimpleTomato}
  \end{subfigure}
  \begin{subfigure}[t]{\rightratio\linewidth}
  \centering
    \includegraphics[width=\rw]{images/recipes/TL}
  \end{subfigure}
  \begin{subfigure}[t]{\rightratio\linewidth}
  \centering
    \includegraphics[width=\rw]{images/recipes/Salad}
  \end{subfigure}

  \begin{subfigure}[c]{\leftratio\linewidth}
  \centering
      \includegraphics[width=\gw]{images/levels/open.png}
  \end{subfigure}
  \begin{subfigure}[c]{\rightratio\linewidth}
  \centering
    \includegraphics[width=\graphw]{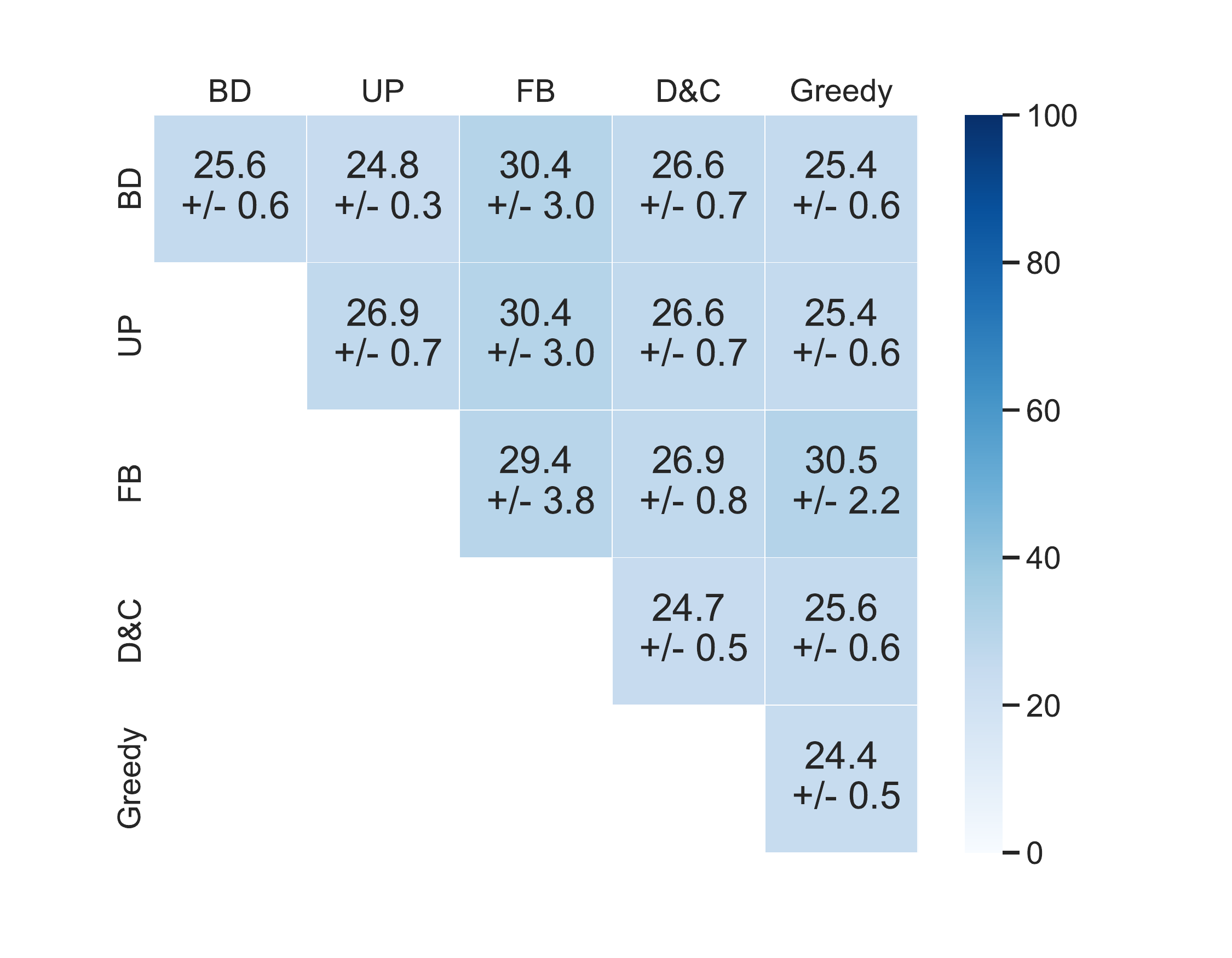}
  \end{subfigure}
  \begin{subfigure}[c]{\rightratio\linewidth}
  \centering
    \includegraphics[width=\graphw]{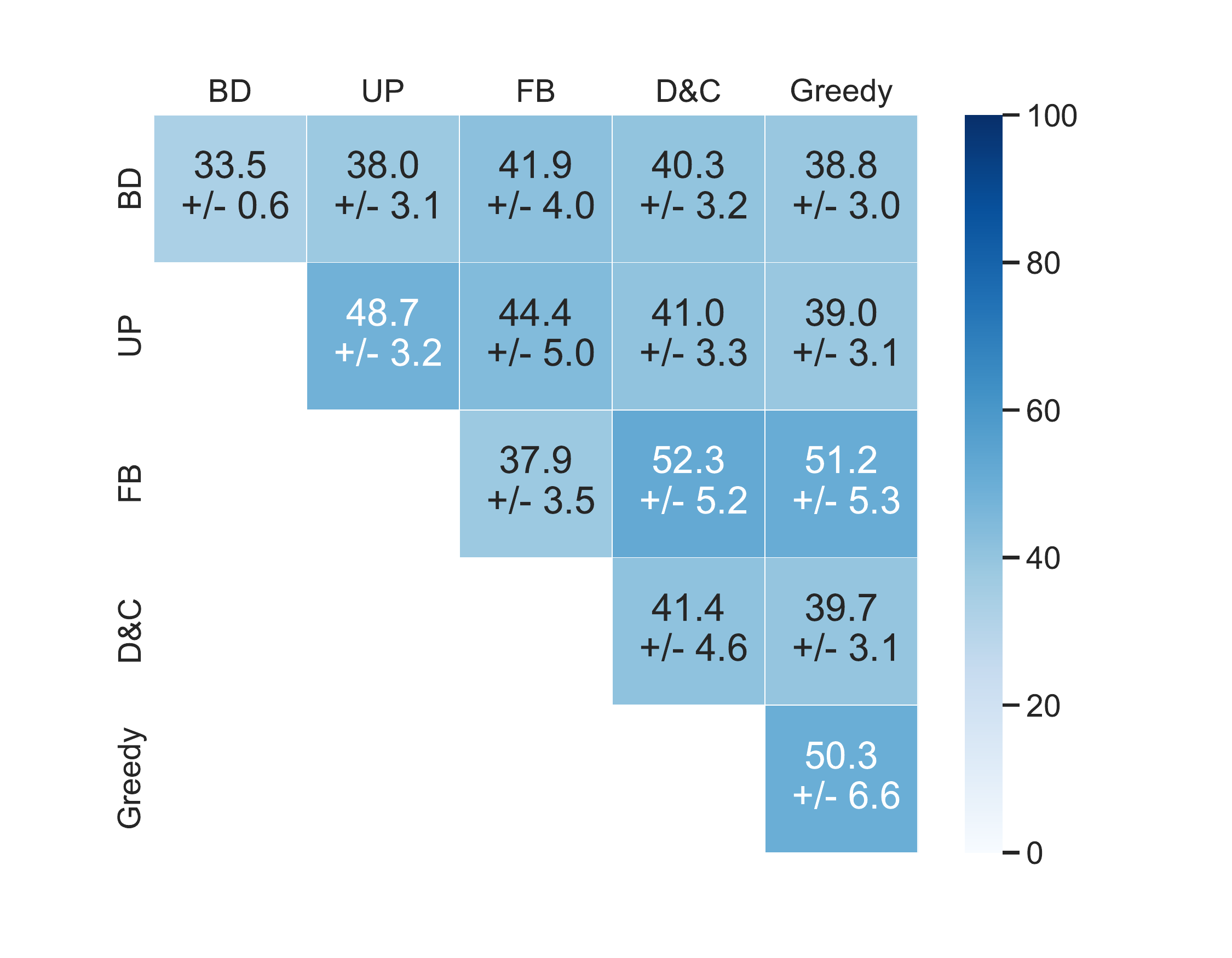}
  \end{subfigure}
  \begin{subfigure}[c]{\rightratio\linewidth}
  \centering
    \includegraphics[width=\graphw]{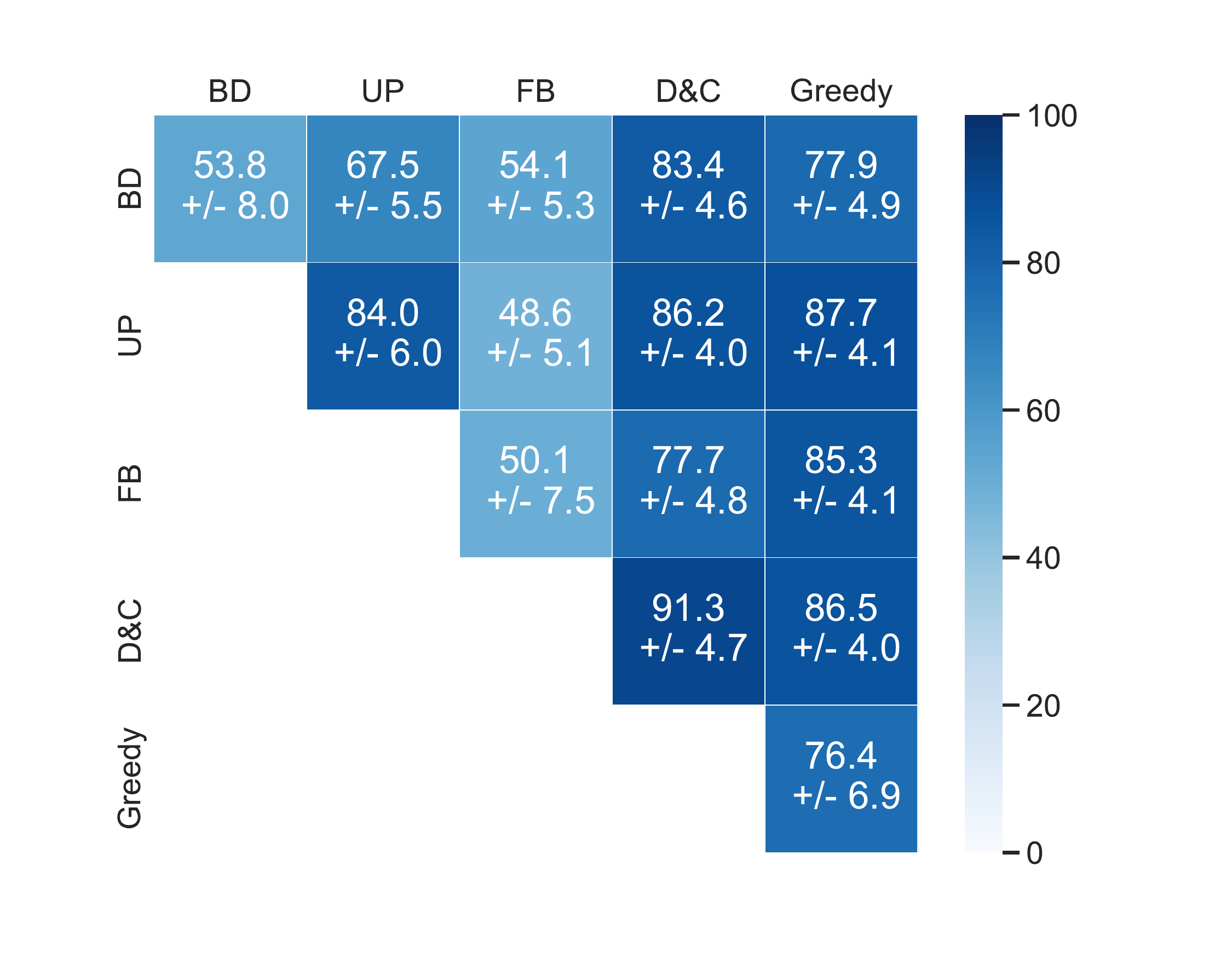}
  \end{subfigure}

  \vspace{\verticalspace}
  \begin{subfigure}[c]{\leftratio\linewidth}
  \centering
    \includegraphics[width=\gw]{images/levels/partial.png}
  \end{subfigure}
  \begin{subfigure}[c]{\rightratio\linewidth}
  \centering
    \includegraphics[width=\graphw]{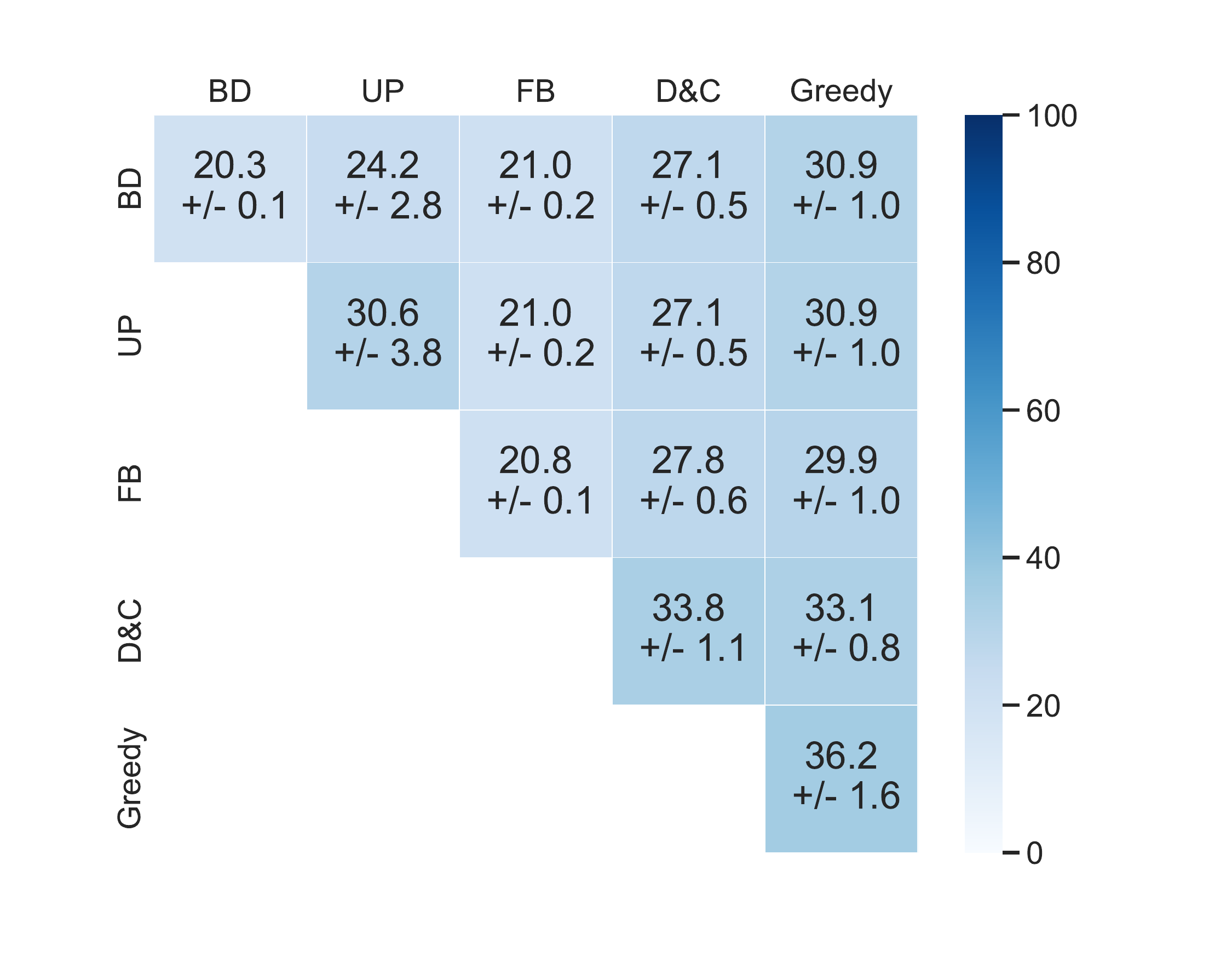}
  \end{subfigure}
  \begin{subfigure}[c]{\rightratio\linewidth}
  \centering
    \includegraphics[width=\graphw]{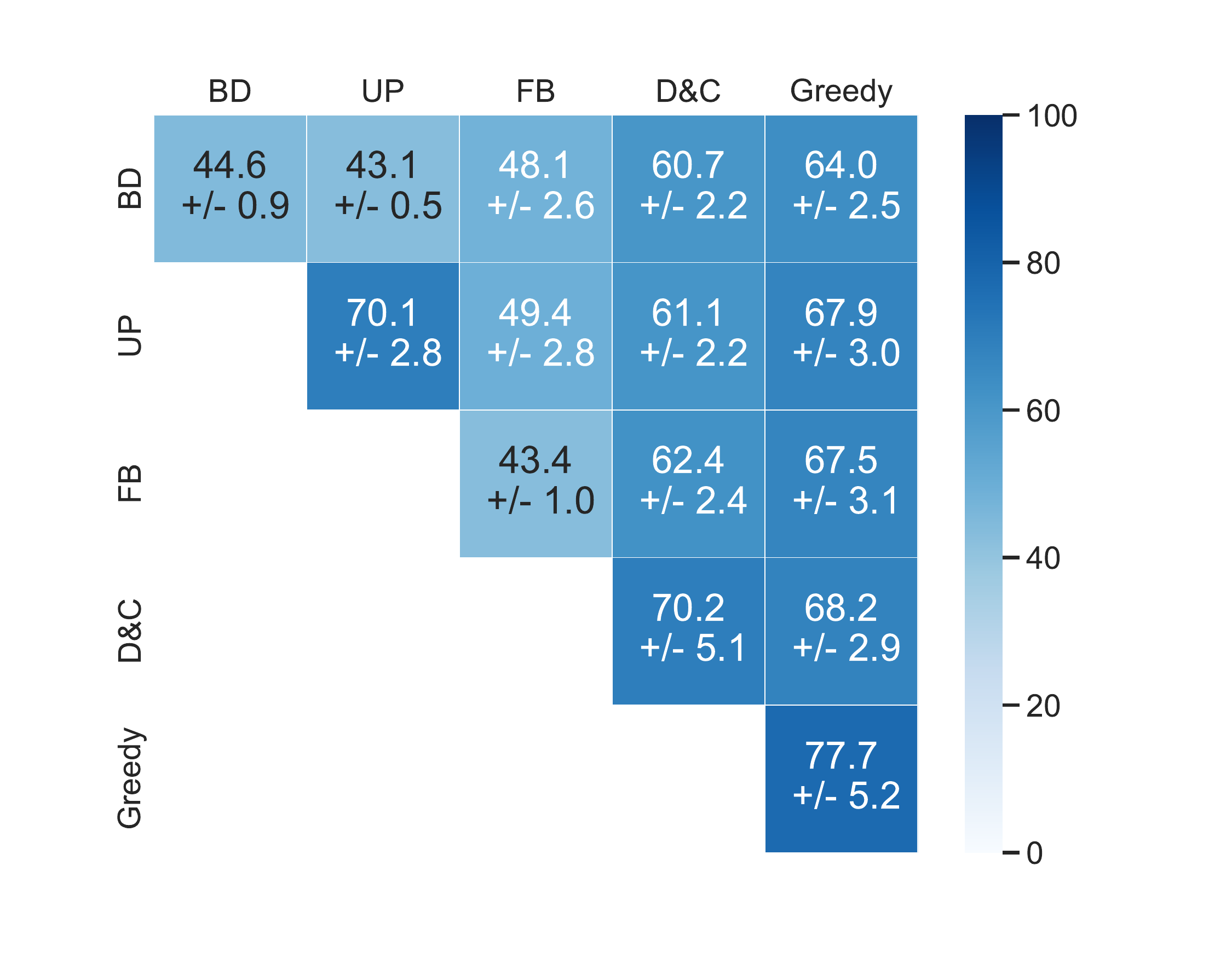}
  \end{subfigure}
  \begin{subfigure}[c]{\rightratio\linewidth}
  \centering
    \includegraphics[width=\graphw]{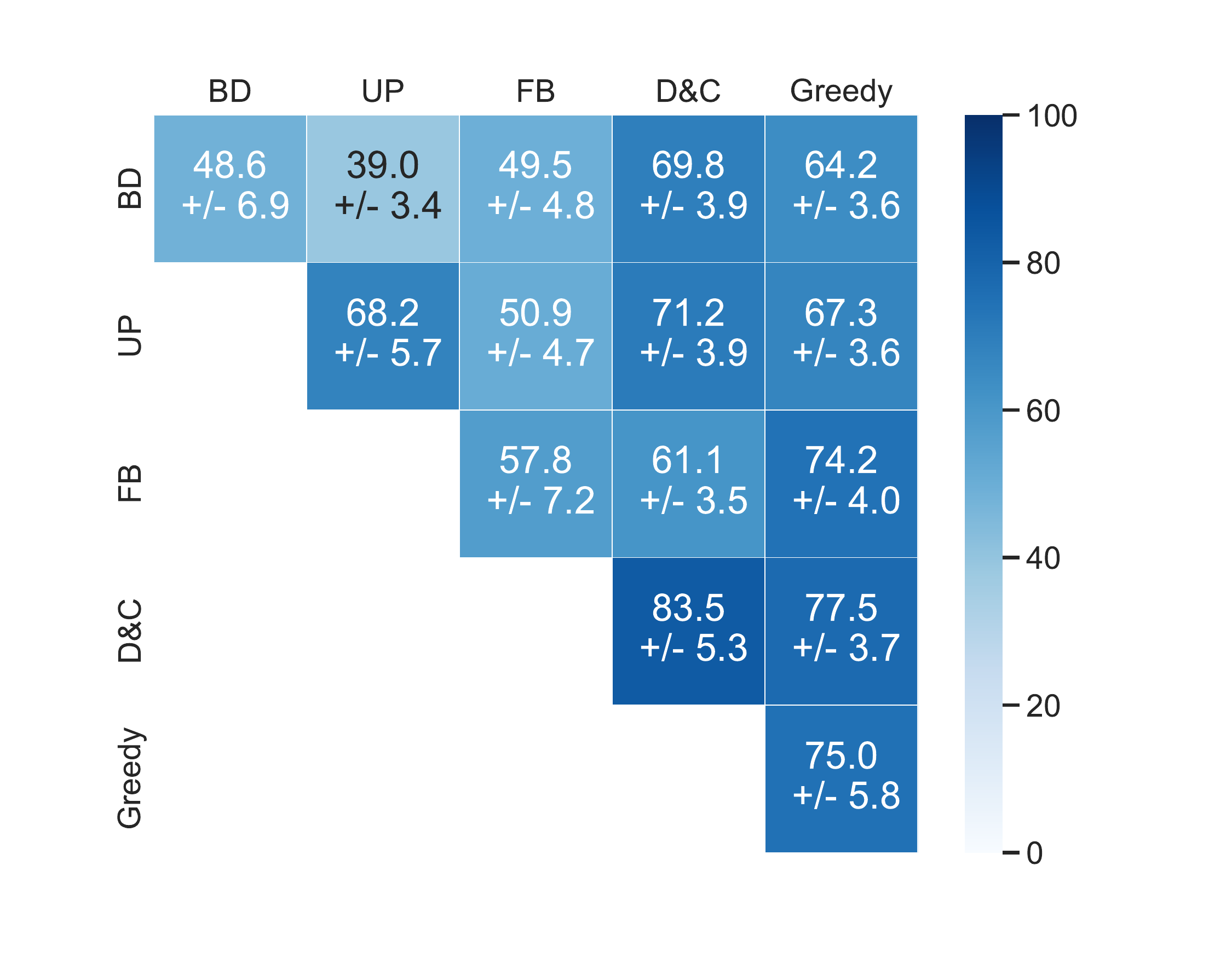}
  \end{subfigure}
  
  \vspace{\verticalspace}
  \begin{subfigure}[c]{\leftratio\linewidth}
  \centering
      \includegraphics[width=\gw]{images/levels/full.png}
  \end{subfigure}
  \begin{subfigure}[c]{\rightratio\linewidth}
  \centering
    \includegraphics[width=\graphw]{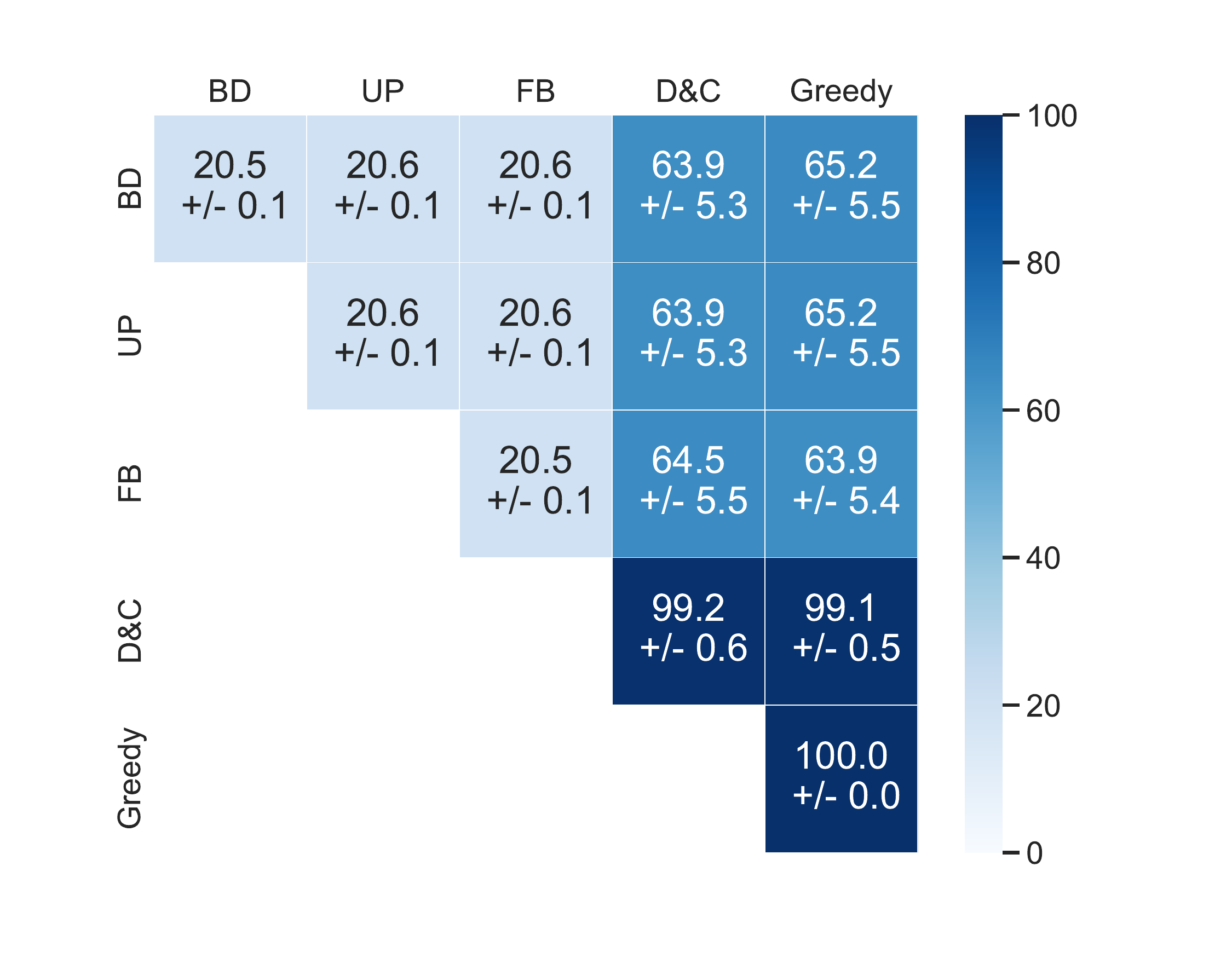}
  \end{subfigure}
  \begin{subfigure}[c]{\rightratio\linewidth}
  \centering
    \includegraphics[width=\graphw]{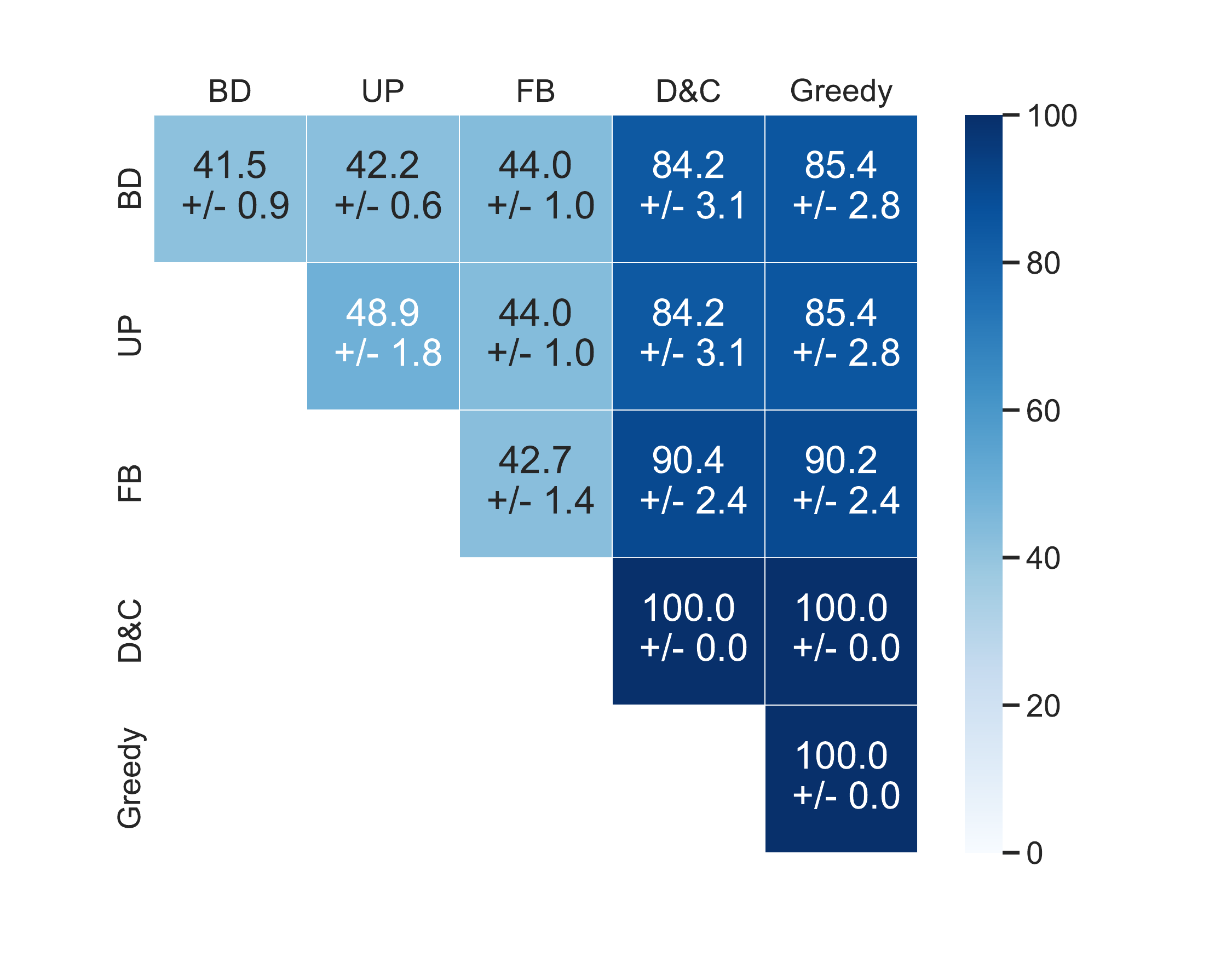}
  \end{subfigure}
  \begin{subfigure}[c]{\rightratio\linewidth}
  \centering
    \includegraphics[width=\graphw]{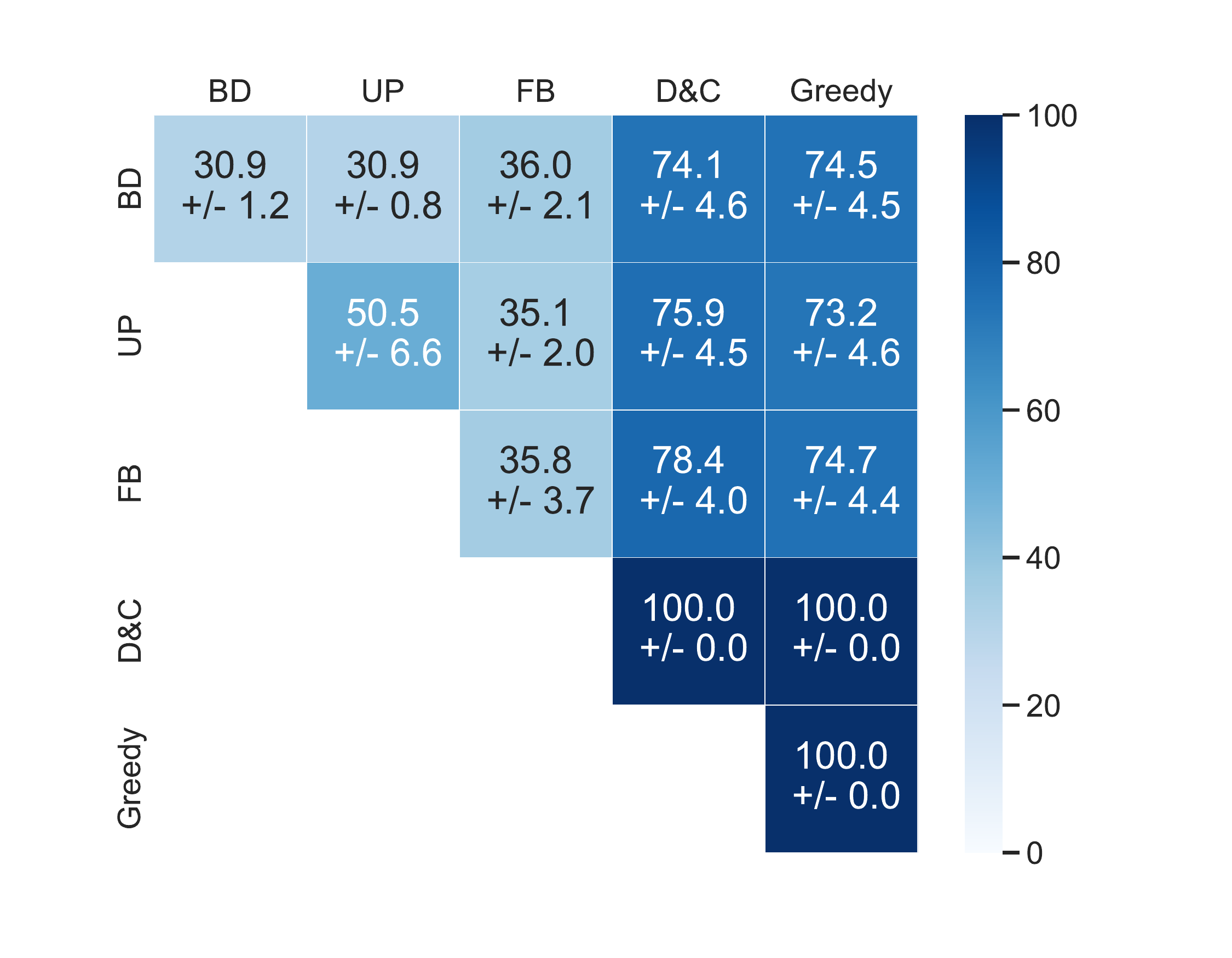}
  \end{subfigure}
\caption{Heat map performance for each kitchen-recipe composition (lower is better) for two agents using different models. The figure breaks down the aggregate performance from Figure~\ref{fig:adhoc} by kitchen (row) and recipe(column). In most compositions, as models perform as better adhoc coordinators as they become more ``Bayesian Delegation''-like (going bottom to top by row, right to left by column). \label{fig:adhoc_breakdown}}
\end{figure}

\newpage
\section{Environment Details \label{appendix:env_details}}
\begin{SCfigure}[50][tbh]
    \newcommand{\gw}{25mm}
    \begin{subfigure}[b]{.19\columnwidth}
      \centering
      \includegraphics[width=\gw]{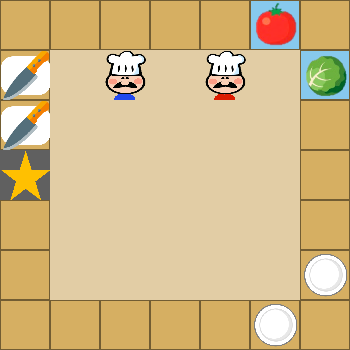}
      \caption{}
    \end{subfigure}
    \begin{subfigure}[b]{.19\columnwidth}
      \centering
      \includegraphics[width=\gw]{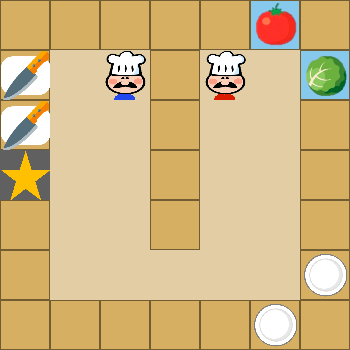}
      \caption{}
    \end{subfigure}
    \begin{subfigure}[b]{.19\columnwidth}
      \centering
      \includegraphics[width=\gw]{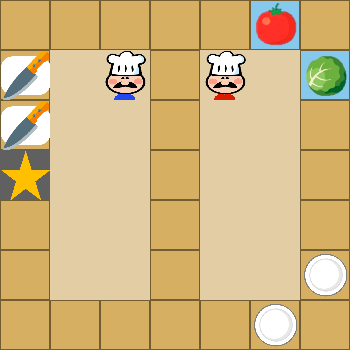}
      \caption{}
    \end{subfigure}
    \caption{The Overcooked kitchens only differ in counter placements.
    In (a) \textit{Open-Divider}, agents can move between both sides of the kitchen. In (b) \textit{Partial-Divider}, agents must pass through a narrow bottleneck. In (c) \textit{Full-Divider}, agents are confined to one half of the space.
    \label{fig:env_all}}
\end{SCfigure}

\begin{figure}[tbh]
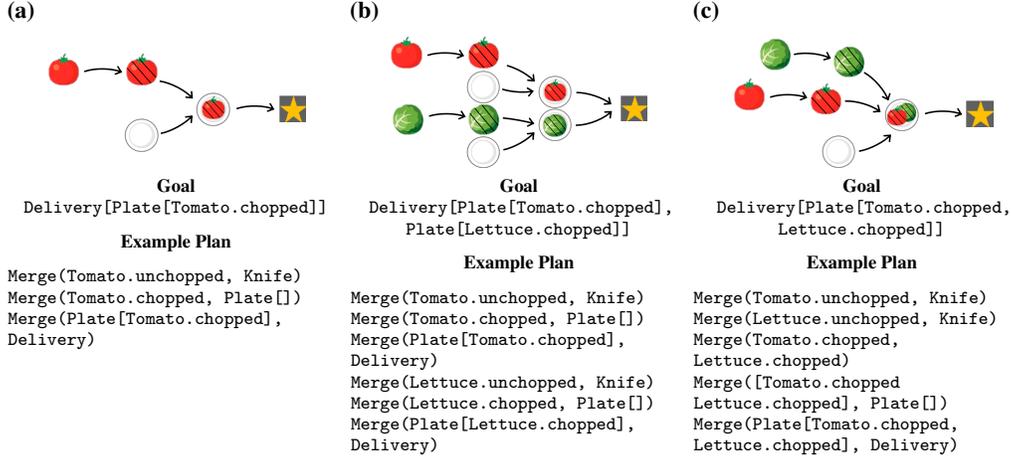

  \centering
  \newcommand{\gw}{35mm}
  \begin{subfigure}[t]{.32\linewidth}
    \small{\textbf{(a)}}
    \scriptsize{
      \begin{center}
        \includegraphics[width=\gw]{images/recipes/SimpleTomato}

        \textbf{Goal}\\
        \texttt{Delivery[Plate[Tomato.chopped]]} \\
      \end{center} 
      \begin{center}\textbf{Example Plan}\end{center}
      \texttt{Merge(Tomato.unchopped, Knife) \\
        Merge(Tomato.chopped, Plate[]) \\
        Merge(Plate[Tomato.chopped], Delivery)
      }
    }
    \phantomcaption
    \label{fig:rec_tomato}
  \end{subfigure}
  \begin{subfigure}[t]{.32\linewidth}
    \small{\textbf{(b)}}
    \scriptsize{
      \begin{center}
        \includegraphics[width=\gw]{images/recipes/TL}

        \textbf{Goal} \\
        \texttt{Delivery[Plate[Tomato.chopped], Plate[Lettuce.chopped]]} \\
      \end{center}
      \begin{center}\textbf{Example Plan}\end{center}
      \texttt{Merge(Tomato.unchopped, Knife)  \\
        Merge(Tomato.chopped, Plate[]) \\
        Merge(Plate[Tomato.chopped], Delivery) \\
        Merge(Lettuce.unchopped, Knife) \\
        Merge(Lettuce.chopped, Plate[]) \\
        Merge(Plate[Lettuce.chopped], Delivery)
      }
    }
  \phantomcaption
  \label{fig:rec_tl}
  \end{subfigure}
  \begin{subfigure}[t]{.32\linewidth}
    \small{\textbf{(c)}}
    \scriptsize{
      \begin{center}
        \includegraphics[width=\gw]{images/recipes/Salad}

        \textbf{Goal} \\
        \texttt{Delivery[Plate[Tomato.chopped, Lettuce.chopped]]} \\
      \end{center}
      \begin{center}\textbf{Example Plan}\end{center}
      \texttt{Merge(Tomato.unchopped, Knife) \\
        Merge(Lettuce.unchopped, Knife) \\
        Merge(Tomato.chopped, Lettuce.chopped) \\
        Merge([Tomato.chopped Lettuce.chopped], Plate[]) \\
        Merge(Plate[Tomato.chopped, Lettuce.chopped], Delivery)
      }
    }
  \phantomcaption
  \label{fig:rec_salad}
  \end{subfigure}
  \caption{Recipes and example partial orderings. All sub-tasks are expressed in the \texttt{Merge} operator. In (a) \textit{Tomato}, the task is to take an unchopped tomato and then chop, plate, and deliver it. In (b) \textit{Tomato+Lettuce}, the task builds on \textit{Tomato} and adds chopping, plating, and delivering a piece of lettuce. In (c) \textit{Salad}, the two chopped foods are combined on a single plate and delivered. The example plans show one of many possible orderings for completing the recipe. \label{fig:recipe}}
\end{figure}

The full set of environments and tasks used in the experiments is shown in Figure~\ref{fig:env_all} and Figure~\ref{fig:recipe}.
The state of the environment is described by the objects and dynamics in Table~\ref{tab:objects}. The partially ordered set of sub-tasks $\mathcal{T}$ is given in the environment and generated by representing each recipe as an instance of STRIPS, an action language \cite{strips}. Each instance consists of an initial state, a specification of the goal state, and a set of actions with preconditions that dictate what must be true/false for the action to be executable, and postconditions that dictate what is made true/false when the action is executed. For instance, for the STRIPS instance of the recipe \textit{Tomato}, the initial state is the initial configuration of the environment (i.e. all objects and their states), the specification of the goal state is \texttt{Delivery[Plate[Tomato.chopped]]}, and the actions are the \texttt{Merge} sub-tasks e.g. the ones show in Figure~\ref{fig:rec_tomato}. A plan for a STRIPS instance is a sequence of actions that can be executed from the initial state and results in a goal state. Examples of such plans are shown in Figure~\ref{fig:recipe}. To generate these partial orderings, we construct a graph for each recipe in which the nodes are the states of the environment objects and the edges are valid actions. We then run breadth-first-search starting from the initial state to determine the nearest goal state, and explore all shortest ``recipe paths" between the two states. We then return $\mathcal{T}$, the set of actions collectively taken on those recipe paths terminating at a completed recipe.

\begin{table}[bt]
\textbf{Object state representation:}\\
\begin{center}
  \begin{tabular}{l | c | l}
    Type & Location & Status \\ \hline
    Agent & \{x, y\} & [] \\
    Plate & \{x, y\} & [] \\
    Counter & \{x, y\} & [] \\
    Delivery & \{x, y\} & [] \\
    Knife & \{x, y\} & N/A \\
    Tomato & \{x, y\} & \{chopped, unchopped\} \\
    Lettuce & \{x, y\} & \{chopped, unchopped\}
  \end{tabular}
\end{center}    

\vspace{1em}

  \textbf{Interaction dynamics:}\\
  \flushleft
\qquad  Food.unchopped + Knife $\xrightarrow{}$ Food.chopped + Knife \\
\qquad  Food1 + Food2 $\xrightarrow{}$ [Food1, Food2] \\
\qquad  X + Y[] $\xrightarrow{}$ Y[X] \\
\vspace{1em}

  \caption{State representation and transitions for the objects and interactions in the Overcooked environments. The two food items (tomato and lettuce) can be in either chopped or unchopped states. Objects with status $[]$ are able to ``hold'' other objects. For example, an Agent holding a Plate holding an unchopped tomato would be denoted Agent[Plate[Tomato.unchopped]]. Once combined, these nested objects share the same $\{x,y\}$ coordinates and movement. Interaction dynamics occur when the two objects are in the same $\{x,y\}$ coordinates. \label{tab:objects}}
\end{table}

\newpage
\section{Behavioral Experiment}

\begin{figure}[tb]
    \centering 

  \newcommand{\plotswidth}{0.32}
  \newcommand{\rw}{23mm}
  \newcommand{\gw}{17mm}
  \newcommand{\bw}{40mm}
  \newcommand{\bww}{28mm}
  \captionsetup[subfigure]{justification=justified,singlelinecheck=false}

\begin{subfigure}{\plotswidth\textwidth}
    \caption{Recipe: \textit{Tomato}} \label{fig:traj1}
    \centering
    \includegraphics[width=\rw]{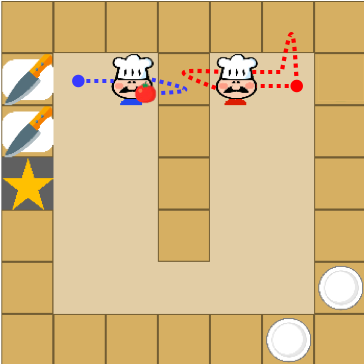}
    \includegraphics[width=\gw]{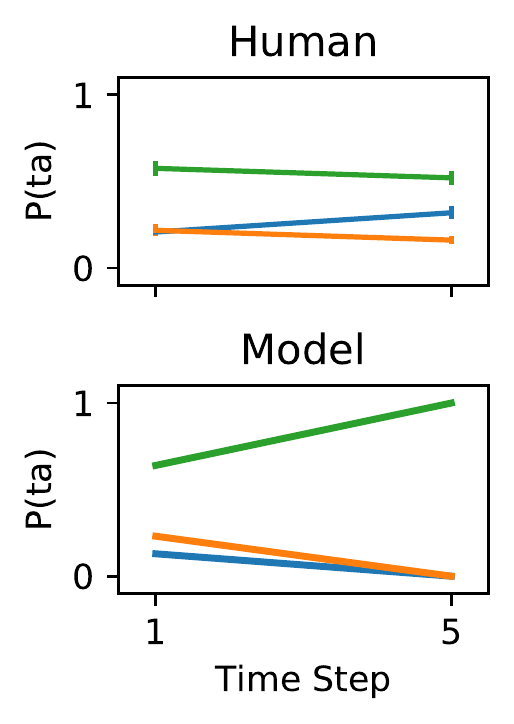}
    \vfill
    \includegraphics[width=\bw]{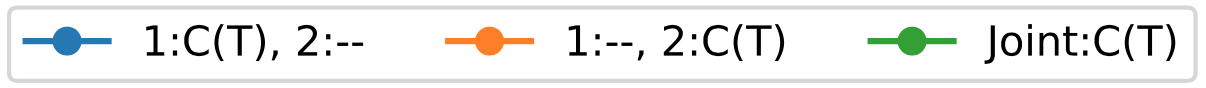}
\end{subfigure} 
\begin{subfigure}{\plotswidth\textwidth}
    \caption{Recipe: \textit{Tomato}} \label{fig:traj2}
    \centering
    \includegraphics[width=\rw]{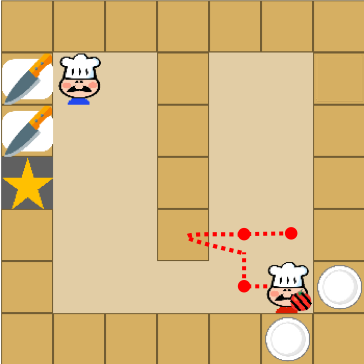}
    \includegraphics[width=\gw]{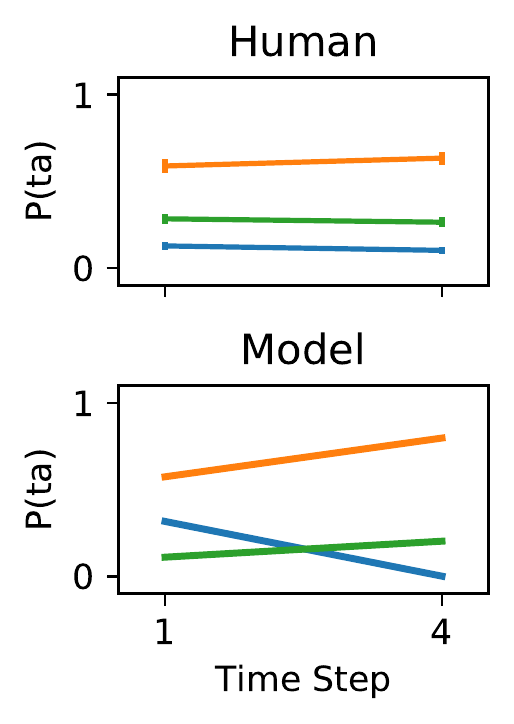}
    \vfill
    \includegraphics[width=\bw]{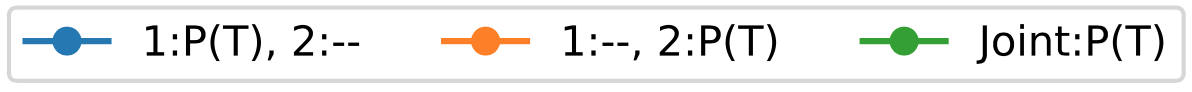}
\end{subfigure} 
\begin{subfigure}{\plotswidth\textwidth}
    \caption{Recipe: \textit{Tomato}} \label{fig:traj3}
    \centering
    \includegraphics[width=\rw]{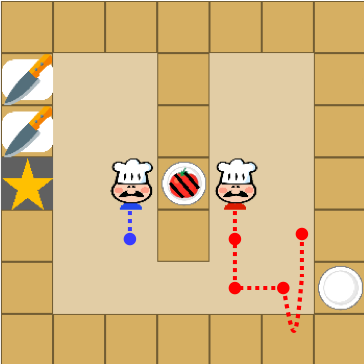}
    \vspace{1mm}
    \includegraphics[width=\gw]{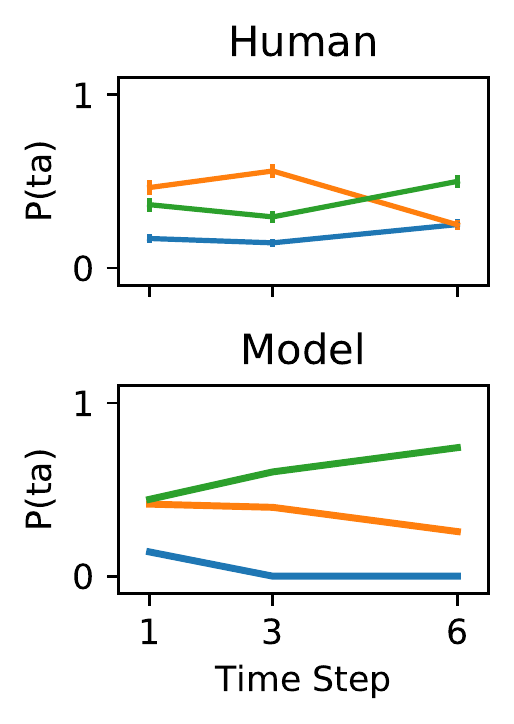}
    \vfill
    \includegraphics[width=\bw]{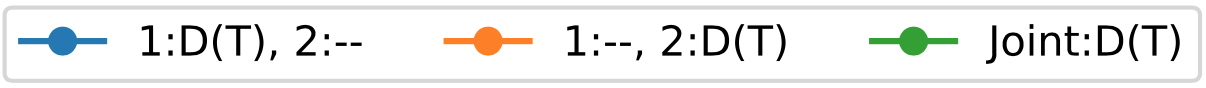}
\end{subfigure}

\medskip

\begin{subfigure}{\plotswidth\textwidth}
    \caption{Recipe: \textit{Salad}} \label{fig:traj4}
    \centering
    \includegraphics[width=\rw]{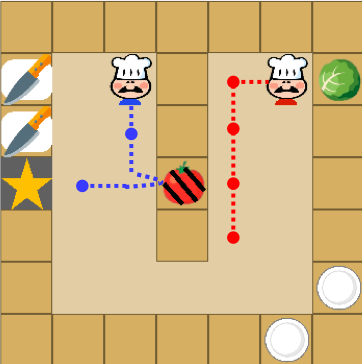}
    \includegraphics[width=\gw]{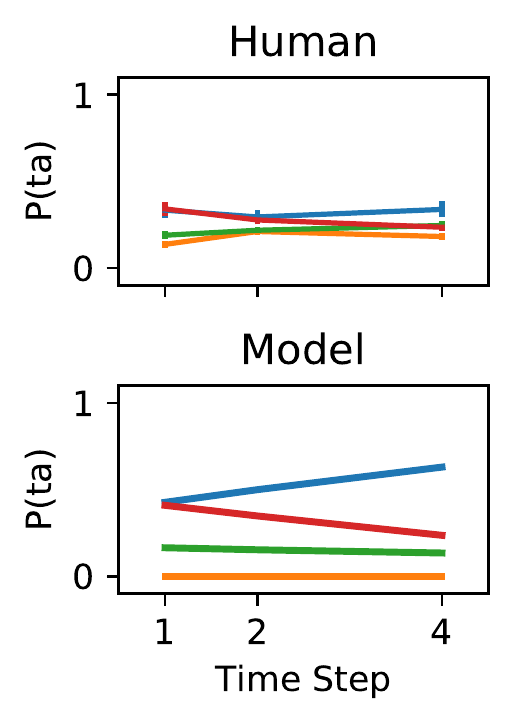}
    \vfill
    \includegraphics[width=\bww]{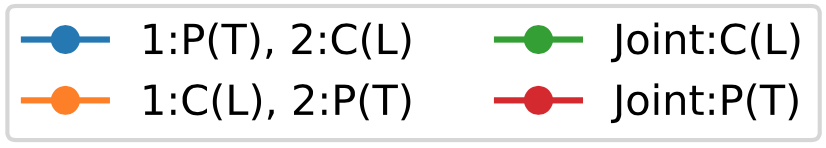}
\end{subfigure} \hfil 
\begin{subfigure}{\plotswidth\textwidth}
    \caption{Recipe: \textit{Salad}} \label{fig:traj5}
    \centering
    \includegraphics[width=\rw]{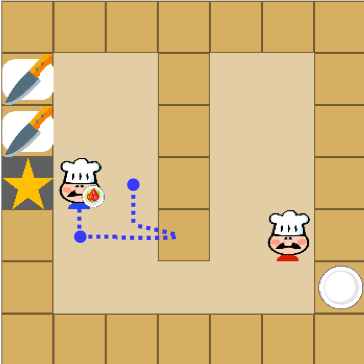}
    \vspace{1mm}
    \includegraphics[width=\gw]{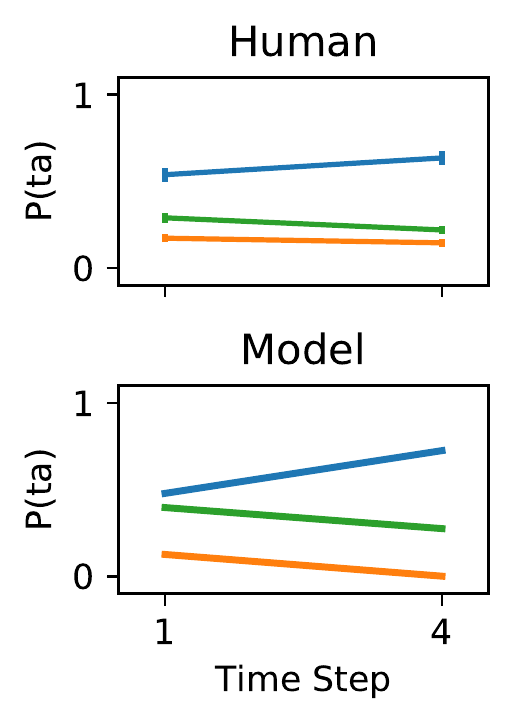}
    \vfill
    \includegraphics[width=\bw]{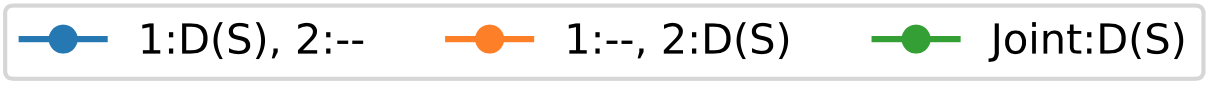}
\end{subfigure} \hfil 
\begin{subfigure}{\plotswidth\textwidth}
    \caption{Recipe: \textit{Salad}} \label{fig:traj6}
    \centering
    \includegraphics[width=\rw]{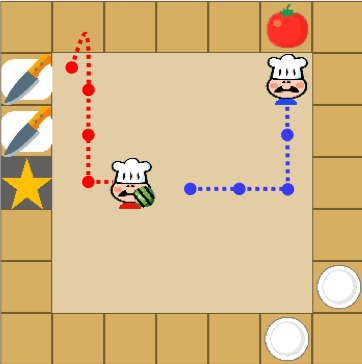}
    \includegraphics[width=\gw]{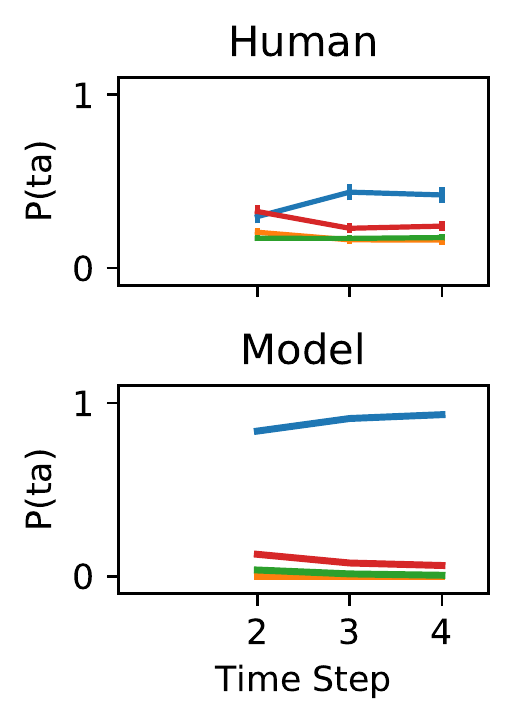}
    \vfill
    \includegraphics[width=\bww]{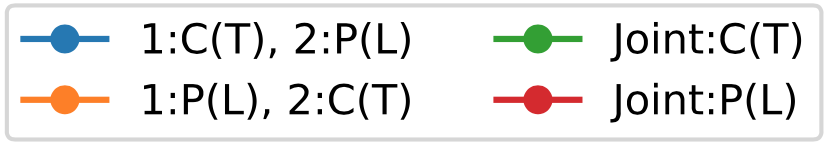}
\end{subfigure}
\caption{Scenarios and results for the human behavioral experiment. Each agent's past trajectory is illustrated by a dotted path, with sharp curves into counters representing picking up or putting down an object. To the right of each trajectory are the inferences made by the model and the average participant inference. The legend notes the possible task allocations of agents (1 or 2) working individually or together (Joint): C = chop, P = plate, D = deliver, T = tomato, L = lettuce, and S = salad. E.g., 1:C(T) refers to Agent 1 chopping the tomato.  Error bars are the standard error of the mean. }
\label{fig:traj}
\end{figure}

\begin{figure}
    \centering
  \newcommand{\gw}{30mm}
  \begin{subfigure}[t]{\linewidth}
    \includegraphics[width=\linewidth]{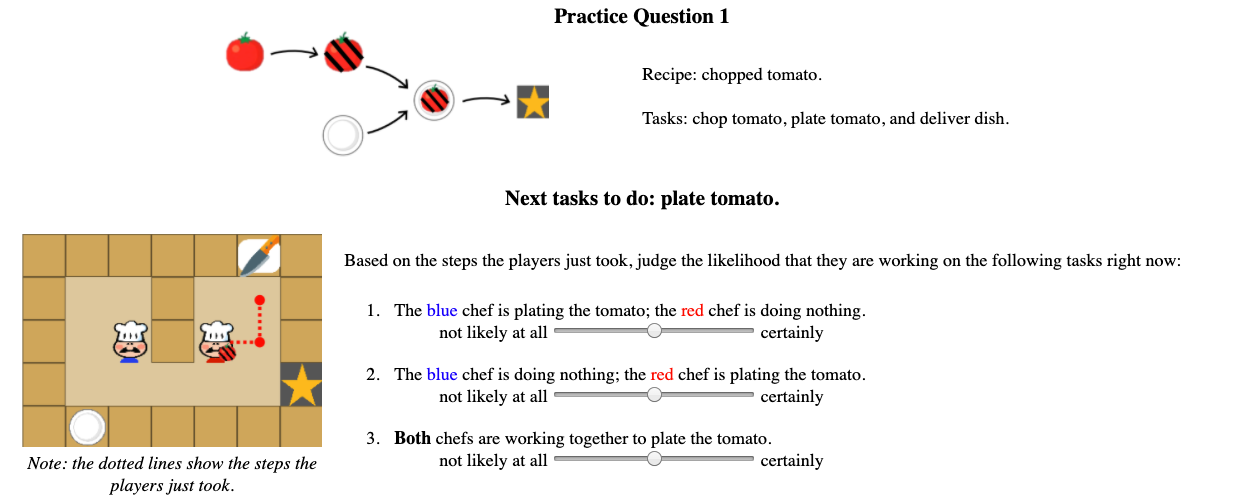}
  \end{subfigure}
  \caption{Example question for the behavioral experiment. Each question contains a description of the recipe at hand and all of the tasks needed to complete that recipe. The question also shows all the feasible tasks (see \textbf{Next tasks to do}) in the current environment state shown on the left gridworld. Below are the likelihood judgements for the task allocations of the feasible tasks.  \label{fig:human_expt_example}}
\end{figure}

Our experiment can be viewed at: \href{https://manycooks.github.io/website/experiment.html}{\texttt{https://manycooks.github.io/website/experiment.html}}.
60 participants were recruited from Amazon Mechanical Turk and were paid \$0.80 for the experiment, which took 10-15 minutes to complete. Participants first answered two practice questions designed to test their understanding of the instructions, which were to observe each scene and make judgements about what tasks the two agents shown are doing and whether they are working together or not. After the practice questions, each participant was presented with the set of trajectories shown in Figure~\ref{fig:traj} spread out over 15 questions. Each question was modelled like Figure~\ref{fig:human_expt_example}: the page displayed the recipe information followed by a past trajectory of the agents and a series of judgement queries. After filtering for correct understanding based on the responses to the practice questions, we used $N = 45$ participants in our final analysis.

\section{Computational Experiments
\label{appendix:compexperiments}}

We used a computing cluster to run all computational experiments and model predictions for the behavioral experiment trajectories shown in Figure~\ref{fig:traj}. Each simulation with two agents was run in parallel (3 recipes, 3 environments, 5 models, 20 seeds) each on 1 CPU core, which took up to 15 GB of memory and roughly 3 hours to complete.

We next describe the details of our BRTDP implementation: \cite{mcmahan2005bounded}. 
\begin{align*}
V_{\mathcal{T}_i}^{b}(s) = \min_{a \in \mathcal{A}_i} Q_{\mathcal{T}_i}^{b}(s, a), \quad V_{\mathcal{T}_i}^{b}(g) = 0
\\
Q_{\mathcal{T}_i}^{b}(s, a) = C_{\mathcal{T}_i}(s, a) + \sum_{s' \in S}T(s' | s, a)V_{\mathcal{T}_i}^{b}(s')
\end{align*}
where $C$ is cost and $b = [l, u]$ is the lower and upper bound respectively. Each time step is penalized by 1 and movement (as opposed to staying still) by an additional 0.1. This cost structure incentivizes efficiency. The lower-bound was initialized to the Manhattan distance between objects (which ignores barriers). The upper-bound was the sum of the shortest-paths between objects which ignores the possibility of more efficiently passing objects. While BRTDP and these heuristics are useful for the specific spatial environments and subtask structures we develop here, it could be replaced with any other algorithm for finding an approximately optimal single-agent policy for a given sub-task. For details on how BRTDP updates on $V$ and $Q$, see \cite{mcmahan2005bounded}. BRTDP was run until the bounds converged ($\alpha = 0.01, \tau=2$) or for a maximum of 100 trajectories each with up to 75 roll-outs for all models. The softmax during inference used $\beta=1.3$, which was optimized to data from the behavioral experiment. At each time step, agents select the action with the highest value for their sub-task.

When agents do not have any valid sub-tasks, i.e. sub-task is None, they take a random action (uniform across the movement and stay-in-place actions). This greatly improves the performance of the alternative models from Section~\ref{section:compexperiments}: Without this noise, they often get stuck and block each other from completing the recipe. It has no effect on Bayesian Delegation.  

\paragraph{Code for reproducibility}
\label{a:repo}
Our code for the environment and models is included in a zip file under supplemental materials. It can also be found at \href{https://github.com/manycooks/manycooks.github.io}{\texttt{https://github.com/manycooks/manycooks.github.io}}.

\newpage
\section{Pseudocode} 

\begin{algorithm}
\caption{\textit{Bayesian Delegation from agent $i$'s perspective}}
\begin{algorithmic}[1]

\STATE \textbf{Input:} $BRTDP$ \quad \# BRTDP plans in <$s$, $T$, $\mathcal{T}$, $\mathcal{A}$> and outputs $V_{\mathcal{T}}, Q_{\mathcal{T}}$, and $\pi_{\mathcal{T}}$.
\STATE \textbf{Input:} $env$ \quad \# The environment.
\STATE \textbf{Input:} Other terms defined in the main text.

\STATE $s_0 = env.reset()$
\STATE $t = 0$
\WHILE{not $env.done()$}
    \STATE $\mathbf{ta}_t \leftarrow$ task allocation set at time $t$ whose preconditions have been satisfied (Appendix~\ref{appendix:env_details})
    \IF{$t = 0$ or $\mathbf{ta_{t-1}} \neq \mathbf{ta_t}$}
        \STATE \#  Initialize beliefs on the first time step or when task allocation set is updated.
        \FOR{$ta$ in $\mathbf{ta}_t$} 
        \STATE \# Run BRTDP to approximate $V_{\mathcal{T}}, Q_{\mathcal{T}}$.
        \STATE $P(\mathbf{ta_t}) = \Sigma_{\mathcal{T} \in ta}\frac{1}{V_{ \mathcal{T}}(s_t)}$
        \ENDFOR
    \ELSE 
        \STATE \# Otherwise, update beliefs based on action likelihoods.
        \FOR{$ta$ in $\mathbf{ta}_t$} 
        \STATE \# Calculate likelihood (Equation~\ref{eq:softmax}). 
        \STATE $P(\mathbf{a_{t-1}}|s_{t-1}, ta)= \Pi_{\mathcal{T} \in ta} \frac{e^{\beta * Q_{ \mathcal{T}}(s_{t-1}, \mathbf{a_{t-1}})}}{\sum_{\textbf{a} \in \mathcal{A}} e^{\beta * Q_{\mathcal{T}}(s_{t-1}, \mathbf{a})}}$
        \STATE \# Update posterior (Equation~\ref{eq:bayes}). 
        \STATE $P(ta) = P(ta) * P(\mathbf{a_{t-1}}|s_{t-1}, ta) $
        \ENDFOR
    \ENDIF
    \STATE $P(ta) = P(ta) / \sum_{ta} P(ta)$ \quad \# Normalize $P(ta)$. 
    \STATE \# Pick the maximum a posteriori task allocation ($ta^*$).
    \STATE $ta^*$ = $\argmax_{ta}P(ta)$
    \STATE $\mathcal{T}_{i} = ta^*.get(i)$ \quad \# Get own task, $\mathcal{T}_{i}$
    \STATE \# Update transitions with models of others performing $\mathcal{T}_{-i} \in ta^*$  \\
    \STATE $\pi^0_{\mathcal{T}_{-i}} = BRTDP(s_t, T, \mathcal{T}_{-i}, \mathcal{A}_{-i})$
    \STATE $T'(s_{t+1}|s_t,a_{-i}) = \sum_{a_i}T(s_{t+1}|s_t, a_{-i}, a_{i})\Pi_{\mathcal{T}_{-i} \neq \mathcal{T}_{i}} \  \pi^0_{\mathcal{T}_{-i}}(s_t)$
    \IF{$\exists \ \mathcal{T}_{-i} \in ta^* s.t. \mathcal{T}_{-i} = \mathcal{T}_{i}$}
        \STATE \# Plan cooperatively using the joint policy returned by BRTDP.
        \STATE $\mathcal{A'} = \mathcal{A}_i \times \mathcal{A}_{-i}$
        \STATE $\pi^{J}_{\mathcal{T}_{i}} = BRTDP(s_t, T', \mathcal{T}_i, \mathcal{A'})$
        \STATE $a_{i, t} = \argmax_{a^J}\pi^{J}_{\mathcal{T}_{i}}(a^J|s_t, T')[i]$
    \ELSE
        \STATE \# Plan a best response to the level-0 models of other agents.
        \STATE $\pi_{\mathcal{T}_{i}} = BRTDP(s_t, T', \mathcal{T}_i, \mathcal{A}_i)$
        \STATE $a_{i, t} = \argmax_{a_i}\pi_{\mathcal{T}_{i}}(a_i |s_t, T')$
    \ENDIF
    \STATE \# Step all agents' actions and update the environment.
    \STATE $\mathbf{a_{t}}, s_{t+1} \leftarrow env.step(a_{i, t})$
    \STATE $t = t + 1$
\ENDWHILE

\end{algorithmic}
\end{algorithm}

\end{document}